\documentclass[11pt]{article}
\usepackage[margin=1in]{geometry}
\usepackage[T1]{fontenc}
\usepackage{lmodern}
\usepackage{amsmath,amssymb,amsthm,mathtools}
\usepackage{booktabs}
\usepackage{longtable}
\usepackage{multirow}
\usepackage{array}
\usepackage{graphicx}
\usepackage{xcolor}
\usepackage{pgfplots}
\pgfplotsset{compat=1.17}
\usepackage[numbers,sort&compress]{natbib}
\usepackage[colorlinks=true,linkcolor=blue!50!black,citecolor=blue!50!black,urlcolor=blue!50!black]{hyperref}
\usepackage{cleveref}

\newtheorem{theorem}{Theorem}
\newtheorem{lemma}[theorem]{Lemma}
\newtheorem{proposition}[theorem]{Proposition}
\newtheorem{corollary}[theorem]{Corollary}
\theoremstyle{definition}
\newtheorem{definition}[theorem]{Definition}
\newtheorem{assumption}{Assumption}
\theoremstyle{remark}
\newtheorem{remark}[theorem]{Remark}

\newcommand{\todo}[1]{\textcolor{red}{[TODO: #1]}}
\newcommand{\XeightCost}{about 20 tokens in natural language (the compact code of \Cref{sec:exp-rate} needs about 10.5 but is decoded less reliably)}
\IfFileExists{theory/macros.tex}{\input{theory/macros.tex}}{}

\title{Agent Memory as Rate-Limited Filtering:\\ Explicit State Outlasts Long Context Beyond a Reader-Dependent Horizon}
\author{Tianhao Guo \and Chongyao Guo}
\date{}

\begin{document}
\maketitle

\begin{abstract}
Can a language model that reads the full history of an access log say whether a service may act now? On an executable benchmark of grants, withdrawals, reinstatements, expiries, groups and change windows under a known policy, scored on pairs of histories that differ only in the order or one field of the deciding events, open models of 7 to 32 billion parameters fall from 72 to 86 percent after eight or sixteen events to 48 to 55 percent after 256, while reasoning-trained 32B models reading the current state reach 96 and 98 percent. A reasoning model of 120 billion parameters trails the state by four and seven points at 256 and 512 events and by about ten at 1{,}024, so the length beyond which explicit state outlasts long context grows with the reader. We treat agent memory as recursive filtering. The decision depends on the history only through a state whose size is set by the number of tracked keys. A store holding fewer component records than the state forgets on a time scale set by its capacity, and a reader model predicts its accuracy within 0.02 on average in a registered test. Retaining log lines without regard to time order errs with probability approaching one half, a bound that is attained. A filter in which a model parses each line and an explicit rule applies the updates answers 80 to 89 percent at 256 events and beats relevance retrieval, model-kept ledgers, a reimplementation of StateMem and the Mem0 library. A retriever that selects at query time the lines the queried state depends on matches it at 256 events on the first domain, so what decides is selection by the state's dependencies, which the filter performs when writing, without the query. Applying parsed updates themselves, the models fall to 58 and 59 percent. A frozen replication on a second domain with fresh wording reproduces the horizon.
\end{abstract}

\section{Introduction}

Agent memory is usually built as text compression. A summary is rewritten as events arrive, a retriever pulls passages back in, or a scoring rule decides which raw records survive a budget \citep{packer2023memgpt,park2023generative,jiang2023llmlingua,yu2026memagent}. Each design asks how to keep the most informative text. Few ask what the downstream decision depends on, and those that do treat memory as a one-shot partition of independent contexts \citep{zou2026remember} rather than as a state updated over time.

For a large class of agent tasks the answer is a state. Whether a service may write to a database depends on the last effective grant or withdrawal for that pair, on the rank of the sources that issued them, on whether a grant has expired, on group membership and on whether a change window is open. The history matters only through this state. Signal processing has a name for maintaining such a quantity recursively under a resource limit: filtering. Information theory has the matching question: how many bits per step a filter needs before estimation fails \citep{nair2004stabilizability,tatikonda2004control,nair2007feedback}.

This paper takes the filtering view literally and tests it.

\paragraph{Long context does not track state.} We build an executable benchmark in which every answer is produced by replaying events through a state machine (\Cref{sec:benchmark}). Episodes come in pairs that share background, positions and wording and differ only in the order or one field of the decisive events, so that the correct answers swap. Given the full log, Qwen2.5-14B-Instruct reaches a balanced accuracy of 0.83 after 8 events and 0.48 after 256 events. Its ranking accuracy (AUC) falls monotonically from 0.85 to 0.53 across six horizons. Qwen2.5-7B follows the same curve one step lower, phi-4, from a different family, falls from 0.90 to 0.68, and Qwen2.5-32B falls from 0.96 at 16 events to 0.65 at 256. Long inputs are known to hurt reasoning and the retrieval of overwritten values \citep{levy2024same,du2025context,wang2025unable}, and long-horizon memory benchmarks report losses on knowledge updates \citep{wu2025longmemeval,tavakoli2026beyond}. The paired design adds two things. Each pair isolates one order or field change with everything else held fixed, and the margin lets us separate a shift of the decision threshold from a loss of discrimination. Both are present, and at 256 events the 7B and 14B Qwen readers no longer discriminate at all. The failure is asymmetric. Prohibitions survive, reinstatements do not: in the restore family (grant, withdrawal, later reconfirmation) the 14B model refuses 11 of the 19 cases that should be approved at 64 events and the 7B model 15 of 19, and an alarm-worded withdrawal aimed at a different service is enough to make the 7B model refuse the service that holds a valid grant.

\paragraph{State restores the signal.} The same model, given only the current state recovered by replay, ranks correctly with AUC between 0.95 and 0.99 at every horizon, and phi-4 between 0.98 and 0.99 with a balanced accuracy that shows no detectable dependence on the horizon (\Cref{sec:exp-state}). For the Qwen readers a calibration offset remains, and it grows with the number of state lines, not with the number of events. Showing the raw log alongside the state pulls the model back toward its long-context behaviour.

\paragraph{The horizon moves with the reader.} Stronger readers push the failure out but do not remove it. Before running them we registered how their results would change our claims (\Cref{sec:exp-reasoning}). gpt-oss-120b, a reasoning model of 120 billion parameters, reads the 256-event log at 0.95, and the state built from parsed events leads it by only $+0.04$, short of the registered margin. At 1{,}024 events its reading falls to 0.87 while the state stays at 0.99, a lead of $+0.12$ ($[0.07,0.17]$) at about one twentieth of the tokens; with a second sampling seed the lead is $+0.09$ ($[0.04,0.13]$), and at 512 events it is $+0.07$. Qwen2.5-72B is already at 0.72 against 0.94 at 256 events, and QwQ-32B, at 0.85 on 256-event logs, answers 0.52 on 1{,}024. By the rule we fixed in advance, the claim is therefore conditional on length. We define a reader's operational horizon as the shortest log length at which the state beats the log by the registered margin (\Cref{sec:exp-reasoning}); for gpt-oss-120b it lies at about 1{,}024 events. Below it a strong reader can use the log as its memory, at a cost that grows with the log. Beyond it the state keeps its accuracy at a cost set by the number of tracked keys. No theorem of this paper limits a reader that sees the whole ordered log, since exact replay of the log is sufficient (\Cref{sec:theory}); the horizon is a measured property of readers.

\paragraph{Length more than same-key interference.} A factorial design on chain counterfactuals varies the number of events on the queried key, the number of events on other keys and the total length independently. Raising the updates of the queried key from 3 to 17 moves AUC by $+0.007$ and $-0.015$ for Qwen 14B and Qwen 7B, with intervals of about $\pm0.05$ around these values, while more events on other keys and a longer log each cost 0.07 to 0.13 with intervals that exclude zero (\Cref{sec:exp-decay}). Within the tested range the point estimates for the queried key are near zero and its intervals reach about 0.07 AUC, so the ordering is an upper bound, not an absence. Most of the loss comes from dilution by everything else, which is exactly what a filter removes. Same-key interference is not absent, since a short view of the key's own events still misleads the 14B reader in \Cref{sec:exp-separation}.

\paragraph{The filtering view predicts which memories can work.} The decision-relevant statistic has a description length set by the number of tracked keys. A recursive filter that stores it needs a rate independent of the horizon when leases are bounded, and no memory below that rate can answer all queries (\Cref{sec:theory}). A memory that retains raw events by any score computed without regard to order, whether lexical relevance, salience or a model's judgment of each line, cannot know which events have been superseded. On reordered histories its error is at least $\tfrac12-b/(m-1)$ for a budget of $b$ lines and $m$ events on the key, and this bound is attained. Keeping the last applied event of each state component is exact for a reader that applies the policy. Controlled chains confirm the separation and both retention results with an exact reader, and a slot-aware reader of a random subset of withdrawals attains the bound (\Cref{sec:exp-separation}), and a rate sweep shows that the budget needed by state memory tracks the size of the state (\Cref{sec:exp-rate}).

\paragraph{Models do not maintain the filter on request.} Asked to keep a ledger of the current state, rewritten every 16 events within a token budget, Qwen2.5-14B reaches chance after 256 events. It does not run out of space. Its final ledgers use 171 of 256 tokens and mention the queried key in 8 of 80 streams, because the writer drops components it judges inactive and a dropped component never returns (\Cref{sec:exp-recursive}). Yet the same model parses 35{,}743 of 35{,}744 event lines correctly when shown one line at a time. A filter built from these parses and an explicit transition rule answers 0.80 at 256 events with the Qwen reader, 0.84 with phi-4 and 0.89 with Qwen2.5-32B. The strongest competitor in which the model reads the log, a reasoning trace of up to 4096 tokens, reaches 0.64 and 0.73, and retrieval at most 0.68 (\Cref{sec:exp-buildable}). Handing the model the same parsed updates and asking it to apply them to a state table drops it to 0.59 and 0.58, and our re-implementation of Mem0-style fact editing ends at 0.59 and 0.54. A table kept by code, which the model may overwrite but not delete from, closes about half of the remaining gap for one reader. The model can read events. It loses components when it rewrites its own record, and even a table it cannot delete from answers below the rule, and for the Qwen reader, the more of the transition the memory executes, the better the result. Asked to write the transition code itself, a 32B code model succeeds on a simple second domain and fails on the rank rule of this one, while its failure on reinstatement traces to a gap in our update format. A frontier coding agent given the same policy text writes a rule that agrees with replay once that gap is closed, a change we made after seeing its result.

\paragraph{What decides is selection by the state's dependencies.} A filter whose store is smaller than the state forgets, at a rate the theory predicts within 0.02 of measured accuracy on average, and the published design closest to ours, StateMem, stays at or near chance on the priority family because its model resolves supersession by recency. The Mem0 library retrieves by similarity to the query, without preferring the latest line, and is at or below chance on most families. A retriever that keeps the whole log and, once the query is known, selects the lines the queried state depends on reads as well as the filter at 256 events on the first domain. The advantage of the filter is therefore not that it compresses the log but that it selects by the state's dependencies when it writes, before any query is known, with storage bounded by the state (\Cref{sec:exp-buildable,sec:exp-budget}).

\paragraph{Contributions.}
\begin{enumerate}
\item A tight bound for order-oblivious retention. On histories that differ only in the order of two events, any rule that retains $b$ log lines chosen without regard to their time order, such as lexical, embedding or per-line salience scores on text stripped of time stamps, gives every reader an error of at least $\tfrac12-b/(m-1)$ averaged over wording, where $m$ is the number of events on the queried key. The error tends to one half when $b=o(m)$, the bound is attained, keeping the last applied event of each state component is exact with a budget set by the number of components, and scorers that read the log in context are covered only up to an empirically estimated slack (\Cref{sec:theory}).
\item A noisy-parser analysis and a buildable filter. Per-event parsing errors are forgotten at the next applied event of one rank and persist behind blocked events of higher rank, a simulation from calibrated per-line error rates ranks parsers of 0.5 to 14 billion parameters with a rank correlation of 0.83 although their errors are not independent, and a calibrated confusion correction, whether line by line or sequential, brings the 0.5B and 1.5B parsers to the error of the 3B parser. A filter that parses each line with a model and applies an explicit rule beats relevance retrieval, model-maintained ledgers, fact stores, a reimplementation of StateMem and the Mem0 library when read by the same model, beats reasoning over the full log for reasoning-trained readers and, after multiplicity adjustment, for one of two instruction-tuned readers, and beats the best code-kept ledger protocol significantly for one of two readers. A retriever that selects at query time the lines the queried state depends on matches it at 256 events, which locates the advantage in selection by state dependencies rather than in compression (\Cref{sec:theory,sec:exp-buildable,sec:exp-reasoning,sec:exp-parsers}).
\item An executable benchmark with paired counterfactual streams, a replay oracle, a paraphrased variant and a pre-registered reader-validity protocol (\Cref{sec:benchmark}).
\item Measurements across eight readers of horizon decay and its causes, of a reader-dependent horizon up to 1{,}024 events, of state-based reading, of retention against the bounds, of recursive memories under a budget including model-maintained ledgers, and of the token cost of state in natural and compact encodings (\Cref{sec:experiments}).
\item The cost of state in three units. In bits it is a classical instantiation, with thresholds far below every budget tested. In components the reader can decode, a calibrated component-eviction model of the reader predicts, in a registered test, how accuracy falls when the reader sees only part of the state, within 0.03 on average. For a bounded store that forgets what it evicts, the same model with the store's own error gives a retention horizon set by capacity and insertion rate (\Cref{sec:theory,sec:exp-budget}). In reader tokens, the two renderings we tested cost about twenty tokens per component, and building the state costs one parser call per event (\Cref{sec:exp-cost}).
\end{enumerate}
All hypotheses, thresholds and seed ranges were written before the corresponding runs. The pre-registration log is reproduced in \Cref{app:prereg}.

\IfFileExists{theory/theory.tex}{
\makeatletter
\@ifundefined{theorem}{\newtheorem{theorem}{Theorem}}{}
\@ifundefined{lemma}{\newtheorem{lemma}[theorem]{Lemma}}{}
\@ifundefined{proposition}{\newtheorem{proposition}[theorem]{Proposition}}{}
\@ifundefined{corollary}{\newtheorem{corollary}[theorem]{Corollary}}{}
\@ifundefined{conjecture}{\newtheorem{conjecture}[theorem]{Conjecture}}{}
\@ifundefined{definition}{\theoremstyle{definition}\newtheorem{definition}[theorem]{Definition}}{}
\@ifundefined{assumption}{\theoremstyle{definition}\newtheorem{assumption}{Assumption}}{}
\@ifundefined{remark}{\theoremstyle{remark}\newtheorem{remark}[theorem]{Remark}}{}
\makeatother
\providecommand{\cO}{\mathcal{O}}
\providecommand{\cQ}{\mathcal{Q}}
\providecommand{\cM}{\mathcal{M}}
\providecommand{\cK}{\mathcal{K}}
\providecommand{\cD}{\mathcal{D}}
\providecommand{\Prob}{\mathbb{P}}
\providecommand{\E}{\mathbb{E}}
\providecommand{\TV}{\mathrm{TV}}
\providecommand{\ind}{\mathbf{1}}
\providecommand{\rk}{\mathrm{rk}}
\providecommand{\Ret}{\mathrm{Ret}}
\providecommand{\NQ}{N_{\cQ}}
\makeatletter
\providecommand{\thlabel}[2]{\@ifpackageloaded{cleveref}{\label[#1]{#2}}{\label{#2}}}
\@ifpackageloaded{cleveref}{%
  \@ifundefined{cref@conjecture@name}{\crefname{conjecture}{conjecture}{conjectures}\Crefname{conjecture}{Conjecture}{Conjectures}}{}%
  \@ifundefined{cref@assumption@name}{\crefname{assumption}{assumption}{assumptions}\Crefname{assumption}{Assumption}{Assumptions}}{}%
}{}
\makeatother
\providecommand{\XeightCost}{\textbf{[X8: tokens per component]}}

\section{Theory: supersession, retention and the cost of state}\label{sec:theory}

An agent that acts on a log never needs the log itself.
It needs the answers to the questions it may be asked, now and after any further events.
We make this precise for the access-control process of our benchmark.
Two results do the work in this paper.
First, a memory that keeps raw log lines chosen without regard to their time order fails on histories that differ only in the order of two events, for every reader, and the bound is attained, while keeping the last applied event of each state component is exact (Section~\ref{sec:theory-extract}).
Second, when a language model parses each line into a structured event and an exact rule updates the state, parse errors on the queried key are forgotten after the next applied event, while rank errors on events that the rule blocks persist and accumulate (Section~\ref{sec:theory-parser}).
We then state what the state costs. These rate results are classical (Myhill--Nerode, a fooling-set argument and Fano's inequality) instantiated for this process, and their thresholds lie far below every budget we test (Section~\ref{sec:theory-rate}).
Finally we propose a falsifiable model of why long-context readers lose the state as $T$ grows (Section~\ref{sec:theory-reader}). That part is a model with stated assumptions, not a theorem.

\subsection{Model}\label{sec:theory-model}

\begin{definition}[Access process]\thlabel{definition}{def:process}
Let $\mathcal{P}$ and $\mathcal{R}$ be finite nonempty sets of principals and resources, $\mathcal{G}$ a finite (possibly empty) set of groups, and $[L]=\{1,\dots,L\}$ with $L\ge1$ the source ranks.
Nonemptiness matters. With $\mathcal{P}=\varnothing$ the query family is empty and every history is equivalent, so the counts below would fail.
A \emph{key} is a pair $\kappa=(p,r)\in\cK:=\mathcal{P}\times\mathcal{R}$, and $K:=|\cK|$.
At each time $t=1,2,\dots$ one observation $O_t\in\cO$ arrives. It is one of
\begin{itemize}
\item a \emph{direct event} $(\kappa,a,\ell,\lambda)$ with decision $a\in\{0,1\}$ (grant or reconfirmation if $a=1$, withdrawal if $a=0$), source rank $\ell\in[L]$ and lease $\lambda\in\{0,\dots,\Lambda\}\cup\{\infty\}$, where $\lambda=\infty$ whenever $a=0$ and a lease grant issued at time $t$ is valid through time $t+\lambda$,
\item a group event $\mathrm{gg}(g,r)$ (group $g$ may write to $r$), $\mathrm{add}(p,g)$ or $\mathrm{rem}(p,g)$,
\item a window event $\mathrm{req}(r)$ ($r$ accepts writes only while its window is open), $\mathrm{open}(r)$ or $\mathrm{close}(r)$,
\item the null event $\varnothing$ (log lines that change nothing).
\end{itemize}
The state is $S_t=(d_t,\mu_t,\gamma_t,\nu_t,\omega_t)$ with direct records $d_t:\cK\to\{\bot\}\cup\{0,1\}\times[L]\times(\mathbb{N}\cup\{\infty\})$, memberships $\mu_t\subseteq\mathcal{P}\times\mathcal{G}$, group grants $\gamma_t\subseteq\mathcal{G}\times\mathcal{R}$, window requirements $\nu_t\subseteq\mathcal{R}$ and open windows $\omega_t\subseteq\mathcal{R}$, all empty (or $\bot$) at $t=0$.
The \emph{rank at time $t$} of a record is $\rk_t(\bot)=0$, $\rk_t(a,\ell,u)=\ell$ if $u\ge t$ and $\rk_t(a,\ell,u)=0$ if $u<t$.
A direct event $(\kappa,a,\ell,\lambda)$ at time $t$ is \emph{applied} if $\ell\ge\rk_t(d_{t-1}(\kappa))$, and then $d_t(\kappa)=(a,\ell,t+\lambda)$. Otherwise $d_t=d_{t-1}$.
Group and window events act on $\mu,\gamma,\nu,\omega$ as set insertions and deletions, and $\mathrm{gg}$ and $\mathrm{req}$ are never undone.
A query $q=(p,r)$ posed after time $t$ has the answer
\begin{equation}\label{eq:answer}
f_q(S_t)=\Bigl(\bigl[d_t(\kappa)=(1,\ell,u),\ u\ge t\bigr]\ \vee\ \bigl[\exists g:(p,g)\in\mu_t,(g,r)\in\gamma_t\bigr]\Bigr)\wedge\bigl[r\notin\nu_t\ \vee\ r\in\omega_t\bigr],
\end{equation}
with $1=\textsf{APPROVE}$. The query family is $\cQ=\cK$.
\end{definition}

Definition~\ref{def:process} transcribes the benchmark's executable ground truth line by line (\texttt{replay()} in the released generator).
Grants and reconfirmations are $a=1$, both withdrawal wordings are $a=0$, and $\mathrm{add}$ covers both membership wordings.
The generator writes absolute deadlines (``valid through Event $N$'') and the reader sees the event index, so the relative lease $\lambda=N-t$ is observable.
We use relative leases because they make the process time homogeneous.
The benchmark has $|\mathcal{P}|=6$, $|\mathcal{R}|=4$, $|\mathcal{G}|=4$ and $L=3$, so $K=24$.
Its generator draws leases only in the expiry family, with $\lambda\le T+19$.
Write $C_{\mathrm{aux}}:=|\mathcal{P}||\mathcal{G}|+|\mathcal{G}||\mathcal{R}|+2|\mathcal{R}|$ for the number of binary group and window components and $K_{\mathrm{tot}}:=K+C_{\mathrm{aux}}$ for the number of state components.

\begin{definition}[Recursive memory, memory size and budget]\thlabel{definition}{def:memory}
A \emph{recursive memory} of memory size $R$ bits is an alphabet $\cM$ with $\log_2|\cM|\le R$, an initial state $M_0$, update maps $M_t=U_t(M_{t-1},O_t)$ and decision maps $\hat A_{t,q}=g_t(M_t,q)$.
Updates and decisions may be randomized with private randomness.
Since $M_t$ is the only quantity passed from step $t$ to step $t+1$, the memory size $R$ is also the rate, in bits per step, of the channel that carries information forward in time.
A memory held as at most $B$ tokens over a vocabulary with $|\mathcal{V}|\ge2$ has $|\cM|\le\sum_{j=0}^{B}|\mathcal{V}|^j\le|\mathcal{V}|^{B+1}$, so its capacity is at most $(B+1)\log_2|\mathcal{V}|$ bits.
\end{definition}

\begin{definition}[Distortion]\thlabel{definition}{def:distortion}
The \emph{anytime worst-case distortion} over horizon $T$ is
\[
D_{\mathrm{wc}}(T)=\sup_{h\in\cO^T}\ \max_{t\le T}\ \max_{q\in\cQ}\ \Prob\bigl(\hat A_{t,q}\neq f_q(S_t)\bigr).
\]
The decision at time $t$ is made on the prefix of length $t$, so $D_{\mathrm{wc}}(T)$ also covers every shorter history followed by a continuation, up to length $T$.
For a distribution of histories, the \emph{average distortion} is the expected fraction of wrong decisions.
A memory that must work for every horizon without knowing it in advance is held to the anytime criterion, because every time may be the last.
We also consider the \emph{end-of-horizon} criterion, where only $t=T$ counts and $T$ is known. For a recursive memory with one alphabet it does not lower the size requirement (Proposition~\ref{prop:recursion}). It does for a one-shot memory computed from the whole log at time $T$.
\end{definition}

\subsection{Extractive memories}\label{sec:theory-extract}

Many practical memories do not compute a state. They keep some of the observed lines.
Let the \emph{content} $c_t$ of $O_t$ be the observation without its time stamp, so $O_t=(t,c_t)$.

\begin{definition}[Extractive memory]\thlabel{definition}{def:extractive}
An extractive memory with budget $b$ retains a random set $S\subseteq[T]$ with $|S|\le b$, possibly depending on the query, and the reader sees the \emph{view} $V=\{(t,c_t):t\in S\}$, together with $q$ and $T$.
If every retained record costs at least $c_{\min}$ tokens, a token budget $B$ implies $b\le\lfloor B/c_{\min}\rfloor$. (An upper bound on the cost per line gives no such implication: cheap lines would let more of them fit.)
\end{definition}

\begin{definition}[Order-oblivious retention]\thlabel{definition}{def:oblivious}
Let $\xi_1,\dots,\xi_T$ be auxiliary random keys, i.i.d.\ with a continuous law and independent of the complete input log. A retention rule $S=\Ret(c_{1:T},\xi_{1:T},q)$ is \emph{order oblivious} if
$\Ret(c\circ\pi,\xi\circ\pi,q)=\pi^{-1}\bigl(\Ret(c,\xi,q)\bigr)$ for every permutation $\pi$ of $[T]$, where $(c\circ\pi)_t=c_{\pi(t)}$.
In words, the rule is an equivariant selection of event occurrences: it sees the events, each tagged with its random key, as an unordered collection.
\end{definition}

Top-$b$ retention by any score $\sigma(c_t,q)$ with ties broken by the continuous keys $\xi_t$ is order oblivious.
(Being a function of the multiset of contents is not enough. The constant output $\{1\}$ is such a function and is not equivariant.)
This covers BM25 and TF--IDF over the line text (corpus statistics depend only on the multiset), dense similarity to the query, keyword or alarm-word salience, and importance judged by a model one line at a time.
It also covers ``keep the lines that mention the queried key'', truncated to $b$ lines by the random keys when there are more.
The time stamp must be excluded from the scored text. Otherwise lexical scorers see position, as BM25 would by matching the event index in the request line.
Removing the explicit time stamp is necessary but not sufficient: increasing ticket numbers or any other position marker in the content would also leak order.
Scores computed with the whole log in context, such as gradient attributions, attention mass or surprisal, depend on order through the scoring model. Corollary~\ref{cor:delta} covers them with a measurable slack.
Recency and sliding windows are not order oblivious. They are treated in Proposition~\ref{prop:recency}.

\begin{definition}[Chain counterfactuals]\thlabel{definition}{def:chain}
Fix a key $\kappa=(p,r)$, a rank $\ell$ and an odd $m\ge3$.
The key receives $m$ direct events at slots $s_1<\dots<s_m$, all of rank $\ell$ and without leases.
Their decisions in version \textsf{A} are $1,0,1,0,\dots,0,1$ (grant, then alternating withdrawals and reconfirmations, ending with a reconfirmation).
Each key event carries a nuisance $\nu_j$ (wording, ticket) drawn i.i.d.\ from one fixed law common to all key events, independent of the slots and of the background.
The other slots hold background events that are not direct events on $\kappa$, not membership events of $p$, and not group or window events on $r$.
Version \textsf{D} exchanges the contents at slots $s_{m-1}$ and $s_m$, so $c^{\textsf{D}}=c^{\textsf{A}}\circ\tau$ with $\tau$ the transposition of these slots.
The answer to $q=\kappa$ at time $T$ is \textsf{APPROVE} in version \textsf{A} and \textsf{DENY} in version \textsf{D}, because all key events are applied and the last one decides.
\end{definition}

This is the restore and order families of the benchmark with a chain whose length grows.

\begin{lemma}[Two-point bound, cf.~\cite{lecam1973convergence}]\thlabel{lemma}{lem:lecam}
For any reader, possibly randomized, the error averaged over the two versions with equal weight is at least $\frac12\bigl(1-\TV(P_{V^{\textsf{A}}},P_{V^{\textsf{D}}})\bigr)$, with equality for the best reader.
\end{lemma}

\begin{proof}
Let the reader answer \textsf{APPROVE} with probability $\phi(v)\in[0,1]$ on view $v$. Its error is $\frac12\bigl(1-\E\phi(V^{\textsf{A}})+\E\phi(V^{\textsf{D}})\bigr)$.
Let $p_{\textsf{A}},p_{\textsf{D}}$ be densities of the two view laws with respect to their sum.
Then $\E\phi(V^{\textsf{A}})-\E\phi(V^{\textsf{D}})=\int\phi\,(p_{\textsf{A}}-p_{\textsf{D}})\le\int(p_{\textsf{A}}-p_{\textsf{D}})_+=\TV$, with equality for $\phi=\ind\{p_{\textsf{A}}>p_{\textsf{D}}\}$.
\end{proof}

\begin{theorem}[A tight bound for order-oblivious retention]\label{thm:separation}
For every order-oblivious extractive memory with budget $b$ and every reader, the error on the chain counterfactual pair, averaged over the two versions, the nuisances and the auxiliary randomness, satisfies
\begin{equation}\label{eq:separation}
\max\Bigl\{0,\ \frac12-\frac{b}{m-1}\Bigr\}\ \le\ \mathrm{Err}^\star\ \le\ \frac12,
\end{equation}
where $\mathrm{Err}^\star$ is the error of the best reader. Every reader has error at least $\mathrm{Err}^\star$.
The bound is attained (Proposition~\ref{prop:tight}).
If $b=o(m)$, in particular for a fixed budget while the key's chain grows with $T$, every reader's error is at least $\frac12-o(1)$.
\end{theorem}

\begin{proof}
Couple the versions by running the rule on $(c^{\textsf{A}},\xi)$ and on $(c^{\textsf{D}},\xi\circ\tau)$. Because $\xi$ is i.i.d.\ and independent of the log, the second input has the joint law of version \textsf{D}.
By Definition~\ref{def:oblivious}, $S^{\textsf{D}}=\tau(S^{\textsf{A}})$.
On the event $S^{\textsf{A}}\cap\{s_{m-1},s_m\}=\varnothing$ we get $S^{\textsf{D}}=S^{\textsf{A}}$, and since $\tau$ fixes $S^{\textsf{A}}$ pointwise the views coincide.
Hence $\TV(P_{V^{\textsf{A}}},P_{V^{\textsf{D}}})\le\Prob(s_{m-1}\in S^{\textsf{A}})+\Prob(s_m\in S^{\textsf{A}})$.

Let $n=(m-1)/2$, and let $W$ and $C$ be the sets of withdrawal and reconfirmation slots of version \textsf{A}, each of size $n$.
Fix the slots and the background and condition on them. For any permutation $\pi$ of $W$, the joint conditional law of $(c\circ\pi,\xi\circ\pi)$ equals that of $(c,\xi)$, because the withdrawal contents are one map applied to i.i.d.\ nuisances and $\xi$ is i.i.d.\ and independent of the log. The background need not be exchangeable.
Pathwise equivariance then gives $\Prob(w\in S^{\textsf{A}})=\Prob(\pi(w)\in S^{\textsf{A}})$, so every withdrawal slot is retained with probability $\E X_W/n$ where $X_W=|S^{\textsf{A}}\cap W|$. Likewise every reconfirmation slot is retained with probability $\E X_C/n$, $X_C=|S^{\textsf{A}}\cap C|$.
Slot $s_{m-1}$ is in $W$ and $s_m$ is in $C$, and the two counts share one budget, so
$\Prob(s_{m-1}\in S^{\textsf{A}})+\Prob(s_m\in S^{\textsf{A}})=\E(X_W+X_C)/n\le b/n=2b/(m-1)$.
Lemma~\ref{lem:lecam} gives $\mathrm{Err}^\star\ge\frac12(1-\min\{1,2b/(m-1)\})$.
\end{proof}

The failure is a memory failure, not a reading failure, since the bound holds for an unbounded reader.
The bound is an average over nuisances and auxiliary randomness. It does not hold for each fixed nuisance realization, and it needs $\xi$ to be independent of the log, not merely i.i.d.\ (a key $\xi_T=1-\nu_T$ that is correlated with the last nuisance would let an equivariant rule single out the last event).
\begin{remark}[What the proof uses]\thlabel{remark}{rem:general}
The proof of Theorem~\ref{thm:separation} uses no property of the access rule beyond two. First, the two versions differ by transposing the contents of two slots $i$ and $j$, and the query answer is a deterministic function of the history that differs between the versions (a query-separated pair of updates that do not commute). Second, $i$ lies in a set $U$ and $j$ in a disjoint set $V$ of slots such that the joint law of the version-\textsf{A} contents and the auxiliary keys is invariant under permutations within $U$ and within $V$, given the other slots (exchangeable nuisances).
Under these two conditions alone, for any process whose answers are deterministic functions of the history, in particular any finite-state process, every order-oblivious memory of budget $b$ gives $\mathrm{Err}^\star\ge\frac12\bigl(1-\min\{1,\E X_U/|U|+\E X_V/|V|\}\bigr)\ge\frac12-\frac{b}{2\min\{|U|,|V|\}}$, where $X_U=|S^{\textsf{A}}\cap U|$ and $X_V=|S^{\textsf{A}}\cap V|$.
The proof is the one above with $W,C,n$ replaced by $U,V,|U|,|V|$, and the last inequality uses $X_U+X_V\le b$. When the contents in $U$ form a class that a content-only score can recognize and $|U|=|V|$, keeping $\min\{b,|U|\}$ of them uniformly at random attains the bound, by the argument of Proposition~\ref{prop:tight}.
\end{remark}

The theorem also holds when the nuisance stays with the slot rather than with the event, as in the benchmark generator. Because withdrawals and reconfirmations draw their nuisance from one common law, the two nuisances at the swapped slots are i.i.d., so version \textsf{D} built either way has the same law, and the total variation bound depends only on the laws.

\begin{proposition}[The bound is attained]\thlabel{proposition}{prop:tight}
Let the rule keep $b'=\min\{b,n\}$ of the key's withdrawal events, chosen uniformly by the random keys, with $n=(m-1)/2$.
It is order oblivious, $\TV(P_{V^{\textsf{A}}},P_{V^{\textsf{D}}})=b'/n$, and $\mathrm{Err}^\star=\frac12(1-b'/n)=\max\{0,\frac12-\frac{b}{m-1}\}$.
\end{proposition}

\begin{proof}
The withdrawal slots of the two versions share $n-1$ slots, and each version has one more ($s_{m-1}$ in \textsf{A}, $s_m$ in \textsf{D}).
The retained set avoids the special slot with probability $\binom{n-1}{b'}/\binom{n}{b'}=1-b'/n$, and then it is a uniform $b'$-subset of the common slots in both versions, so the views have the same law.
Otherwise the view contains a withdrawal at $s_{m-1}$ or at $s_m$, which identifies the version. Hence $\TV=b'/n$, and Lemma~\ref{lem:lecam} gives the error.
\end{proof}

The computation conditions on the slots, as Definition~\ref{def:chain} does, so attainment is for a reader that knows where the key's events sit. With random slots unknown to the reader the views carry less information, the total variation can only fall, and the error of this rule can only rise. The lower bound of Theorem~\ref{thm:separation} holds in both cases.

\begin{corollary}[Approximately order-oblivious rules]\thlabel{corollary}{cor:delta}
Let $\Ret^\circ$ be order oblivious with budget $b$, and run $\Ret$ and $\Ret^\circ$ on the same realized log of version $X\in\{\textsf{A},\textsf{D}\}$, with $\delta_X=\Prob\bigl(\Ret(c^X,\cdot)\ne\Ret^\circ(c^X,\cdot)\bigr)$ under some coupling of their randomness.
Then every reader's error is at least $\max\bigl\{0,\frac12-\frac{b}{m-1}-\frac{\delta_{\textsf{A}}+\delta_{\textsf{D}}}{2}\bigr\}$.
\end{corollary}

\begin{proof}
On the same log, equal retained sets give equal views, so $\TV(P_{V^X},P_{V^{\circ X}})\le\delta_X$.
The triangle inequality gives $\TV(P_{V^{\textsf{A}}},P_{V^{\textsf{D}}})\le2b/(m-1)+\delta_{\textsf{A}}+\delta_{\textsf{D}}$, and Lemma~\ref{lem:lecam} concludes.
\end{proof}

The coupling must be on the same realized log. Coupling only the laws of the retained index sets is not enough, because a rule can choose which withdrawal to keep according to the version and so carry the answer in the content of the kept line.
For a scorer that reads the log in context, a natural comparator $\Ret^\circ$ is: attach the continuous random keys, sort the events by key, run the scorer on the sorted log, and map the selected positions back to the original indices. Without the map back, equivariance fails.
The rate at which the two retained sets differ is an empirical estimate $\hat\delta$ of $\delta$. It gives a guaranteed bound only through an upper confidence limit, and otherwise it is an empirical slack.

\begin{proposition}[A filter with raw-line state encoding is exact]\thlabel{proposition}{prop:aware}
Keep, for each state component, the last \emph{applied} event that set it: for each key the last applied direct event, for each membership the last $\mathrm{add}$ or $\mathrm{rem}$, for each group grant and window requirement its first occurrence, and for each window the last $\mathrm{open}$ or $\mathrm{close}$.
This rule is a filter: to decide which line to keep it must execute the transition rule (it compares ranks and deadlines), and it stores the resulting state as raw lines. It can be maintained online, never holds more than $K_{\mathrm{tot}}$ events (a budget of $K_{\mathrm{tot}}\,c_{\max}$ tokens when every line costs at most $c_{\max}$ tokens), and a reader that replays the retained events in time order answers every query exactly at every time, with or without leases.
\end{proposition}

\begin{proof}
Whether a new direct event is applied depends only on its rank, the current time and the retained record of its key, so the rule is computable online.
After the last applied event $e^\ast$ of a key, no later event on that key was applied, so the full replay ends with the record that $e^\ast$ wrote. Expiry depends only on that record and the current time.
Replaying the retained events alone writes the same record, because $e^\ast$ is then the only event on its key and is applied to $\bot$.
Direct events on different keys do not interact, and the set components depend only on their last insertion or deletion.
\end{proof}

\begin{proposition}[Authority-blind retention fails behind blocked events]\thlabel{proposition}{prop:recency}
Call a retention rule \emph{authority blind} if its retained set is a function of the events with their rank and decision fields removed (slots, time stamps, key mentions and the remaining wording), and of auxiliary randomness independent of the log.
Global recency (the last $b$ lines), key-filtered recency (the last $b$ lines that mention $\kappa$), and any rule that scores time, keys or wording but not sources or decisions are authority blind.
Let $\kappa=(p,r)$ receive one top-rank direct event $e^\ast$ at slot $s^\ast$ and $b'$ other direct events of lower rank, and let two versions differ only in the decision $a\in\{0,1\}$ of $e^\ast$.
As in Definition~\ref{def:chain}, assume that nothing masks the direct record: the log has no membership event of $p$ and no group or window event on $r$, and $e^\ast$ has no lease or, in the version with $a=1$, a lease valid through the query time.
\begin{enumerate}
\item[(a)] For every authority-blind rule and every reader, $\mathrm{Err}^\star\ge\frac12\bigl(1-\Prob(s^\ast\in S)\bigr)$. Key-filtered recency with at least $b$ of the lower-rank events after $s^\ast$ has $\Prob(s^\ast\in S)=0$ and error exactly $\frac12$, and so does global recency whenever at least $b$ lines follow $s^\ast$ that are identical in both versions and do not change the answer.
\item[(b)] If $s^\ast$ is uniform over the $b'+1$ slots of $\kappa$'s events, and the authority-free contents of the events on $\kappa$ are i.i.d.\ and independent of $s^\ast$ and of the rest of the log, then every authority-blind rule has $\Prob(s^\ast\in S)\le b/(b'+1)$ and $\mathrm{Err}^\star\ge\frac12-\frac{b}{2(b'+1)}$.
\end{enumerate}
\end{proposition}

\begin{proof}
Earlier events on $\kappa$ have lower rank and are replaced by $e^\ast$. Later ones are blocked, because $e^\ast$ holds the top rank through the query time. With no membership of $p$ and no group or window event on $r$, the answer is the direct decision, so it is $a$.
The two versions agree on every field the rule sees, so with the same auxiliary randomness they retain the same set $S$. If $s^\ast\notin S$ the views coincide, hence $\TV\le\Prob(s^\ast\in S)$ and Lemma~\ref{lem:lecam} gives (a). For key-filtered recency the retained lines are the last $b$ events on $\kappa$, all after $s^\ast$.
For (b), $S$ is a function of authority-free contents and auxiliary randomness whose joint law does not depend on $s^\ast$, so $S$ is independent of $s^\ast$, and $\Prob(s^\ast\in S)=\E|S\cap\mathrm{slots}(\kappa)|/(b'+1)\le b/(b'+1)$.
The rule of Proposition~\ref{prop:aware} retains the decisive event.
\end{proof}

Each hypothesis is needed. If $p$ belongs to a group that may write to $r$, both versions answer \textsf{APPROVE}. If $r$ requires a window that is closed, both answer \textsf{DENY}. If the lease of $e^\ast$ ends before the query, the version with $a=1$ can answer \textsf{DENY} like the other, with no later event or with a later lower-rank withdrawal that the expiry lets through. In each case the versions agree and no error is forced.
The blocked family of the benchmark (experiment X5) satisfies the hypotheses. Its background holds direct events on other keys only, and $e^\ast$ carries no lease (\texttt{theory/boundary\_cases.py}, case A, on $240$ episodes).

\begin{lemma}[Supersession is a scalar order without leases]\thlabel{lemma}{lem:scalar}
Without leases, the last applied direct event on a key is the one that maximizes (rank, time) in lexicographic order among the key's events.
\end{lemma}

\begin{proof}
Without leases the rank of the record never decreases, because an event is applied only if its rank is at least the current one.
If an event of rank higher than that of the last applied event $e^\ast$ occurred before $e^\ast$, it would have been applied and $e^\ast$ would have been blocked. If it occurred after, it would have been applied.
So $e^\ast$ has maximal rank, and among events of maximal rank, all of which are applied, it is the latest.
\end{proof}

Lemma~\ref{lem:scalar} shows that a single-event score $\sigma(O_t;\kappa)=\ind\{O_t\text{ is on }\kappa\}\cdot(\ell,t)$ with budget one per key is exact without leases.
This score is not supersession unaware. It encodes the rank rule.
The bound of Theorem~\ref{thm:separation} is therefore about order, not about scoring one line at a time. Its content is the exact rate $b/(m-1)$ and its attainment (Proposition~\ref{prop:tight}), not the existence of a gap, which follows from symmetry alone.
Leases break the scalar structure.
Whether a later low-rank event was applied depends on the pair formed by its time and an earlier grant's deadline.
For example, with a top-rank lease grant valid through time $10$ followed by one rank-one grant, the answer at time $20$ is \textsf{APPROVE} if the rank-one grant came after time $10$ and \textsf{DENY} if it came before. With budget one per key, both versions retain the same top-scoring line under $(\ell,t)$, namely the lease grant.
Similar pairs defeat the score that first discards grants expired at query time.

Call a score \emph{fixed} if it is a function $\sigma(c_t,\ell_t,t,T_q)$ of one event's content, rank and time stamp and of the query time $T_q$, chosen before the history is seen.
With one rank the question is settled. Every direct event is then applied, so keeping the last event per key and checking its expiry at the query time is exact with leases (checked exhaustively on all histories of up to five events on one key, \texttt{theory/boundary\_cases.py}, case B).

\begin{conjecture}\thlabel{conjecture}{conj:lease}
With leases and $L\ge2$ ranks, no fixed single-event score with top-$b$ retention and $b$ bounded per key is exact on all histories.
\end{conjecture}

We tried only the two natural scores above, $(\ell,t)$ and the rank discounted by expiry at query time, and found the counterexamples given (both checked on the exact rule, case B of the same script). We did not search systematically over classes of scores.

\paragraph{A taxonomy of retention.}
The three results order extractive memories by what their selection can see.
Retention that ignores time order cannot tell which retained line is still in force, and on order counterfactuals its error is at least $\frac12-b/(m-1)$ (Theorem~\ref{thm:separation}, attained by Proposition~\ref{prop:tight}).
Retention that sees time but not authority fails behind blocked events, with error at least $\frac12(1-\Prob(s^\ast\in S))$ (Proposition~\ref{prop:recency}).
Retention that executes the transition rule is exact with a budget set by the number of components (Proposition~\ref{prop:aware}), and such a rule is a filter that stores its state as raw lines.

\paragraph{Implication for the benchmark.}
The generator places at most four events on the queried key, so Theorem~\ref{thm:separation} only bites when $b<m$.
Testing it requires a chain family whose length grows at fixed budget. The theorem predicts error rising toward $\frac12$, at least $\frac12-b/(m-1)$, for content-only retrieval, against zero for the rule of Proposition~\ref{prop:aware}.
Selection is the only restriction of Theorem~\ref{thm:separation}. The reader of Definition~\ref{def:extractive} already sees the time stamp of every retained line, so re-sorting the selected lines by time or showing their event indices does not leave the class. A rule whose selection depends on position does, for example one that breaks ties toward later lines. Where such a rule fails on the benchmark streams, the theorem does not explain the failure.

\subsection{A noisy parser feeding an exact filter}\label{sec:theory-parser}

A practical recursive memory can split the work. A language model parses each log line into a structured event, and an exact transition rule updates the state.
The parser is then the only source of error, and the question is how a parse error on one event propagates to later decisions.
We treat the queried key $\kappa=(p,r)$ without leases and without masks (no group or window events that involve $p$ and $r$), so that the answer is the direct decision of $\kappa$.

\begin{definition}[Noisy parser]\thlabel{definition}{def:parser}
Each direct event on $\kappa$ is parsed independently of all others. With probability $\delta_0$ it is dropped (parsed as a null event), with probability $\delta_a$ its decision is flipped and its rank kept, with probability $\delta_\ell$ its rank is replaced by a uniformly random other rank and its decision kept, and with probability $\gamma:=1-\delta_0-\delta_a-\delta_\ell$ it is parsed correctly. All other events are parsed correctly and none is attributed to $\kappa$. The exact rule of Definition~\ref{def:process} is applied to the parsed stream.
\end{definition}

Let $e_1,\dots,e_N$, with $N\ge1$, be the true events on $\kappa$ in time order, with ranks $\ell_j$ and decisions $a_j$.
By Lemma~\ref{lem:scalar} the true answer is the decision $a^\ast$ of the event $e^\ast=e_{J^\ast}$ that maximizes (rank, time), with $\ell^\ast$ its rank. The parsed stream is also lease free, so the parsed answer is the decision of the parsed event that maximizes (parsed rank, time), or \textsf{DENY} if no event survives.
Let $\mathrm{Err}$ be the probability that the parsed answer differs from $a^\ast$, and let $N_{\mathrm{after}}=N-J^\ast$ and $N_{\mathrm{before}}=J^\ast-1$ be the numbers of events on $\kappa$ after and before $e^\ast$.

\begin{proposition}[Propagation of parse errors]\thlabel{proposition}{prop:parser}
\begin{enumerate}
\item[(a)] \emph{One rank.} If $L=1$ (so $\delta_\ell=0$), every event is applied and
\[
\mathrm{Err}=\sum_{k=0}^{N-1}\delta_0^{\,k}\Bigl[\delta_a\,\ind\{a_{N-k}=a_N\}+(1-\delta_0-\delta_a)\,\ind\{a_{N-k}\ne a_N\}\Bigr]+\delta_0^{\,N}\,\ind\{a_N=1\}.
\]
A parse error on $e_j$ reaches the final answer only if all $N-j$ later events are dropped, which has probability $\delta_0^{\,N-j}$. For an alternating chain, $\mathrm{Err}\to\bigl(\delta_a+\delta_0(1-\delta_0-\delta_a)\bigr)/(1-\delta_0^2)$ as $N\to\infty$. For a run of equal decisions, $\mathrm{Err}=\delta_a(1-\delta_0^{\,N})/(1-\delta_0)+\delta_0^{\,N}\ind\{a_N=1\}$.
\item[(b)] \emph{Ranks.} For any $L\ge2$, with $p_<=\delta_\ell(L-\ell^\ast+1)/(L-1)$ and $p_>=\delta_\ell(L-\ell^\ast)/(L-1)$,
\[
\mathrm{Err}\ \le\ 1-\gamma\,(1-p_<)^{N_{\mathrm{after}}}(1-p_>)^{N_{\mathrm{before}}}.
\]
If $N_{\mathrm{before}}=0$ and every later event has decision $1-a^\ast$, then $\mathrm{Err}\ge\gamma\bigl(1-(1-p_<)^{N_{\mathrm{after}}}\bigr)$.
\end{enumerate}
\end{proposition}

\begin{proof}
(a) With one rank every parsed event is applied, so the parsed answer is the parsed decision of the last event that is not dropped, or \textsf{DENY} if all are dropped. Event $e_{N-k}$ is the last survivor with probability $\delta_0^{\,k}(1-\delta_0)$, and given that it survives its decision is flipped with probability $\delta_a/(1-\delta_0)$. The answer is wrong if the survivor has the decision $a_N$ and is flipped, or has the other decision and is not flipped, which gives the two terms. All $N$ events are dropped with probability $\delta_0^{\,N}$, and then the answer is wrong exactly when $a_N=1$. The two closed forms are geometric sums.

(b) A later event has rank below $\ell^\ast$, by maximality of $e^\ast$. It beats $e^\ast$ in the parsed order exactly when its parsed rank is at least $\ell^\ast$, which requires a rank error landing on one of the $L-\ell^\ast+1$ ranks $\ge\ell^\ast$, none of which is its own rank. This has probability $p_<$. An earlier event has rank at most $\ell^\ast$ and beats $e^\ast$ exactly when its parsed rank exceeds $\ell^\ast$, with probability $p_>$. Decision flips and drops of other events never make them beat $e^\ast$.
Let $A$ be the event that $e^\ast$ is parsed correctly and no other event beats it. By independence $\Prob(A)=\gamma(1-p_<)^{N_{\mathrm{after}}}(1-p_>)^{N_{\mathrm{before}}}$, and on $A$ the parsed maximizer is $e^\ast$, so the answer is right. This is the upper bound.
For the lower bound, suppose $e^\ast$ is parsed correctly and some later event beats it. The parsed maximizer is then one of the later events that beat it. A rank error keeps the decision, so its decision is $1-a^\ast$, and the answer is wrong. With $N_{\mathrm{before}}=0$ these events have probability $\gamma\bigl(1-(1-p_<)^{N_{\mathrm{after}}}\bigr)$.
\end{proof}

The two parts behave differently.
In (a) errors are forgotten: the final answer depends only on the last event that survives parsing, every applied event renews the state, and the error stays bounded as the chain grows.
In (b) an upward rank error on a blocked event persists, because no later event on the key ranks high enough to replace it. Each of the $N_{\mathrm{after}}$ blocked events is a fresh chance to override the decisive one, and the error grows like $N_{\mathrm{after}}\,\delta_\ell(L-\ell^\ast+1)/(L-1)$ until it saturates. A downward rank error on the decisive event also persists: another event, earlier or later, becomes the parsed maximizer, and nothing later restores the decisive one.
So the parser must be accurate on ranks in proportion to the number of blocked events, while drops and decision flips on the queried key cost only through the last few events.
A Monte Carlo check on the exact rule (\texttt{theory/parser\_check.py}, $\delta_0=0.1$, $\delta_a=0.05$) confirms both parts. With one rank ($L=1$, $\delta_\ell=0$) the error is $0.136$ for alternating chains of $3$, $9$ and $33$ events, as formula (a) gives. With $L=3$ and $\delta_\ell=0.06$, a top-rank decisive event followed by $2$, $8$ and $32$ blocked events gives $0.24$, $0.37$ and $0.70$, just below the upper bound of (b), because in that family a mis-parsed decisive event is itself almost always an error. Under the same $L=3$ parameters an alternating chain at the top rank has error near $0.18$ for every length (upper bound $1-\gamma=0.21$), while a chain at the lowest rank grows from $0.17$ to $0.44$ between $3$ and $33$ events, because each earlier event can now be misread upward and block the later ones.

\begin{remark}[Attribution errors and leases]\thlabel{remark}{rem:parser}
Suppose in addition that each direct event on another key is attributed to $\kappa$ with probability $\mu$, keeping its rank and decision, and that $L=1$. If $n_{\mathrm{after}}$ events on other keys follow $e_N$ and their decisions are independent and uniform, the last misattributed event after $e_N$, if there is one, decides the answer and is wrong with probability $\frac12$, so $\mathrm{Err}\ge\frac12\bigl(1-(1-\mu)^{n_{\mathrm{after}}}\bigr)$. Like blocked-event rank errors, misattributions accumulate with the number of other-key events after the last update.
This mirrors the per-line costs of Section~\ref{sec:theory-reader}, where events on other keys break the binding of a full-log reader while further events on the queried key do not. The parallel is an interpretation, not a test, because a full-log reader is not a parser feeding an exact filter.
With leases, a misread expiry does not wear off at the next applied event as the errors of part (a) do. If a top-rank grant with true expiry $10$ is read as valid through $20$, the two decisions differ at times $11$ to $20$ unless an event applied in both streams intervenes. A rank-one grant at time $15$ is applied in the true stream and blocked in the parsed one, and the difference then lasts beyond time $20$, until the next event that is applied in both (\texttt{theory/boundary\_cases.py}, case C).
\end{remark}

\paragraph{Predicting a parser-plus-filter memory from measured rates.}
Proposition~\ref{prop:parser} gives closed forms for chain and blocked families. On general streams the prediction is a simulation from calibrated per-line rates assuming independent errors, which can be registered before the memory is run.
\emph{Rates.} On calibration streams disjoint from the test streams, classify every event line by the first applicable outcome: dropped (parsed as no event or unparseable, $\delta_0$), attributed to another component (another principal, resource or group, $\mu$ per line, which drops it for its own component and inserts it into the other), decision flipped with key and rank right ($\delta_a$), rank replaced with key and decision right ($\delta_\ell$), lease wrong with all else right ($\delta_\lambda$), other error, or correct. Classify filler lines as correct or inserted ($\iota$ per filler line). Estimate each rate separately for each true event type.
\emph{Closed form.} On single-rank chains (Definition~\ref{def:chain} with $L=1$), the predicted filter error is formula (a) of Proposition~\ref{prop:parser} with the queried key's $\delta_0$ and $\delta_a$. On blocked families it is the interval of (b).
\emph{General streams.} With backgrounds, groups, windows and leases, the prediction is a Monte Carlo over the test streams. Replace each line independently by an outcome drawn from its type's calibrated outcome frequencies, with a substituted field drawn uniformly from the alternatives seen in calibration. Apply the exact rule, and record the balanced error $\bar P^{\mathrm{parse}}_e$ of the exact answer of the resulting state.
\emph{Reader level.} For an LLM reader of that state, the reader model of Proposition~\ref{prop:budget} gives $\mathrm{BA}-\frac12=(\mathrm{BA}_{\mathrm{state}}-\frac12)(1-2\bar P^{\mathrm{parse}}_e)$, where $\mathrm{BA}_{\mathrm{state}}$ is the same reader's balanced accuracy on the exact state at the same checkpoint.
The main risk to the prediction is the independence in Definition~\ref{def:parser}. A parser that reads one line at a time with greedy decoding errs deterministically on a given text, so independence across lines holds only to the extent that wording and tickets vary.

\paragraph{When the plug-in filter is the Bayes decision.}
The plug-in filter applies the exact rule to the parsed events. The Bayes filter instead propagates a posterior over each component's state, with the probability of each possible true event given the parsed line as emission, a forward recursion over the supersession automaton.

\begin{proposition}[Plug-in equals MAP in the one-rank drop and flip model]\thlabel{proposition}{prop:plugin-map}
Under Definition~\ref{def:parser} with $L=1$, suppose the decisions of the true events on $\kappa$ are i.i.d.\ uniform and independent of the number and positions of these events and of the parse outcomes, and that $q:=\delta_a/(1-\delta_0)<\frac12$. Then for every parsed stream the MAP answer equals the plug-in answer, with ties broken toward the last parsed decision on $\kappa$, or toward \textsf{DENY} if none is parsed.
A tie with a parsed event requires the filter to be certain that a dropped event on $\kappa$ follows the last parsed one. This cannot happen when the number and positions of $\kappa$'s events are unknown to the filter, so that every later slot may hold a line that changes nothing.
\end{proposition}

\begin{proof}
If some event on $\kappa$ is parsed, let $\hat a$ be the parsed decision of the last one. Either it is the last true event, and then the answer equals $\hat a$ with probability $1-q$, or a dropped true event follows it, and then the answer is that event's decision, which is uniform and independent of the observation. Let $\rho$ be the posterior probability of the second case. Then $\Prob(\text{answer}=\hat a\mid\text{parse})=(1-\rho)(1-q)+\rho/2=\frac12+(1-\rho)(\frac12-q)$. This exceeds $\frac12$ when $\rho<1$, and when $\rho=1$ it equals $\frac12$ and the tie-break picks $\hat a$.
If no event on $\kappa$ is parsed, the answer is \textsf{APPROVE} only if some event was dropped and the last dropped decision is a grant, which has posterior probability at most $\frac12$. With ties broken toward \textsf{DENY}, MAP answers \textsf{DENY}, as the plug-in filter does.
\end{proof}

With the slots of $\kappa$'s events known, as in Definition~\ref{def:chain}, a dropped last event gives $\rho=1$ and an exact tie. With unknown slots, MAP and plug-in agree on every parsed stream of a small model with four slots, without ties (\texttt{theory/boundary\_cases.py}, case D).

Plug-in and MAP part where the parse channel or the prior is structured.
With ranks, a late event parsed at the top rank overrides the decisive one in the plug-in filter (Proposition~\ref{prop:parser}(b)). Under a prior in which top-rank events are rare, the posterior that such a line is a misread lower-rank event can exceed one half, and MAP keeps the earlier decision.
Systematic confusions matter more in practice. A parser that always reads an opened window as a window requirement, or attaches the ticket number as a lease, leaves a fixed signature that a calibrated correction can invert, and the plug-in filter cannot. The correction need not be sequential, as the audit below shows.

\paragraph{A calibrated Bayes filter.}
We fit the emissions by counting on calibration streams only. For a component $c$ (a direct record, membership, group grant or window), the probability of each true effect is estimated given four things: the parsed type, where $c$'s identifiers appear among the parsed fields, the parsed source and lease class for direct records, and a coarse summary of $c$'s current state. The last term makes the prior a Markov chain on the component.
The estimates are tied across components of the same kind and smoothed hierarchically. The filter propagates each relevant component exactly and decides by MAP on the product of the component marginals (\texttt{theory/bayes\_filter.py}).
On the test streams, as registered, it lowers the balanced error of the 0.5B parser from $0.31$ to $0.04$ at $64$ events and from $0.24$ to $0.09$ at $256$, and of the 1.5B parser from $0.45$ to $0.04$ and from $0.33$ to $0.09$, and it leaves the 3B, 7B and 14B parsers unchanged.

\paragraph{What does the work.}
An audit registered after that result was known, and therefore diagnostic (\texttt{theory/REGISTRATION\_round5.md}, \texttt{theory/bayes\_audit.py}), asks where the gain comes from, on the same test streams.
\emph{(i) Query blindness.} No component's recursion involves the query. A filter that keeps the posteriors of all $68$ component records online, with no query in sight, makes the same decisions on every stream, parser and checkpoint. It stores $210$ to $500$ support points on average, $8$ to $21$ kbit at $32$-bit precision.
\emph{(ii) Factorization.} An exact joint forward filter over the $10$ query-relevant components, with a joint emission over the true event behind each line, decides as the product of its own marginals on every test episode. For the 0.5B, 1.5B and 3B parsers it also has the balanced error of the registered filter. The product of marginals costs nothing here.
\emph{(iii) Remapping.} A memoryless remap replaces each parsed line by the most frequent true tuple seen for its parsed form on the calibration lines, and then applies the exact rule. It has exactly the balanced error of the Bayes filter in all ten parser and checkpoint cells, and under every family weighting of (v). So on these parsers sequential inference never beats remapping, and the gain is calibrated confusion correction.
\emph{(iv) Held-out wordings.} With leave-template-out calibration, the gain of the 0.5B parser at $256$ events disappears ($0.26$ against $0.24$ for the plug-in filter, interval of the difference $[-0.09,+0.14]$), so there it rests on wordings seen in calibration. The other three cells keep more than half of their gain. In one cell the Bayes filter then beats the remap (0.5B at $64$ events, $0.04$ against $0.19$, interval $[-0.26,-0.03]$). Its emissions back off to coarser parsed features where the remap's lookup table has no entry, and this design does not separate that smoothing from the sequential state.
\emph{(v) Family prior.} Reweighting the seven families of the test streams, uniformly or with one family tripled, keeps the gain over the plug-in filter for the 0.5B and 1.5B parsers, with upper $95\%$ limits below $-0.05$ under every weighting.
The residual error, $0.04$ and $0.09$, is the same for the 0.5B, 1.5B and 3B filters and lies entirely in the window family. The 3B parser reads every opened window as a window requirement, the smaller ones do so for part of them, and neither corrector separates the two readings.
The same calibration data also bear on independence. Errors are largely fixed by the wording template, with intraclass correlations within templates of $0.7$ or more for several event types, so Definition~\ref{def:parser}'s independence across lines fails at the template level.

\subsection{What the state costs}\label{sec:theory-rate}

This section states what a memory must spend to hold the state. We start with the unit that binds for a language-model reader, decodable components, and the reader model built on it.
The classical bit-level statements follow in compressed form, with proofs in Appendix~\ref{app:theory-proofs}. Their thresholds, $110$ to $281$ bits on the benchmark, lie below the smallest budget we test, so they bound no experiment.

\begin{remark}[Bits and tokens]\thlabel{remark}{rem:tokens}
The converses count bits. A memory of $B$ tokens over a vocabulary $\mathcal{V}$ holds at most $(B+1)\log_2|\mathcal{V}|$ bits (Definition~\ref{def:memory}), about $17$ bits per token for a vocabulary of $1.5\times10^5$, so the $110$ to $281$ bits of Corollary~\ref{cor:v2} fit in $7$ to $17$ tokens.
Budgets of $64$ to $512$ tokens are therefore far above the information-theoretic threshold, and the converses say nothing about them.
What a reader needs in practice is set by the encoding it can decode. If the cheapest rendering a reader decodes reliably costs $c_{\mathrm{comp}}$ tokens per state component, a full state memory needs about $K_{\mathrm{tot}}\,c_{\mathrm{comp}}$ tokens, and the ratio $c_{\mathrm{comp}}\log_2|\mathcal{V}|$ to the information per component, at most $\log_2|\cD|$ bits, measures the redundancy the reader requires.
The two natural-language renderings we tested, with one line per component, cost about twenty tokens per component. That is a measured cost of these renderings, not the cheapest code a reader could decode. They are redundant by two orders of magnitude, and the converses bound neither that redundancy nor a reader's ability to cope with less.
Three quantities must be kept apart: the size of the minimal state in bits (Section~\ref{sec:theory-minimal}), the length of a stored representation in tokens per component, which we measure only for the renderings tested, and the input cost of a reader, which also counts the rest of its prompt.
For the benchmark the most compact rendering the reader still decodes costs \XeightCost{} per component (experiment X8).
\end{remark}

\subsubsection{In units of decodable components}\label{sec:theory-components}

The bit thresholds below lie two orders of magnitude below any budget that a language-model reader can use (Remark~\ref{rem:tokens}). What binds in practice is the number of state components the reader can decode. If a reader needs $c$ tokens to decode one rendered component (about twenty for the two renderings tested), a budget of $B$ tokens holds $b=\lfloor B/c\rfloor$ components, and the memory can carry a fraction $f=\min\{1,b/|S|\}$ of a state with $|S|$ rendered lines.
In these units Theorem~\ref{thm:rd} and Proposition~\ref{prop:timeshare} (Section~\ref{sec:theory-rd}) read as follows. A memory that stores whole components, $b$ of the $K$ keys in the direct-only uniform model, is time sharing with $K'=b$, so its key error is $(1-b/K)(1-1/n)$. A memory that may pack information across components is still bounded by~\eqref{eq:rd} with $R=b\log_2 n$, that is $\bar\delta\ge\varphi^{-1}\bigl((1-b/K)\log_2 n\bigr)$.
The following proposition turns the component count into a prediction for a reader.

\begin{definition}[Budgeted memory]\thlabel{definition}{def:budget}
A budgeted memory holds at most $k$ state components, each rendered as one line. Let $\hat y$ be the answer of the exact rule on the memory's contents, recursive or projected, with a missing component read as absent, the closed-world default of the rule. Let $p_e$ be the probability that $\hat y$ differs from the true answer $y$ of the episode, and $\bar P_e$ the balanced error, the mean of $p_e$ over \textsf{APPROVE} episodes and over \textsf{DENY} episodes, averaged.
A \emph{read-time projection} computes the exact filter state, renders it as $|S|$ lines (one per non-default component), and keeps $k=\lfloor f|S|+\frac12\rfloor$ of them, either a uniformly random $k$-subset (\emph{random eviction}) or the $k$ lines whose component was touched most recently (\emph{recency eviction}). We write $\bar P_e(f)$ for its balanced error. Its hidden state is unbounded, and only the reader's view is bounded.
A bounded recursive store (Definition~\ref{def:store}) has no hidden state.
\end{definition}

\begin{proposition}[A reader model for budgeted memories]\thlabel{proposition}{prop:budget}
Suppose the reader's decision depends on the memory's contents only through $\hat y$, and that given $\hat y$ it outputs $\hat y$ with probability $a_{\hat y}$, the class-conditional accuracy it has on full states. Then its balanced accuracy at fraction $f$ (for a store, at the capacity $C$ that $f$ sets) is
\[
\mathrm{BA}(f)-\tfrac12=\bigl(\mathrm{BA}(1)-\tfrac12\bigr)\bigl(1-2\bar P_e(f)\bigr),\qquad \mathrm{BA}(1)=\tfrac12(a_{\textsf{A}}+a_{\textsf{D}}).
\]
For a read-time projection, if every \textsf{APPROVE} episode rests on one line and no \textsf{DENY} episode can be turned into \textsf{APPROVE} by removing lines, random eviction gives $\bar P_e(f)=(1-f)/2$ up to rounding of $k$, hence $\mathrm{BA}(f)-\frac12=f\bigl(\mathrm{BA}(1)-\frac12\bigr)$.
\end{proposition}

\begin{proof}
For an episode with answer $y$ the reader is right with probability $(1-p_e)a_y+p_e(1-a_{1-y})$, since a wrong $\hat y$ equals $1-y$. Let $e_y$ be the mean of $p_e$ over episodes with answer $y$. Averaging over the two classes gives $\mathrm{BA}(f)=\frac12\bigl[(1-e_{\textsf{A}})a_{\textsf{A}}+e_{\textsf{A}}(1-a_{\textsf{D}})+(1-e_{\textsf{D}})a_{\textsf{D}}+e_{\textsf{D}}(1-a_{\textsf{A}})\bigr]$, and collecting terms gives $\mathrm{BA}(f)-\frac12=\frac12(a_{\textsf{A}}+a_{\textsf{D}}-1)(1-e_{\textsf{A}}-e_{\textsf{D}})$. For the special case, the deciding line of an \textsf{APPROVE} episode survives random eviction with probability $k/|S|\approx f$, and \textsf{DENY} episodes have $p_e=0$.
\end{proof}

The factor $1-2\bar P_e(f)$ depends only on the generator and the eviction rule, so it can be computed before any reader is run (\texttt{theory/budget\_predict.py}). Table~\ref{tab:budget} gives it on the controlled backgrounds of the rate sweep ($40$ pairs per cell, the sweep's seeds). Random eviction is close to the line $f$, except where $|S|\approx4$ makes the rounding of $k$ coarse. Recency eviction is far below it, because the queried key's lines are often old.
The identity of Proposition~\ref{prop:budget} holds for any $\hat y$, so it applies to the bounded store below. Its special case does not (Proposition~\ref{prop:store}(c)).

\begin{table}[t]
\centering
\caption{Predicted factor $1-2\bar P_e(f)$ of Proposition~\ref{prop:budget} on the controlled rate-sweep backgrounds ($K$ background keys, horizon $T$, $40$ pairs). Predicted balanced accuracy is $\frac12+(\mathrm{BA}(1)-\frac12)$ times the entry.}\label{tab:budget}
\resizebox{\textwidth}{!}{%
\begin{tabular}{llrrrrrrrrrrr}
\hline
eviction & $K$ & $T$ & $\overline{|S|}$ & $f=0$ & $0.125$ & $0.25$ & $0.375$ & $0.5$ & $0.625$ & $0.75$ & $0.875$ & $1$\\
\hline
random & 3 & 64 & 4.3 & .000 & .161 & .161 & .355 & .450 & .581 & .709 & .840 & 1\\
random & 3 & 256 & 4.2 & .000 & .194 & .194 & .410 & .455 & .649 & .716 & .910 & 1\\
random & 7 & 64 & 8.0 & .000 & .098 & .203 & .314 & .457 & .572 & .708 & .851 & 1\\
random & 7 & 256 & 8.2 & .000 & .101 & .208 & .319 & .468 & .594 & .724 & .860 & 1\\
random & 15 & 64 & 13.0 & .000 & .118 & .212 & .341 & .470 & .578 & .713 & .853 & 1\\
random & 15 & 256 & 16.2 & .000 & .084 & .178 & .282 & .417 & .548 & .688 & .839 & 1\\
recency & 3 & 64 & 4.3 & .000 & .050 & .050 & .075 & .075 & .225 & .225 & .525 & 1\\
recency & 3 & 256 & 4.2 & .000 & .050 & .050 & .075 & .075 & .175 & .175 & .775 & 1\\
recency & 7 & 64 & 8.0 & .000 & .075 & .175 & .225 & .325 & .450 & .700 & .775 & 1\\
recency & 7 & 256 & 8.2 & .000 & .000 & .025 & .050 & .075 & .200 & .325 & .350 & 1\\
recency & 15 & 64 & 13.0 & .000 & .075 & .225 & .275 & .425 & .450 & .575 & .725 & 1\\
recency & 15 & 256 & 16.2 & .000 & .025 & .075 & .075 & .100 & .125 & .175 & .375 & 1\\
\hline
\end{tabular}}
\end{table}

Proposition~\ref{prop:budget} is a reader model, not a theorem about readers: its hypothesis on the reader is what the budget experiment tests. That experiment is a read-time projection, so it tests the reader hypothesis and not a storage constraint. It fails if the reader guesses differently when a line is missing, or reads a partial state worse than a full one, beyond the closed-world default. A model-maintained ledger can be placed on the same scale through its effective fraction, the fraction of the true state's lines that its final memory states correctly.

\paragraph{A bounded recursive store.}
A read-time projection keeps the full state out of the reader's sight but still computes it. A memory that is genuinely rate limited must process every event against what it has kept.

\begin{definition}[Bounded recursive store]\thlabel{definition}{def:store}
A bounded recursive store of capacity $C$ holds at most $C$ component records and nothing else. Components are those of Definition~\ref{def:process}, with the window requirement and the window state of a resource held in one record.
Each event on a component $c$ is processed by the exact rule against the stored record of $c$, or against the default if $c$ is absent, because it was never written, was evicted or was deleted.
A removal from a group deletes the membership record. Every other event that is applied writes the record of $c$, the window record even when both of its flags are off. A write updates the record if $c$ is stored, and otherwise it is an \emph{insertion}. When the store already holds $C$ records, an insertion first evicts one record, the \emph{victim}, which is lost.
An event on a stored record \emph{touches} it, whether the event is applied or blocked, and so does an insertion.
Under \emph{random eviction} the victim is uniform over the stored records. Under \emph{LRU eviction} it is the record whose last touch is earliest, which is unique because each line touches at most one component.
The answer $\hat y$ is that of the exact rule on the stored records, as in Definition~\ref{def:budget}.
\end{definition}

\begin{proposition}[Retention in a bounded store]\thlabel{proposition}{prop:store}
Let a record $x$ be stored after time $s$, and let no event on $x$ occur in $(s,T]$.
\begin{enumerate}
\item[(a)] Under random eviction with $C\ge2$, $x$ is still stored at $T$ with probability
\[
\E_Q\bigl[(1-1/C)^{N}\bigr]\ \ge\ (1-1/C)^{\E_Q N},
\]
where $N$ is the number of evictions in $(s,T]$ and $Q$ is the law of the store in which $x$ is exempt and each victim is uniform over the other $C-1$ records. If the store is full at $s$, no record is deleted in $(s,T]$ and the events in $(s,T]$ fall on $N$ distinct components with no earlier event, one event each, then $N$ is fixed and the probability is $(1-1/C)^N$.
\item[(b)] Under LRU eviction, if no record is deleted in $(s,T]$, $x$ is still stored at $T$ if and only if fewer than $C$ distinct components other than $x$ are touched in $(s,T]$. Without the condition on deletions, the ``if'' direction holds.
\item[(c)] If $C$ is at least the largest number of records that the same store holds up to $T$ without a capacity limit, no eviction occurs and the store holds the exact state, so $\hat y=y$ under every eviction rule. Below that size the special case of Proposition~\ref{prop:budget} fails for $L\ge2$. A top-rank withdrawal can be evicted and a later rank-one grant then applied, turning a \textsf{DENY} episode into \textsf{APPROVE}, and a top-rank grant can be lost to a later rank-one withdrawal in the same way.
\end{enumerate}
\end{proposition}

\begin{proof}
(a) Fix the events. A path of the store is its sequence of victims. On a path on which $x$ is never chosen, each eviction has probability $1/C$ under the store and $1/(C-1)$ under $Q$, so the path has probability $\bigl((C-1)/C\bigr)^{N}$ times its probability under $Q$. These paths are exactly the paths of $Q$, and summing over them gives the identity. Jensen's inequality gives the bound. Under the stated condition every event in $(s,T]$ inserts a new component into a full store, so $N$ is the number of these events.
(b) Suppose an eviction at time $t\in(s,T]$, inserting $y$, chooses $x$. Since $x$ was last touched at $s$ and is the oldest record, each of the other $C-1$ stored records was touched after $s$, and with $y$ these are $C$ distinct components touched in $(s,t]$. So fewer than $C$ rules out the eviction of $x$. Conversely, suppose $x$ survives while $C$ distinct components other than $x$ are touched in $(s,T]$. A touch leaves its component stored, and while $x$ is stored no component touched after $s$ can be a victim, because $x$ is older. Without deletions all $C$ of them and $x$ are stored at $T$, which exceeds the capacity.
(c) Without a capacity limit every event is processed against the exact record of its component, since an absent record and the exact default agree, so the unbounded store holds the exact state, together with some records whose value is the default. While no eviction has occurred, the bounded store coincides with the unbounded one, so it holds at most $C$ records and no insertion meets a full store. By induction no eviction is ever needed. For the examples, let $C=2$, and let a top-rank withdrawal on $\kappa$ be followed by grants on two other keys and then by a rank-one grant on $\kappa$. LRU evicts the withdrawal at the second grant, and the rank-one grant is then applied to the default. So the store answers \textsf{APPROVE}, while the exact rule blocks the grant and answers \textsf{DENY}. Under random eviction the same happens with positive probability. Exchanging grant and withdrawal gives the second example.
\end{proof}

The script \texttt{theory/boundary\_cases.py} (case E) checks (a) to (c) by simulation. Part (b) agrees on all $2{,}868$ records of runs on the rate-sweep streams, and part (a) is within $0.001$ of Monte Carlo in two designs, one with $N$ fixed and one with evicted records returning.
The proposition gives the store a retention time that a read-time projection lacks. In a projection a line survives random eviction with probability $k/|S|$, whatever its age. In a store, a record written $t$ events before the query survives random eviction with probability about $(1-1/C)^{\lambda t}\approx e^{-\lambda t/C}$ when evictions occur at rate $\lambda$ per event, a retention time of about $C/\lambda$ events. It survives LRU eviction exactly while fewer than $C$ other components have been touched since its last touch.
Above the peak state size there is no retention limit. So a rate-limited recursive filter below its state size forgets with a time constant set by its capacity and its insertion rate, and above it does not forget. We call this time constant the \emph{retention time}, to keep it apart from the binding decay length of a full-log reader (Section~\ref{sec:theory-reader}).

Table~\ref{tab:store} gives the factor $1-2\bar P_e$ of the store and of a read-time projection with the same capacity, on fresh seeds of the rate-sweep design. These are the registered predictions of the reader experiment (\texttt{experiments/r5\_bounded.py}, \texttt{r5\_bounded\_pred.json}). The independent implementation \texttt{theory/bounded\_store.py} gives the same answer on every run we compared ($27{,}720$ runs with shared random seeds).
Under random eviction at $f=\frac12$ and $K=15$ the store's factor falls from $0.36$ at $T=64$ to $0.11$ at $T=256$ and $0.05$ at $T=1{,}024$, while the projection with the same capacity stays between $0.39$ and $0.45$.
LRU stays within $0.05$ of the recency projection with the same capacity in six of the seven cells, and within $0.075$ in the seventh ($K=7$, $T=256$). Recency eviction already keeps the recently touched components, which is what LRU keeps.
Evicting records equivalent to the default first (rank-one withdrawals and expired grants), and otherwise by LRU, never does worse than LRU here and gains at most $0.075$ ($K=15$, $T=64$).
At $f=1$ the entries are below one, because a capacity set by the mean state size is below the peak size of some episodes.

\begin{table}[t]
\centering
\caption{Factor $1-2\bar P_e$ of the bounded recursive store of Definition~\ref{def:store} and of a read-time projection that keeps the same number $C=\mathrm{round}(f\cdot\overline{|S|})$ of components, fixed per cell, on fresh seeds of the rate-sweep backgrounds ($40$ pairs per cell, $50$ runs of random eviction per episode). These are the registered predictions of the reader experiment.}\label{tab:store}
\resizebox{\textwidth}{!}{%
\begin{tabular}{llrrrrrrrrrrr}
\hline
eviction & $K$ & $T$ & $\overline{|S|}$ & $f=0$ & $0.125$ & $0.25$ & $0.375$ & $0.5$ & $0.625$ & $0.75$ & $0.875$ & $1$\\
\hline
store, random & 3 & 64 & 4.2 & .000 & .025 & .025 & .106 & .110 & .322 & .313 & .768 & .769\\
projection, random & 3 & 64 & 4.2 & .000 & .207 & .175 & .400 & .408 & .630 & .621 & .898 & .889\\
store, LRU & 3 & 64 & 4.2 & .000 & .025 & .025 & .050 & .050 & .225 & .225 & .675 & .675\\
projection, recency & 3 & 64 & 4.2 & .000 & .025 & .025 & .075 & .075 & .225 & .225 & .675 & .675\\
store, random & 3 & 256 & 4.2 & .000 & .000 & .000 & .028 & .029 & .120 & .118 & .678 & .685\\
projection, random & 3 & 256 & 4.2 & .000 & .187 & .177 & .390 & .387 & .589 & .613 & .869 & .859\\
store, LRU & 3 & 256 & 4.2 & .000 & .000 & .000 & .025 & .025 & .075 & .075 & .650 & .650\\
projection, recency & 3 & 256 & 4.2 & .000 & .000 & .000 & .025 & .025 & .100 & .100 & .650 & .650\\
store, random & 7 & 64 & 8.1 & .000 & .000 & .032 & .093 & .216 & .346 & .500 & .710 & .910\\
projection, random & 7 & 64 & 8.1 & .000 & .077 & .182 & .263 & .409 & .511 & .667 & .801 & .932\\
store, LRU & 7 & 64 & 8.1 & .000 & .000 & .000 & .100 & .175 & .275 & .425 & .675 & .875\\
projection, recency & 7 & 64 & 8.1 & .000 & .000 & .000 & .125 & .200 & .300 & .450 & .675 & .875\\
store, random & 7 & 256 & 8.3 & .000 & .000 & .025 & .054 & .086 & .137 & .218 & .379 & .756\\
projection, random & 7 & 256 & 8.3 & .000 & .089 & .180 & .308 & .417 & .510 & .650 & .776 & .923\\
store, LRU & 7 & 256 & 8.3 & .000 & .000 & .025 & .050 & .100 & .100 & .125 & .225 & .575\\
projection, recency & 7 & 256 & 8.3 & .000 & .025 & .050 & .075 & .150 & .150 & .200 & .275 & .575\\
store, random & 15 & 64 & 12.8 & .000 & .090 & .162 & .303 & .362 & .531 & .709 & .802 & .951\\
projection, random & 15 & 64 & 12.8 & .000 & .131 & .198 & .343 & .427 & .554 & .732 & .828 & .966\\
store, LRU & 15 & 64 & 12.8 & .000 & .075 & .200 & .350 & .425 & .525 & .725 & .875 & .975\\
projection, recency & 15 & 64 & 12.8 & .000 & .075 & .250 & .375 & .450 & .575 & .750 & .875 & .975\\
store, random & 15 & 256 & 16.3 & .000 & .025 & .039 & .070 & .110 & .205 & .364 & .566 & .886\\
projection, random & 15 & 256 & 16.3 & .000 & .081 & .173 & .267 & .387 & .526 & .675 & .805 & .953\\
store, LRU & 15 & 256 & 16.3 & .000 & .025 & .025 & .025 & .100 & .150 & .225 & .300 & .650\\
projection, recency & 15 & 256 & 16.3 & .000 & .025 & .025 & .025 & .100 & .150 & .275 & .325 & .650\\
store, random & 15 & 1024 & 16.1 & .000 & .000 & .008 & .028 & .053 & .091 & .153 & .310 & .899\\
projection, random & 15 & 1024 & 16.1 & .000 & .109 & .218 & .349 & .451 & .580 & .709 & .841 & .985\\
store, LRU & 15 & 1024 & 16.1 & .000 & .000 & .000 & .000 & .025 & .050 & .075 & .125 & .875\\
projection, recency & 15 & 1024 & 16.1 & .000 & .000 & .000 & .000 & .025 & .075 & .125 & .175 & .875\\
\hline
\end{tabular}}
\end{table}

\subsubsection{Minimal state and the zero-error converse}\label{sec:theory-minimal}

\begin{definition}[Filter state]\thlabel{definition}{def:filter}
Histories $h,h'\in\cO^*$ are \emph{equivalent}, $h\equiv h'$, if $f_q(S(hw))=f_q(S(h'w))$ for every continuation $w\in\cO^*$ and every $q\in\cQ$, where $S(h)$ is the state after $h$.
The \emph{filter state} $Z(h)$ is the equivalence class of $h$ and $\NQ$ is the number of classes.
For $t,k\ge0$, $\NQ(t,k)$ is the largest number of histories of length $t$ that are pairwise separated by some continuation of length at most $k$ and some query.
\end{definition}

Prepending $\varnothing$ does not change the class of a history, because leases are relative.
Hence $\NQ(t,k)$ is nondecreasing in both arguments and equals $\NQ$ once $t$ and $k$ are large enough, whenever $\NQ<\infty$.
Definition~\ref{def:filter} is the Myhill--Nerode congruence of the query family \cite{nerode1958linear}. It is also the causal-state equivalence of computational mechanics \cite{shalizi2001computational}, taken for an input--output process with deterministic outputs. The general object for such processes is the $\varepsilon$-transducer \cite{barnett2015computational}, and $Z$ is its causal state for the supersession automaton.

\begin{theorem}[Achievability]\label{thm:achieve}
The filter state obeys a recursion $Z(O_{1:t})=\Phi\bigl(Z(O_{1:t-1}),O_t\bigr)$, and every answer $f_q(S_t)$ is a function of $Z(O_{1:t})$.
If $\NQ<\infty$, the deterministic recursive memory $M_t=Z(O_{1:t})$ uses $\lceil\log_2\NQ\rceil$ bits and has $D_{\mathrm{wc}}(T)=0$ for every $T$.
\end{theorem}

\begin{theorem}[Converse]\label{thm:converse}
Let $t+k\le T$ and $N=\NQ(t,k)$. Fix histories $h_1,\dots,h_N\in\cO^t$ that are pairwise separated by continuations of length at most $k$, and a finite set $\Gamma$ of tests $(w,q)$ with $|w|\le k$ that separates every pair.
\begin{enumerate}
\item[(a)] Every deterministic recursive memory with $2^R<N$ has $D_{\mathrm{wc}}(T)=1$. At least $N-2^R$ of the $h_i$ are answered wrongly on some test.
\item[(b)] Every recursive memory with private randomness and $2^R<N$ has $D_{\mathrm{wc}}(T)>0$. Hence a zero-error recursive memory, randomized or not, needs at least $\log_2\NQ(t,k)$ bits.
\end{enumerate}
\end{theorem}

Theorems~\ref{thm:achieve} and~\ref{thm:converse} say that the filter state is the smallest zero-error recursive memory, randomized or not.
Its size does not involve the log length when leases are bounded.
These results are classical in substance. Theorem~\ref{thm:achieve} is the Myhill--Nerode construction of the minimal automaton \cite{nerode1958linear}, equivalently the causal-state construction \cite{shalizi2001computational,barnett2015computational}, Theorem~\ref{thm:converse}(a) is the fooling-set argument for deterministic space lower bounds, and (b) is a derandomization.
Part (b) is qualitative: no quantitative randomized worst-case bound is given. Quantitative statements for positive error are in Section~\ref{sec:theory-rd}. An earlier version of (b) bounded the randomized worst-case error by Fano's inequality, but that bound is at most about $1/\log_2 N$ and we dropped it. The tests $(w,q)$ are the tests of predictive state representations \cite{littman2001psr}.
The closest classical estimation setting is hypothesis testing with finite memory \cite{hellman1970learning}, where a finite-state machine is likewise updated recursively from a stream. There, with an unbounded i.i.d.\ sample and a fixed number of states, the probability of error stays bounded away from zero, at a level set by the number of states and the range of the likelihood ratio, and randomized transitions can be needed to approach the optimum. Here the observations are noiseless and a single event can change the answer, so a memory with $\NQ$ states is exact, no smaller memory can average its errors away, and for zero error randomization does not help (part (b)).
What is new in this section is not the general bound but its content for a supersession automaton, the explicit count of Proposition~\ref{prop:count} with ranks, leases, groups and windows.
The source is a finite automaton driven by noiseless observations and the memory is a noiseless store, so the threshold is the logarithm of a number of classes. There is no entropy production as in the data-rate theorem for unstable plants, and the analogy with that theorem is one of vocabulary.

\subsubsection{The state of the access process}\label{sec:theory-count}

Let $\rho=u-t\ge0$ be the residual lifetime of a live record after time $t$.
The \emph{reduced direct states} are
\begin{equation}\label{eq:reduced}
\cD=\{\bot\}\cup\{(0,\ell):2\le\ell\le L\}\cup\{(1,\ell,\rho):\ell\in[L],\ \rho\in\{1,\dots,\Lambda\}\cup\{\infty\}\}\cup\{(1,\ast,0)\},
\end{equation}
of size $L(\Lambda+2)+1$. Without leases only $\rho=\infty$ occurs and $|\cD|=2L$.

\begin{proposition}[Size of the filter state]\thlabel{proposition}{prop:count}
\begin{enumerate}
\item[(a)] Without leases,
\[
\NQ=(2L)^K\,2^{C_{\mathrm{aux}}},
\]
and $\NQ(t,k)=\NQ$ whenever $t\ge K_{\mathrm{tot}}$ and $k\ge|\mathcal{G}|+2$.
\item[(b)] With lease bound $\Lambda\ge K_{\mathrm{tot}}+|\mathcal{G}|+1$,
\begin{equation}\label{eq:count-lease}
K\log_2\bigl(L(\Lambda-K_{\mathrm{tot}}-|\mathcal{G}|+1)\bigr)+C_{\mathrm{aux}}\ \le\ \log_2\NQ(t,k)\ \le\ \log_2\NQ\ \le\ K\log_2\bigl(L(\Lambda+2)+1\bigr)+C_{\mathrm{aux}}
\end{equation}
for $t\ge K_{\mathrm{tot}}$ and $k\ge\Lambda+|\mathcal{G}|+2$.
In the direct-only submodel with a single key (no group or window events, and no group or window components) the count is exact, $\NQ=L(\Lambda+2)+1$.
\end{enumerate}
\end{proposition}

\begin{corollary}[Benchmark values]\thlabel{corollary}{cor:v2}
For the benchmark's dimensions ($K=24$, $C_{\mathrm{aux}}=48$, $L=3$), without leases $\log_2\NQ=24\log_2 6+48\approx110.0$ bits.
With leases up to $\Lambda=275$, as the generator draws at $T=256$, $269.5\le\log_2\NQ\le280.8$ bits, a sandwich gap of about $11$ bits.
The raw log prompt at $T=256$ measures about $7.0\times10^3$ tokens.
\end{corollary}

These counts are for the process of Definition~\ref{def:process}. The generator reaches a much smaller set of states, since it places leases only on the target key of the expiry family and touches few keys in short logs (Remark~\ref{rem:T}).

\begin{remark}[What the converses say]\thlabel{remark}{rem:graded}
Theorem~\ref{thm:converse} is a zero-error statement. Below $\log_2\NQ$ bits a deterministic memory answers some history and some test wrongly with certainty, and a randomized one with positive probability.
Theorem~\ref{thm:rd} quantifies the average under a uniform state of the direct-only process. The mean per-key error, the probability that some test in a battery of $2L-1$ tests about the key fails, is at least $\varphi^{-1}(\log_2 n-R/K)$. This is zero at $R/K=\log_2 n$, grows continuously as the rate falls, and reaches $1-1/n$ only as $R/K\to0$. The decision error for one random test from the battery is at least this divided by $2L-1$.
So the theory predicts graded degradation below the threshold, not a collapse. A sentence that states exactly this is: \emph{no recursive memory, randomized or not, with fewer than $\log_2\NQ$ bits answers every query correctly on every history, and below that size the mean per-key error is bounded from below by a converse curve that rises continuously from zero and reaches chance only as the memory vanishes.}
\end{remark}

\begin{remark}[Masking and leases]\thlabel{remark}{rem:mask}
With leases, the product of~\eqref{eq:reduced} overcounts when a direct record is hidden by group access.
If $p$ belongs to a group that may write to $r$, a grant expiring now and no record at all are equivalent.
The group path hides the difference now, unmasking takes at least one event, and by then both records are gone.
This is why part~(b) is stated as a sandwich. The gap per key is $\log_2\frac{L(\Lambda+2)+1}{L(\Lambda-K_{\mathrm{tot}}-|\mathcal{G}|+1)}\to0$ as $\Lambda\to\infty$.
\end{remark}

\begin{remark}[Where $T$ enters]\thlabel{remark}{rem:T}
For bounded leases $\NQ$ does not depend on $T$.
In the benchmark the expiry family draws leases that grow with $T$, and then $\log_2\NQ$ grows like $K\log_2 T$.
The raw log needs $\Theta(T)$ bits, but the relevant comparator for an average-case requirement is the entropy of the filter state under the event law, which is at most $\log_2\NQ$ and can be much smaller for sparse logs. For the benchmark generator, the sum of the entropies of the $72$ state components bounds $H(Z_T)$ from above by subadditivity. Plug-in estimates of that sum are $20.1$, $30.8$, $61.7$ and $78.2$ bits at $T=8$, $16$, $64$ and $256$ ($1000$ episodes per horizon, \texttt{theory/hz\_estimate.py}). Plug-in entropy is biased low, so these are not guaranteed upper bounds, and with the Miller--Madow correction the sums are $20.3$, $31.0$, $61.8$ and $78.4$ bits (bootstrap $95\%$ intervals $[19.8,20.4]$, $[30.5,31.1]$, $[61.3,62.0]$ and $[77.7,78.6]$). All lie far below $\log_2\NQ\ge110$. They grow with $T$ because the background touches more keys. If the true entropy follows them, the average-case requirement on this generator depends on the horizon through the number of active components.
The precise form of the thesis is therefore that the sufficient statistic is independent of $T$ when leases are bounded, and logarithmic in $T$ otherwise.
\end{remark}

\paragraph{Machine check.}
The script \texttt{theory/verify\_nq.py} tests these counts against the benchmark's own \texttt{replay()}.
The reduced filter of the proof agrees with \texttt{replay()} on $129{,}600$ queries over $300$ random histories of length $120$ with all event types and leases up to $12$.
Enumerating the reachable graph of exact reduced states and the bounded anytime answer signatures of one representative per state gives $96$ states with $96$ distinct signatures for one key and one group without leases ($=(2\cdot3)\cdot2^4$), and $11$ and $13$ for one key with $(L,\Lambda)=(2,3)$ and $(3,2)$ ($=L(\Lambda+2)+1$). The state count bounds $\NQ$ from above and the signature count from below.
The check supports the proof but does not replace it. In particular, pruning histories by bounded signature instead of by reduced state is unsound, since two states with equal short signatures can have different successors.
It also confirms the masking collapse of Remark~\ref{rem:mask}.

\subsubsection{Below the threshold}\label{sec:theory-rd}

Theorem~\ref{thm:converse} is a zero-error statement about the whole state.
A per-key version shows how the error grows when the rate falls below $\log_2\NQ$.
Consider histories with direct events only and no leases.
Every key then has $n:=2L$ reduced states, and every product state is reachable at time $K$.
For key $\kappa$ let $\Gamma_\kappa$ be the $2L-1$ tests of length at most one from the proof of Proposition~\ref{prop:count} (Appendix~\ref{app:theory-proofs}) (query now, withdraw or grant at rank $\ell'<L$ and query), which separate the $n$ states.
The $2L-1$ tests cannot all be run in one continuation. Let $E_\kappa$ be the event that some test in $\Gamma_\kappa$ is answered wrongly when each test is run on an independent copy of the memory started from the common state $M_t$ at time $t\ge K$ (horizon $T\ge t+1$).

\begin{theorem}[Per-key rate--distortion converse]\label{thm:rd}
Let the reduced key states $Z=(Z_\kappa)_{\kappa\in\cK}$ be uniform on $\cD^{\cK}$, $|\cD|=n$, realized by direct events.
Let $\varphi(\delta)=h(\delta)+\delta\log_2(n-1)$, with $h$ the binary entropy, and let $\varphi^{-1}$ be its inverse on $[0,1-1/n]$, where $\varphi$ increases from $0$ to $\log_2 n$.
For every recursive memory of $R$ bits, the mean key error $\bar\delta:=K^{-1}\sum_\kappa\Prob(E_\kappa)$ satisfies
\begin{equation}\label{eq:rd}
\bar\delta\ \ge\ \varphi^{-1}\Bigl(\max\Bigl\{0,\ \log_2 n-\frac{R}{K}\Bigr\}\Bigr),
\end{equation}
equivalently $R/K\ge\log_2 n-\varphi(\bar\delta)$ whenever $\bar\delta\le1-1/n$. For a deterministic memory the worst-case fraction of keys in error, $\max_{h}K^{-1}\sum_\kappa\ind\{E_\kappa\}$, obeys the same lower bound, since it is at least $\bar\delta$.
\end{theorem}

At $R/K<\log_2(2L)$ the key error is bounded away from zero, and it tends to $1-1/n$ as $R/K\to0$.
The bound is a converse only. A recursive scheme gives an upper curve.

\begin{proposition}[Recursive time sharing over keys]\thlabel{proposition}{prop:timeshare}
Under the hypotheses of Theorem~\ref{thm:rd}, the recursive memory that tracks the reduced states of a fixed set of $K'=\min\{K,\lfloor R/\log_2 n\rfloor\}$ keys exactly and answers the tests of every other key as if its state were a fixed $z_0\in\cD$ uses at most $R$ bits and achieves $\bar\delta=(1-K'/K)(1-1/n)$.
\end{proposition}

The upper curve is linear in $R$ while the converse~\eqref{eq:rd} is convex. At $n=6$ and half the full rate, $R/K=\frac12\log_2 6$, time sharing gives $\bar\delta=0.42$ and the converse $0.23$.
Theorem~\ref{thm:rd} is the Hamming rate--distortion converse for a uniform product source, instantiated. Block codes over keys approach it as $K\to\infty$, but they are not obviously updatable one event at a time.
\emph{Open question.} Does any recursive memory beat time sharing on this source, and does recursion cost anything relative to block coding? We do not know, and we therefore do not call~\eqref{eq:rd} a rate--distortion function.
The bounded stores of Proposition~\ref{prop:store} are other recursive schemes, and they can be placed on the same axis.

\begin{proposition}[Eviction stores on the uniform source]\thlabel{proposition}{prop:store-uniform}
Under the hypotheses of Theorem~\ref{thm:rd}, realize $Z$ by one direct event on each key whose state is not $\bot$, in uniformly random order. A bounded recursive store of $C$ records that evicts only when full, by any rule, achieves $\bar\delta=\E(M-C)_+/K$, where $M\sim\mathrm{Bin}(K,1-1/n)$ is the number of keys not in state $\bot$. Under random eviction it uses $R_C=\log_2\sum_{j=0}^{C}\binom{K}{j}(n-1)^j$ bits.
\end{proposition}

\begin{proof}
Each key receives at most one event, so no evicted record returns, and every insertion into a full store loses one key not in state $\bot$. The store ends with $\min\{M,C\}$ exact records. The other $(M-C)_+$ such keys are answered as $\bot$, which fails some test of $\Gamma_\kappa$, and keys in state $\bot$ are answered correctly. The memory is a set of at most $C$ pairs of a key and a non-default value, and random eviction needs no other state.
\end{proof}

Time sharing is the store with $C$ slots pinned to fixed keys (Proposition~\ref{prop:timeshare} with $K'=C$). It spends $\log_2 n$ bits per tracked key and stores the default too. An eviction store stores no default, but it spends about $\log_2 K+\log_2(n-1)$ bits per record, because it must name the key.
For $K=24$ and $n=6$ the eviction store lies above time sharing at every rate up to $60.6$ of the $62.0$ bits of the full state, and below it only in the last bit and a half, where time sharing cannot use the remaining bits for one more key (\texttt{theory/bounded\_store.py}). At half the full rate, $R/K=\frac12\log_2 6$, the store fits $5$ records and reaches $\bar\delta=0.63$, against $0.42$ for time sharing and $0.23$ for the converse. LRU must also hold the recency order, so it is no cheaper.
Eviction is thus a recursive scheme whose rate goes mostly to addressing. It places the stores of Proposition~\ref{prop:store} on the rate axis of the converse, and it leaves the open question where it was.
If the test of a key is drawn uniformly from $\Gamma_\kappa$, the decision error is at least $\bar\delta/(2L-1)$.
Key error is the probability that some test in a battery of $2L-1$ fails, which is not a decision error for a random query.
For a one-shot memory computed from the whole log at time $T$ only the current decisions matter, and Corollary~\ref{cor:fano} gives the analogous bound with $n=2$. For a recursive memory the end-of-horizon criterion changes nothing (Proposition~\ref{prop:recursion}).

\begin{corollary}[End-of-horizon Fano bound]\thlabel{corollary}{cor:fano}
Let $A=(f_q(S_T))_{q\in\cQ}$ and let $\hat A$ be any decisions computed from an $R$-bit memory at time $T$, recursive or not, with mean error $\bar P_e=|\cQ|^{-1}\sum_q\Prob(\hat A_q\ne A_q)$.
Then $h(\bar P_e)\ge\bigl(H(A)-R\bigr)/|\cQ|$.
If the answers are independent and uniform, $h(\bar P_e)\ge1-R/K$.
\end{corollary}

\begin{proposition}[The price of recursion is the rank]\thlabel{proposition}{prop:recursion}\thlabel{proposition}{rem:rank}
Consider the process without leases and $T\ge K_{\mathrm{tot}}+|\mathcal{G}|+2$.
\begin{enumerate}
\item[(a)] A one-shot memory, computed from the whole history at time $T$ and asked only at time $T$, is exact with one state per distinct answer vector at time $T$. Without groups and windows this is $2^K$ states, so $K$ bits suffice.
\item[(b)] A recursive memory with one alphabet that answers every query correctly at time $T$ on every history of length $T$, randomized or not, needs at least $\NQ=(2L)^K2^{C_{\mathrm{aux}}}$ states, even when $T$ is known. Without groups and windows this is $K\log_2(2L)$ bits.
\end{enumerate}
\end{proposition}

\begin{proof}
(a) is immediate, since only the answers at $T$ are asked.
(b) Take the $\NQ$ pairwise-separated histories of length $t=K_{\mathrm{tot}}$ from Proposition~\ref{prop:count}(a), with separating tests of length at most $|\mathcal{G}|+2$. Without leases a null event changes no state, so every test continuation $w$ can be padded in front with null events to end exactly at $T$, and it still separates the same pair. The argument of Theorem~\ref{thm:converse}(a) at time $t$, with all tests ending at $T$, and its derandomization (b) give the claim.
\end{proof}

For $L=2$ and one key, a rank-2 grant and a rank-1 grant are separated by a rank-1 withdrawal at time $T$, and a rank-2 withdrawal and no record are separated by a rank-1 grant at time $T$.
The extra $\log_2 L$ bits per key record the authority behind each current decision. No query asks for them, yet without them a later lower-ranked event cannot be processed. They are the price of recursion, not of answering at every time.
This is also the information that the priority family tests. As an aside that the theory does not imply (a full-log reader is not a recursive memory), at $T=64$ the 7B reader gets the DENY version of that family right only $57\%$ of the time, against $86\%$ to $93\%$ for the DENY versions of the restore and order families.

\subsection{Binding decay of a full-log reader: a model}\label{sec:theory-reader}

The results above concern recursive memories.
A transformer reader with the full log in context is not one: it has random access to all $T$ lines, so Theorems~\ref{thm:converse} and~\ref{thm:rd} say nothing about it.
Its loss of the state with $T$ is an empirical fact.
This section proposes a model for it and derives consequences that the data can falsify. The fits are reported with the decay experiment.
Everything here is conditional on Assumption~\ref{ass:binding}.

Let $s$ be the reader's decision score (log-probability of \textsf{DENY} minus that of \textsf{APPROVE}) and $y\in\{\textsf{A},\textsf{D}\}$ the true answer.

\begin{assumption}[Binding failure]\label{ass:binding}
\begin{enumerate}
\item[(B1)] For each episode there is a latent indicator $\beta$ that the reader bound the queried key's events correctly, independent of $y$ and of where the decisive event sits. Given $\beta=1$, $s$ has a class-conditional law $P^c_y$ that does not depend on $T$. Given $\beta=0$, $s$ has a law $P^z$ that depends on neither $y$ nor $T$.
\item[(B2)] A context with $n_j$ lines of type $j$ keeps the binding with probability
\[
\sigma=\pi+(1-\pi)\exp\Bigl(-\sum_j\varepsilon_jn_j\Bigr).
\]
Each line of type $j$ breaks it independently with probability $1-e^{-\varepsilon_j}$, and a fraction $\pi$ of episodes is immune. With one line type and $\pi=0$ this is $\sigma=e^{-T/\tau}$, $\tau=1/\varepsilon$. We call $\tau$ the \emph{binding decay length}. It is a parameter of this model, not an operational horizon of a reader.
\end{enumerate}
\end{assumption}

(B2) is a hazard model: every line, before or after the decisive event, multiplies the binding probability by $e^{-\varepsilon_j}$. In channel terms this is a contraction of total variation by $e^{-\varepsilon_j}$ per line \cite{dobrushin1956central}, but nothing below uses a strong data-processing inequality.
What separates the model from a Markov chain of channels along the log is that the contraction is global, not ordered by distance.
The first version of this section had a single line type and no immune fraction. The data forced both extensions (Section~\ref{sec:theory-decay-fits}).

\begin{proposition}[Consequences of Assumption~\ref{ass:binding}]\thlabel{proposition}{prop:binding}
Let $\sigma$ be the binding probability of (B2). Then
\begin{enumerate}
\item[(i)] the distribution functions satisfy $F_{s\mid\textsf{A}}-F_{s\mid\textsf{D}}=\sigma\bigl(F^c_{\textsf{A}}-F^c_{\textsf{D}}\bigr)$ pointwise, so the balanced accuracy of the best threshold rule, $\mathrm{BA}^\star=\frac12\bigl(1+\sup_c[F_{s\mid\textsf{A}}(c)-F_{s\mid\textsf{D}}(c)]\bigr)$, obeys $\mathrm{BA}^\star-\frac12=\sigma\bigl(\mathrm{BA}^\star_c-\frac12\bigr)$, and the same scaling holds for total variation, $\TV(P_{s\mid\textsf{A}},P_{s\mid\textsf{D}})=\sigma\,\TV(P^c_{\textsf{A}},P^c_{\textsf{D}})$,
\item[(ii)] at any fixed threshold, the \textsf{DENY} rate is $q=q_z+(q_c-q_z)\sigma$,
\item[(iii)] if $P^c_{\textsf{D}}$ stochastically dominates $P^c_{\textsf{A}}$ and the laws are continuous, $\mathrm{AUC}-\frac12=\sigma^2\bigl(\mathrm{AUC}_c-\frac12\bigr)+\sigma(1-\sigma)\,\theta$ with $0\le\theta\le\TV(P^c_{\textsf{A}},P^c_{\textsf{D}})$,
\item[(iv)] at fixed composition, accuracy does not depend on the distance between the decisive event and the query.
\end{enumerate}
\end{proposition}

\begin{proof}
(i) $P_{s\mid y}=\sigma P^c_y+(1-\sigma)P^z$, so $P_{s\mid\textsf{A}}-P_{s\mid\textsf{D}}=\sigma(P^c_{\textsf{A}}-P^c_{\textsf{D}})$ as measures, and the same identity holds for their distribution functions. The rule ``\textsf{DENY} if $s>c$'' has balanced accuracy $\frac12\bigl(1+F_{s\mid\textsf{A}}(c)-F_{s\mid\textsf{D}}(c)\bigr)$, so the best threshold attains $\frac12\bigl(1+\sup_c[F_{s\mid\textsf{A}}(c)-F_{s\mid\textsf{D}}(c)]\bigr)$, a one-sided Kolmogorov distance that scales by $\sigma$. Over all decision rules the best balanced accuracy is $\frac12(1+\TV)$, and total variation scales by $\sigma$ as well.
(ii) The \textsf{DENY} rate is linear in the mixture weight.
(iii) Expand $\Prob(s_{\textsf{D}}>s_{\textsf{A}})$ over the four combinations of $\beta$ for two independent episodes. Both corrupted gives $\frac12$. Both clean gives $\mathrm{AUC}_c$. The mixed terms sum to $1+\theta$ with $\theta=\E_{Z\sim P^z}\bigl[\Prob(s^c_{\textsf{D}}>Z)-\Prob(s^c_{\textsf{A}}>Z)\bigr]$, which lies between $0$ and the Kolmogorov distance by dominance. Collecting terms gives the formula.
(iv) is (B1).
\end{proof}

With one line type and $\pi=0$, $\mathrm{AUC}-\frac12$ decays exponentially in $T$, asymptotically at rate $1/\tau$ when $\theta>0$ (the $\sigma(1-\sigma)\theta$ term dominates) and at rate $2/\tau$ when $\theta=0$. An immune fraction $\pi>0$ adds a floor, $\mathrm{BA}^\star\to\frac12+\pi(\mathrm{BA}^\star_c-\frac12)$.
Only (ii) involves the decision threshold.
Suppose the score may in addition undergo a strictly increasing transformation $g_{T,\mathrm{comp}}$ that depends on the horizon and on the log composition but is common to all episodes pooled at that horizon and composition (a criterion shift). Then (i), (iii) and (iv) are unchanged, because total variation, the best balanced accuracy and the AUC of the pooled scores are invariant under one strictly increasing map, while (ii) is no longer implied.
We call this weaker version (B1$'$). The data reject (ii) and so force it (Section~\ref{sec:theory-decay-fits}).
The invariance fails if the transformation differs between episodes, since an episode-specific map changes the ranking across episodes. A metric that tolerates more is the within-pair ordering rate $\Prob(s_{\textsf{D}}>s_{\textsf{A}})$, which is invariant under any strictly increasing map that is common to the two members of a pair, even if it differs from pair to pair.

\paragraph{Two alternatives.}
\emph{Chain contraction.} If the decisive information instead passed through the $d$ later lines as a Markov chain of channels, each with Dobrushin coefficient at most $\eta$, then $\TV\le\eta^d\TV_0$ \cite{dobrushin1956central}.
Accuracy would fall steeply with the distance $d$ at fixed $T$, and averaging over a uniformly placed decisive event gives only a $1/T$ tail.
Matching the drop of $\mathrm{BA}^\star-\frac12$ between $T=16$ and $T=64$ for the smaller reader requires $\eta\approx0.97$, which predicts that at $T=64$ decisive events within $8$ lines of the query are about four times more discriminable than those $32$ or more lines away.
\emph{Interference.} If the decision variable is a signal plus interference that accumulates over the context, the Gaussian discriminability obeys $d'(T)^{-2}=d_0'^{-2}+T/T_c$ with $\mathrm{AUC}=\Phi(d'/\sqrt2)$ \cite{green1966signal}, a power-law tail with no distance dependence.

The fits of this model to the measured decay curves are reported with the decay experiment (Section~\ref{sec:theory-decay-fits}).
}{\section{Theory}\label{sec:theory}\todo{theory from theory session}}

\section{An executable benchmark with paired counterfactuals}
\label{sec:benchmark}

\paragraph{Events and state.} A log is a sequence of $T$ numbered events. Access events grant, withdraw or reconfirm write access of a service to a resource, and a grant may carry an expiry event index. Each source has a rank (security office above access administration above operations console), and an access event changes the current decision only if its source ranks at least as high as the source of the decision it would replace. Membership events add services to groups or remove them, group grants give a group write access, and a resource may require an open change window. Remaining positions hold operational filler lines in the same register. The policy is stated to the model as five rules (\Cref{app:policy}). The ground truth for a query ``may $p$ write to $r$ after event $t$'' is computed by replaying the prefix through the policy.

\paragraph{Families and pairs.} Each pair contains an approve version and a deny version of one target key. Background events, their positions and all wording are shared, and only the target family's lines differ.
\begin{center}
\small
\begin{tabular}{lll}
\toprule
Family & Approve version & Deny version \\
\midrule
restore & grant, withdraw, reconfirm & grant, reconfirm, withdraw \\
order & withdraw, grant & grant, withdraw \\
expiry & grant valid past the request & grant valid until before the request \\
priority & low-rank withdraw, high-rank grant & high-rank withdraw, low-rank grant \\
decoy & alarm-worded withdraw of another service & the same line for the target service \\
implicit & removed from group, added back & added, then removed \\
conjunction & grant, window closed then opened & grant, window opened then closed \\
\bottomrule
\end{tabular}
\end{center}
The restore family is sampled with double weight, so that reinstatements make up a quarter of all episodes. Key events are placed uniformly over the timeline, lines have 10 to 17 words, and no line carries capitalized banners. The replay oracle is asserted against the label each family intends for every generated episode.

\paragraph{Scoring.} After the prompt the assistant turn starts with \texttt{DECISION:}. We score the full-sequence log-probabilities of the continuations `` APPROVE'' and `` DENY'' (two tokens each in the Qwen and phi-4 tokenizers, so the two continuations need no length normalization) with label logits in single precision, and report balanced accuracy at the sign of the margin and AUC of the margin. A four-token greedy generation agrees with the margin decision in 98.9 to 99.6 percent of cases.

\paragraph{Reader validity.} No memory method can be judged against a reader that fails with the full history. We therefore fixed a validity gate before running any memory method: at 64 events the reader must reach recall 0.90 on both classes and on the reinstatement subset. On a separate validation range both Qwen2.5 sizes failed this gate (balanced accuracy 0.55 and reinstatement recall 0.28 for 7B, 0.63 and 0.39 for 14B), which is how the long-context failure was first found. The registered consequence was to stop: no memory method would be scored against a reader that fails the gate. We changed course after this result. Memory comparisons in this paper use the reader on short memory contents, and the validity reference for that use is reading the replayed state (\Cref{sec:exp-state}), not reading the full log. The change was made after seeing the gate results and is part of the record in \Cref{app:prereg}. The state-reading, buildable-filter and paraphrase comparisons are repeated with phi-4, whose state reading shows no detectable dependence on the horizon. The separation experiment, the recursive-memory grid and the factorial design use Qwen readers only.

\paragraph{Relation to other state benchmarks.} \Cref{tab:benchmarks} compares design features with benchmarks that probe whether answers follow the current state of an evolving history. The closest is StateMemBench \citep{fan2026can}, which also generates histories from a symbolic program with a deterministic evaluator. The features that matter for the theory here are the paired order counterfactuals, which make the separation results testable, and the reader-validity check, which separates memory failures from reading failures.

\IfFileExists{related/benchmark_table.tex}{
\begin{table}[t]
  \centering
  \small
  \setlength{\tabcolsep}{3pt}
  \newcolumntype{C}{>{\centering\arraybackslash\hspace{0pt}}p{1.08cm}}
  \caption{Design features of benchmarks that test whether a model or memory system answers from the current state of an evolving history.}
  \label{tab:benchmarks}
  \begin{tabular}{@{}>{\raggedright\arraybackslash}p{3.5cm}CCCCCCCCCC@{}}
    \toprule
    & {\footnotesize Exe\-cut\-able ground truth}
    & {\footnotesize Super\-session or re\-vo\-ca\-tion}
    & {\footnotesize Rein\-state\-ment after re\-vo\-ca\-tion}
    & {\footnotesize Source priority}
    & {\footnotesize Expiry}
    & {\footnotesize Paired order counter\-factuals}
    & {\footnotesize Reader check with correct state}
    & {\footnotesize Query-un\-known writing}
    & {\footnotesize Horizon as con\-trolled variable}
    & {\footnotesize Real logs} \\
    \midrule
    StateMemBench~\cite{fan2026can}
      & \checkmark & \checkmark & $\times$ & $\times$ & -- & $\times$ & $\times$ & \checkmark & \checkmark & $\times$ \\
    STALE~\cite{chao2026stale}
      & $\times$ & \checkmark & $\times$ & $\times$ & $\times$ & $\times$ & $\times$ & \checkmark & $\times$ & $\times$ \\
    Memora~\cite{uddin2026recall}
      & -- & \checkmark & -- & $\times$ & -- & $\times$ & $\times$ & -- & \checkmark & $\times$ \\
    MemoryAgentBench~\cite{hu2026evaluating}
      & $\times$ & \checkmark & $\times$ & $\times$ & $\times$ & $\times$ & $\times$ & \checkmark & \checkmark & $\times$ \\
    MEMAUDIT~\cite{bhargava2026memaudit}
      & \checkmark & \checkmark & $\times$ & $\times$ & -- & $\times$ & -- & \checkmark & $\times$ & $\times$ \\
    OrgAccess~\cite{sanyal2026orgaccess}
      & $\times$ & $\times$ & $\times$ & -- & \checkmark & $\times$ & $\times$ & $\times$ & $\times$ & $\times$ \\
    LongMemEval~\cite{wu2025longmemeval}
      & $\times$ & \checkmark & $\times$ & $\times$ & $\times$ & $\times$ & -- & \checkmark & \checkmark & -- \\
    PI-LLM~\cite{wang2025unable}
      & \checkmark & \checkmark & $\times$ & $\times$ & $\times$ & -- & $\times$ & $\times$ & \checkmark & $\times$ \\
    \midrule
    This paper
      & \checkmark & \checkmark & \checkmark & \checkmark & \checkmark & \checkmark & \checkmark & \checkmark & \checkmark & $\times$ \\
    \bottomrule
  \end{tabular}

  \vspace{2pt}
  {\footnotesize \checkmark: described in the paper. $\times$: not part of the design the paper describes. --: the paper is ambiguous. Evidence for each cell is in Table~\ref{tab:benchmark-evidence}.\par}
\end{table}
}{}

\section{Experiments}
\label{sec:experiments}

\subsection{Horizon decay of long-context reading}
\label{sec:exp-decay}

\begin{figure}[t]
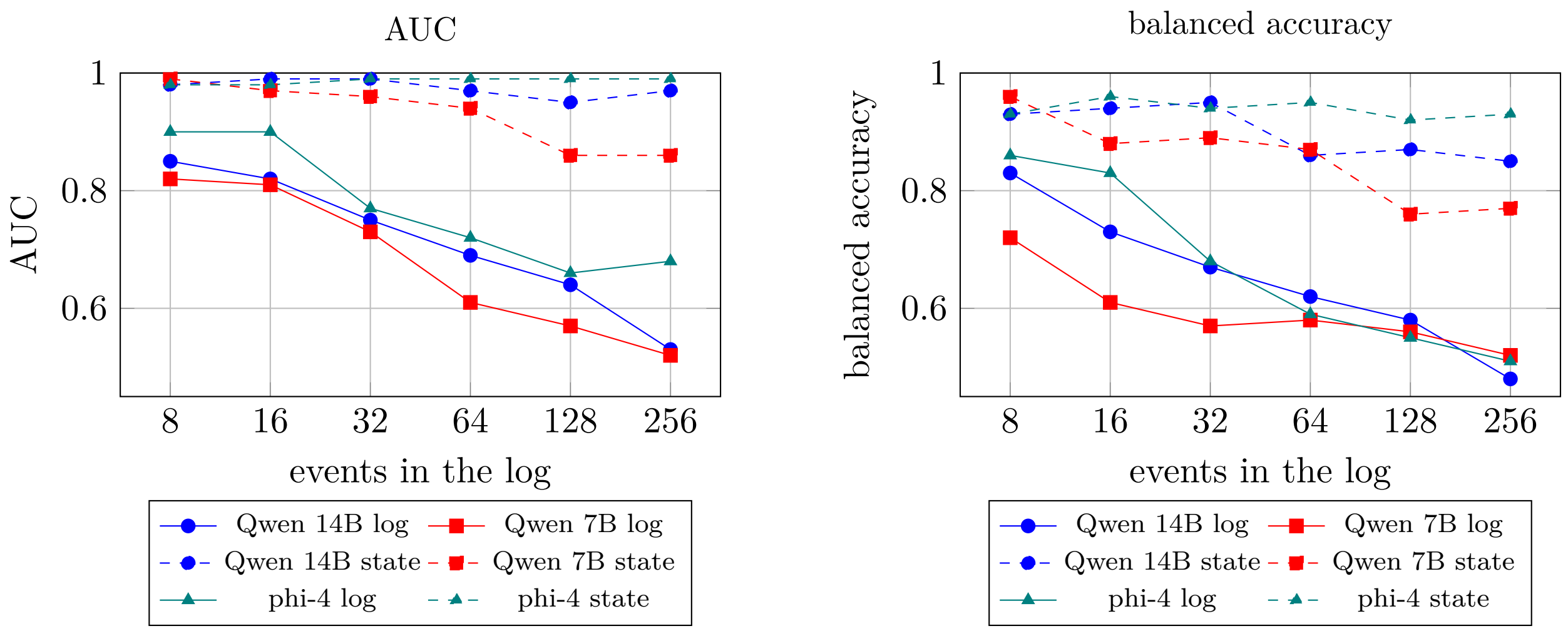

\centering
\begin{tikzpicture}
\begin{axis}[width=0.48\linewidth,height=5cm,xmode=log,log basis x=2,xtick={8,16,32,64,128,256},xticklabels={8,16,32,64,128,256},
  xlabel={events in the log},ylabel={AUC},ymin=0.45,ymax=1.0,legend style={font=\scriptsize,at={(0.5,-0.32)},anchor=north,legend columns=2},grid=major,title={\small AUC}]
\input{generated/x12_auc.tex}
\end{axis}
\end{tikzpicture}
\hfill
\begin{tikzpicture}
\begin{axis}[width=0.48\linewidth,height=5cm,xmode=log,log basis x=2,xtick={8,16,32,64,128,256},xticklabels={8,16,32,64,128,256},
  xlabel={events in the log},ylabel={balanced accuracy},ymin=0.45,ymax=1.0,legend style={font=\scriptsize,at={(0.5,-0.32)},anchor=north,legend columns=2},grid=major,title={\small balanced accuracy}]
\input{generated/x12_bal.tex}
\end{axis}
\end{tikzpicture}
\caption{Reading the full log (solid) against reading the replayed state (dashed). Each point is 120 episodes (60 pairs) from the test seed range. The state is a list of facts and never contains the answer.}
\label{fig:decay}
\end{figure}

Readers are Qwen2.5-7B-Instruct, Qwen2.5-14B-Instruct and phi-4 (14B, a different family and training corpus), all in bfloat16. For each horizon $T\in\{8,16,32,64,128,256\}$ we draw 60 pairs from the test seed range. \Cref{fig:decay} shows the result. AUC decreases with $T$ for all three readers (Spearman correlation $-1$, $-1$ and $-0.89$ over the six horizons). At 256 events the Qwen readers no longer separate the classes and phi-4 keeps an AUC of 0.68.

The share of deny decisions is higher at 256 events than at 8, 0.65 against 0.41 for Qwen 14B, 0.80 against 0.55 for Qwen 7B and 0.94 against 0.43 for phi-4, so part of the loss is a drift toward refusal. The drift does not explain the loss. With the decision threshold chosen after the fact to maximize balanced accuracy, the 7B reader on the validation range reaches only 0.64 at 64 events and 0.54 at 256 events. Replacing the persona ``automated decision and safety governance agent'' by ``access-decision assistant'' changes balanced accuracy at 64 events by $+0.008$ for all three readers.

\paragraph{Reasoning traces.} Asked to list the relevant events in order and apply the policy before answering, with up to 768 generated tokens, Qwen 14B goes from 0.62 to 0.78 at 64 events and from 0.48 to 0.54 at 256, and phi-4 from 0.59 to 0.88 and from 0.51 to 0.68. At 256 events 30 and 15 of 120 traces end without a decision. Reading the state without any trace remains better at both horizons, at 0.86 and 0.85 for Qwen 14B and 0.95 and 0.93 for phi-4. Counting only traces that reach a decision, the 768-token readers are at 0.73 and 0.77 at 256 events. With up to 4096 tokens, on the streams of \Cref{sec:exp-buildable}, 75 of 80 and 80 of 80 traces reach a decision and balanced accuracy at 256 events is 0.64 and 0.73, or 0.69 and 0.73 among decided traces, still below the state. A longer budget mostly turns missing decisions into answers of the same quality.

\paragraph{A larger reader.} Qwen2.5-32B-Instruct, run on 60 pairs at 16, 64 and 256 events, follows the same course. Reading the log, its AUC falls from 0.96 to 0.72 and 0.65 and its balanced accuracy from 0.86 to 0.61 and 0.55. Reading the state, its AUC is 0.99 or above at all three horizons. On the same pairs a model twice as large ranks better (AUC 0.96 against 0.82 at 16 events, 0.65 against 0.53 at 256) but loses about as much over the horizon (0.31 against 0.29).

\paragraph{Length more than same-key interference.} Reports on overwritten values attribute long-context failure to repeated updates of the same key \citep{wang2025unable}. A factorial design separates three causes on chain counterfactuals: events on the queried key ($m\in\{3,9,17\}$), events on other keys ($8$ or $40$) and total length ($T\in\{64,256\}$, filled with operational lines), with 40 pairs per cell. Averaged over the other factors, raising $m$ from 3 to 17 changes AUC by $+0.007$ for Qwen 14B and $-0.015$ for Qwen 7B, with 95 percent bootstrap intervals over pairs of $[-0.045,+0.059]$ and $[-0.068,+0.041]$. Five times more events on other keys lowers AUC by 0.07 and 0.13, and four times the length lowers it by 0.12 and 0.09. Within $m\le17$ the intervals bound the same-key effect at about 0.07 AUC, below the other two effects (0.07 to 0.13), and a bootstrap of the difference between effects excludes zero for both readers, so the claim is one of ordering and of an upper bound, not of absence. The 14B reader still errs on short views made only of the key's own events (\Cref{sec:exp-separation}).

\IfFileExists{theory/theory_decay.tex}{
\subsubsection{Fitting the reader model}\label{sec:theory-decay-fits}

\paragraph{Fits to the test data.}
Table~\ref{tab:decay} uses the six-length measurement on the test seeds (Qwen2.5-7B, Qwen2.5-14B and phi-4, raw log, $T\in\{8,16,32,64,128,256\}$, $60$ pairs per length).
Every law is fitted by least squares weighted with pair-bootstrap standard errors, and judged out of sample twice: leaving out one length at a time, and extrapolating from $T\le64$ to $T\in\{128,256\}$.
\emph{The two Qwen readers.} The exponential law has the smallest leave-one-out error ($0.028$ and $0.030$ AUC root mean square), and the interference and power laws have the largest ($0.042$ to $0.071$). Their in-sample $\chi^2$ on four degrees of freedom is $8.7$ to $12.1$, against $2.2$ and $2.4$ for the exponential.
Extrapolated from $T\le64$, the interference and power laws overpredict the $T=256$ AUC by $0.06$ to $0.09$, the signature of a tail that is too heavy. The exponential is within $0.02$ there, but it underpredicts $T=128$ for 14B by $0.06$.
The criterion of a held-out error below $0.05$ AUC is met by the exponential law in root mean square for both, and in the worst case for 7B ($0.047$) but not for 14B ($0.055$).
A floor parameter is not needed. For 14B it goes to its lower bound $\frac12$, and for 7B to $0.52$.
The fitted binding decay lengths are $\tau=69$ lines for 7B (pair-bootstrap interval $46$ to $108$) and $\tau=108$ for 14B ($76$ to $148$), with the same intercept $\mathrm{AUC}_c\approx0.87$, in agreement with the three-point fits on the separate validation seeds ($68.5$ and $125$).
\emph{phi-4.} Its AUC falls from $0.90$ to $0.66$ by $T=128$ and then stays there ($0.68$ at $T=256$).
The plain exponential fails ($\chi^2=16.9$, leave-one-out $0.080$, and it underpredicts $T=256$ by $0.15$ when extrapolated).
With an immune fraction (a floor, $\mathrm{AUC}=f+Ae^{-T/\tau}$) the fit is good in sample ($\chi^2=3.4$ on three degrees of freedom, floor $f=0.66$, $\tau=38$), but out of sample it ties with the interference and power laws (leave-one-out $0.050$ against $0.045$ and $0.046$), and no law meets the worst-case criterion (maximum errors $0.062$ to $0.093$).
Its floor is concentrated in families that local cues can solve: at $T\in\{128,256\}$ phi-4 orders all $15$ decoy pairs and $14$ of $15$ order pairs correctly, but only $61$ to $62\%$ of the restore and priority pairs.
\emph{Verdict.} The data are consistent with the model, not proof of it. With three readers the exponential law no longer wins everywhere. It is the best law for the two Qwen readers, and for phi-4 it needs the immune fraction of (B2), after which three laws are indistinguishable at six lengths.

\begin{table}[t]
\centering
\caption{Decay laws fitted to the six-length AUC (raw log, test seeds). LOO: leave one length out, root mean square and maximum absolute error. Extrap.: fitted on $T\le64$, error at $T=128$ and $T=256$. $\chi^2$ on four degrees of freedom (three for the floor law). $d'$-based laws use the equal-variance binormal link $\mathrm{AUC}=\Phi(d'/\sqrt2)$ \cite{green1966signal}.}\label{tab:decay}
\resizebox{\textwidth}{!}{%
\begin{tabular}{lccccccccc}
\hline
 & \multicolumn{3}{c}{Qwen2.5-7B} & \multicolumn{3}{c}{Qwen2.5-14B} & \multicolumn{3}{c}{phi-4}\\
Law & LOO & Extrap. & $\chi^2$ & LOO & Extrap. & $\chi^2$ & LOO & Extrap. & $\chi^2$\\
\hline
$\mathrm{AUC}-\frac12\propto e^{-T/\tau}$ & $.028/.047$ & $-.029,-.017$ & $2.2$ & $.030/.055$ & $-.063,-.010$ & $2.4$ & $.080/.138$ & $-.061,-.154$ & $16.9$\\
$\mathrm{AUC}=f+Ae^{-T/\tau}$ & $.032/.047$ & $-.029,-.017$ & $1.5$ & $.054/.108$ & $-.000,+.108$ & $2.4$ & $.050/.093$ & $-.019,-.066$ & $3.4$\\
$d'\propto e^{-T/\tau}$ & $.030/.052$ & $-.036,-.019$ & $2.6$ & $.035/.060$ & $-.073,-.016$ & $3.0$ & $.088/.148$ & $-.081,-.165$ & $20.2$\\
$d'^{-2}=d_0'^{-2}+T/T_c$ & $.048/.081$ & $+.043,+.059$ & $12.1$ & $.042/.076$ & $-.003,+.075$ & $11.0$ & $.046/.062$ & $+.013,-.051$ & $6.1$\\
$d'\propto T^{-\alpha}$ & $.071/.142$ & $+.039,+.057$ & $8.7$ & $.059/.089$ & $+.010,+.094$ & $10.9$ & $.045/.067$ & $+.018,-.041$ & $6.1$\\
\hline
\end{tabular}}
\end{table}

\paragraph{Distance.}
Recovering the decisive slot of every pair, the within-pair ordering rate $\Prob(s_{\textsf{D}}>s_{\textsf{A}})$ shows no decline with the distance between the decisive event and the query.
On the validation seeds at $T=64$ it is $0.63$, $0.63$ and $0.67$ for distances in $[0,8)$, $[8,32)$ and $[32,64)$ for 7B, and $0.76$, $0.74$ and $0.76$ for 14B, with standard errors near $0.07$.
On the test seeds, for all three readers, the correlation between distance and correct ordering lies between $-0.21$ and $+0.19$ at every length where ordering is not already perfect.
The fourfold advantage of near events predicted by chain contraction is absent. Since (iv) restates (B1), this tests the assumption directly rather than a derived consequence.

\paragraph{The criterion.}
Prediction (ii) fails for all three readers. With the binding probability fixed at the AUC fit, the \textsf{DENY} rate curve gives $\chi^2=25.2$, $17.2$ and $25.8$ (7B, 14B, phi-4) on four degrees of freedom.
Fitted freely, the bias saturates within $6$ to $8$ lines for the Qwen readers. phi-4 drifts more slowly but further, from $0.42$ to $0.94$.
The shift toward \textsf{DENY} is thus a separate effect from the loss of discriminability, which is why we keep only (B1$'$).

\paragraph{Which lines break the binding.}
\emph{Resolved state lines.} In the stated-memory arm the reader sees only the reduced state, one line per state component, and the number of lines grows with the number of keys the background touches.
For 14B and phi-4 the AUC stays between $0.95$ and $0.99$ from one to thirty-two state lines, while on raw logs 14B has a binding decay length of $108$ lines and phi-4 falls to its floor within about $128$ lines.
Their balanced accuracy at the default threshold drifts a little (14B from $0.94$ to $0.85$, with the \textsf{DENY} rate rising from $0.49$ to $0.63$), a criterion shift with discriminability intact, as in (B1$'$).
For 7B the AUC does fall, from $0.99$ to $0.86$.
So a resolved state line is at least ten times cheaper than a raw log line for the two stronger readers.
\emph{Raw lines.} The factorial chain experiment X1c varies, for the 7B and 14B readers, the number $m\in\{3,9,17\}$ of events on the queried key, the number $n_o\in\{8,40\}$ of events on other keys and the length $T\in\{64,256\}$, with filler lines making up the rest ($n_f=T-m-n_o$), on $40$ pairs per cell.
We fit three versions of (B2) to the twelve cell AUCs (\texttt{theory/x1c\_lines.py}, $\pi=0$).
Length alone ($\sigma=e^{-T/\tau}$) is rejected, with $\chi^2=59.3$ (7B) and $19.1$ (14B) on ten degrees of freedom, because at fixed $T$ more other-key events lower the AUC.
Separating event lines from filler fits well ($\chi^2=11.5$ and $7.8$ on nine). An event line costs $\varepsilon_{\mathrm{ev}}=0.018$ (7B) and $0.011$ (14B), four to nine times a filler line ($0.002$ and $0.003$).
Separating further the queried key from other keys lowers $\chi^2$ to $6.9$ and $3.4$ (one more parameter, AIC better by $2.6$ and $2.3$). It gives $\varepsilon_o=0.022\pm0.003$ and $0.012\pm0.003$ for other-key events, while queried-key events cost $0.006\pm0.007$ and $-0.002\pm0.007$, no more than filler.
The prediction of the previous version of this section, that events on the queried key would cost more than other lines, is refuted: adding fourteen queried-key events at fixed length changes the AUC by $+0.007$ (14B) and $-0.015$ (7B), with intervals containing zero.
What breaks the binding are unresolved events, and if anything those on other keys, which can be mistaken for the queried key. Same-key supersession chains do not add measurably to the cost.
Applied to the benchmark mix (about one third events, the rest filler), these per-line costs give binding decay lengths $1/\varepsilon_{\mathrm{mix}}$ of about $110$ (7B) and $170$ (14B) lines. The decay length of the AUC fitted directly lies between $1/(2\varepsilon)$ and $1/\varepsilon$ by Proposition~\ref{prop:binding}(iii), and the direct fits, $69$ and $108$ lines, fall in that range. The two designs differ in their event families, so this is a consistency check, not a prediction: the two estimates are consistent with the model.

\paragraph{Binding decay length.}
Under the model a reader is summarized by $(\mathrm{AUC}_c,\pi,\{\varepsilon_j\})$. For a fixed log composition this collapses to $(\mathrm{AUC}_c,\pi,\tau)$.
We call $\tau$ the \emph{binding decay length} for that composition: the number of log lines over which the discriminability above the floor falls by a factor $e$, with half-life $\tau\ln2$. It is a parameter of the fitted model, not an operational horizon of the reader, which is defined by the gap between reading the state and reading the log. It is not a property of the reader alone either, since event lines cost several times more than filler.
On the benchmark mix the two Qwen readers have the same intercept and decay lengths that differ by a factor of about $1.6$, with intervals that overlap only between $76$ and $108$. Doubling the scale appears to lengthen the decay length rather than raise the short-context accuracy. phi-4 combines a short decay length ($\tau=38$ above its floor) with the highest floor.
The estimation protocol is as follows.
(1) Measure each reader at six or more log lengths with paired episodes, and vary the log composition at fixed length.
(2) Fit $\mathrm{AUC}$ and $\mathrm{BA}^\star$ (with cross-validated thresholds) against the line counts by weighted least squares, with intervals from a pair bootstrap.
(3) Select among the exponential (with and without floor), interference and power laws by leave-one-length-out error, and accept a law only if that error is below $0.05$ AUC.
(4) Test (iv) by regressing pair ordering on decisive-event distance at each $T$, and report the criterion separately.
(5) Report $\tau$, $\pi$ and the per-type costs per reader. A recursive filter with $R\ge\log_2\NQ$ has $\tau=\infty$ by Theorem~\ref{thm:achieve}, and the stated-memory arm measures how close a reader gets to it when the filtering is done for it.
}{}

\subsection{Reading the state instead of the log}
\label{sec:exp-state}

The state condition replaces the log by the filter state at the request: one line per key with its current decision and deciding source, expired grants marked as expired, memberships, group grants, and window requirements with the current window status. The reader still has to combine direct grants, group grants and windows. Ranking accuracy returns to 0.95--0.99 for Qwen 14B and to 0.98--0.99 for phi-4 at every horizon (\Cref{fig:decay}, dashed). For phi-4 balanced accuracy stays between 0.92 and 0.96 with no detectable dependence on the horizon: the difference between 256 and 8 events has the bootstrap interval $[-0.075,+0.050]$, which does not exclude a decline of a few points. Qwen2.5-32B reads the state at 0.96, 0.96 and 0.85 at 16, 64 and 256 events, with the same calibration offset as the 14B model at the largest state.

No reader reaches the pre-registered level of 0.95 at every horizon. For Qwen 14B it is 0.93--0.95 up to 32 events and 0.85--0.87 beyond. The errors fall on the approve side, and they track the size of the state rather than the horizon. In this generator the background touches more keys as the log grows, so the state grows from 1 to 32 lines, 2.7 lines on average at 8 events and 26.9 at 256. Descriptively, balanced accuracy of the 14B reader is 0.94 with fewer than 8 state lines and 0.86 with 24 or more. Within the 256-event episodes, splitting by state size at the median gives 0.86 and 0.84. The rate sweep in \Cref{sec:exp-rate} separates the two factors by fixing the number of keys.

Adding the raw log to the state hurts. At 256 events the 14B reader drops from 0.85 with the state alone to 0.67 with state and log together.

\subsection{Which extractive memories can work}
\label{sec:exp-separation}

\Cref{thm:separation} and \Cref{prop:recency} predict two failures and one success for memories that keep raw lines. We test them on two families built to the hypotheses of the theorems. In the chain family one key receives $m\in\{5,9,17,33,65\}$ events of a single rank, alternating withdrawals and reconfirmations after an initial grant, and the deny version swaps the contents of the last two key slots. In the blocked family a security-office decision is followed by $b'\in\{2,4,8,16\}$ operations-console events on the same key, which are never applied and are identical in both versions. Every episode has 128 events, 32 of them access events on other keys. Each cell holds 60 pairs.

Retention keeps $b\in\{2,4\}$ lines. BM25 over the line text without its time stamp, a random subset of the lines that mention the key, and a salience score that ranks withdrawals above grants are order oblivious. A gradient score computed by the reader on the full log is order oblivious up to a measurable slack $\delta$. Global recency and key-filtered recency are position based. Keeping the last applied event of each state component is supersession aware. The exact reader replays the retained events in time order. It is the reader of \Cref{prop:aware} and a natural decoder for the others, though not the optimal one.

\begin{figure}[t]
\centering
\begin{tikzpicture}
\begin{axis}[width=0.62\linewidth,height=5.2cm,xmode=log,log basis x=2,xtick={5,9,17,33,65},xticklabels={5,9,17,33,65},
  xlabel={events on the queried key ($m$)},ylabel={error},ymin=-0.03,ymax=0.55,legend pos=outer north east,legend style={font=\scriptsize},grid=major]
\input{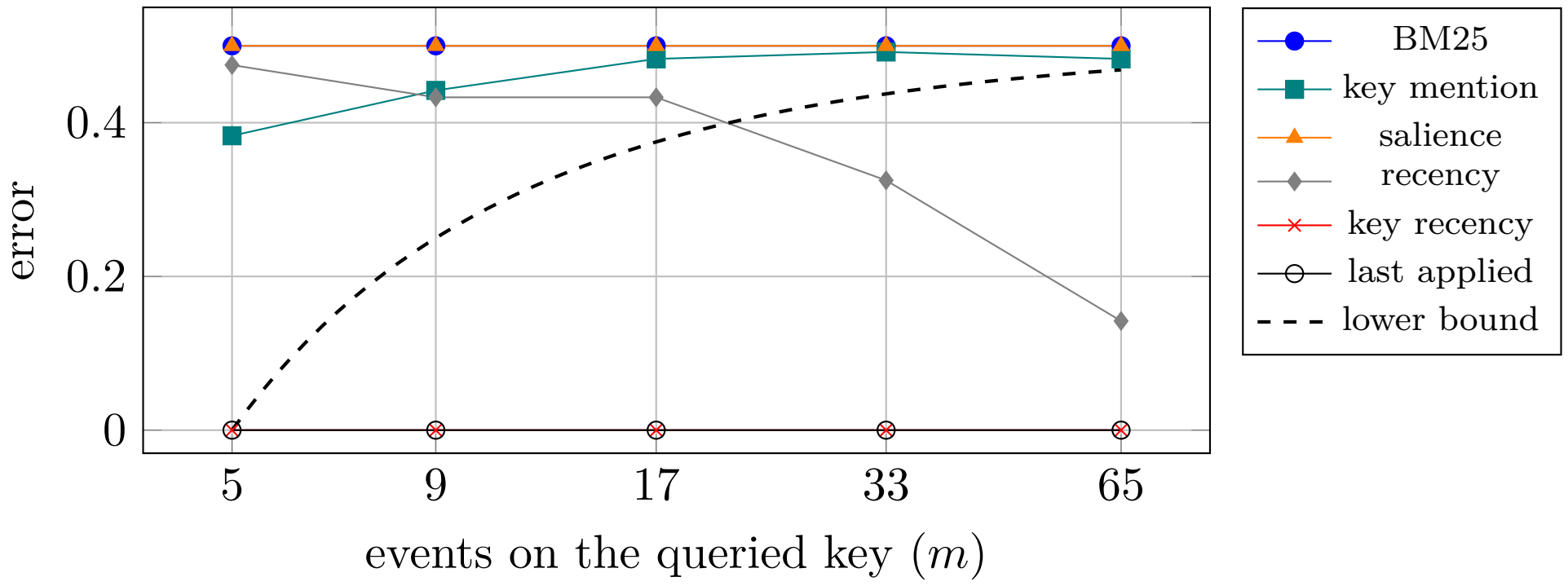}
\end{axis}
\end{tikzpicture}
\caption{Chain counterfactuals with a budget of two lines and the exact reader. The dashed curve is the lower bound $\tfrac12-b/(m-1)$ of \Cref{thm:separation}, valid for every reader of an order-oblivious memory. A rule that keeps a uniformly random subset of the key's withdrawals attains it (\Cref{prop:tight}), and none of the plotted rules does.}
\label{fig:separation}
\end{figure}

\Cref{fig:separation} and \Cref{tab:separation} show the outcome. The three order-oblivious rules stay at or above the bound at every chain length and reach 0.48 to 0.50 at $m=65$, against a bound of 0.47. Key-filtered recency is exact on chains, where the last key event decides, and its error is exactly one half on blocked chains whenever $b'\ge b$, where the last key events are the ones that do not count. With $b=4$ and $b'=2$ the decisive line is still inside the window and the error is zero, as the proposition requires. Keeping the last applied event is exact on both families. The bound itself is attained. Keeping a uniformly random subset of the key's withdrawals, with a reader that knows the slots as the theorem assumes, gives an error within 0.03 of $\tfrac12-\min(b,\tfrac{m-1}2)/(m-1)$ in all 15 combinations of $m\in\{5,\dots,65\}$ and $b\in\{1,2,4\}$, on either side of the bound as sampling allows.

The LLM reader on the same views reproduces the exact reader (\Cref{tab:separation-llm} in \Cref{app:x5}). Order-oblivious rules err 0.48 to 0.50 at $m=65$, key recency errs 0.50 on every blocked family with $b'\ge b$, and the last applied event gives at most one error in a hundred. One difference is instructive. With key recency on chains the exact reader is always right, while the LLM reader errs 0.02 to 0.06 with two retained lines and 0.18 to 0.31 with four, although the extra lines are correct evidence that ends with the deciding event. The same reader given the full 128-event log errs 0.42 to 0.50. The four-line result qualifies the factorial finding of \Cref{sec:exp-decay}. On full logs the number of events on the queried key has no detectable effect, but on a short view in which those events are most of the text, more of them make the reader worse. Interference from the queried key's own history exists. In long logs it is swamped by everything else.

The gradient rule scores lines with the reader's own sensitivity on the full log. It is not order oblivious in practice. Shuffling the log changes its retained set in every episode, so \Cref{cor:delta} gives no bound. It still errs 0.44 to 0.53 everywhere, no better than chance. A sensitivity computed on the full log does not tell which event was applied last.

\begin{table}[t]
\centering
\small
\begin{tabular}{llccccccccc}
\toprule
 & & \multicolumn{5}{c}{chain length $m$} & \multicolumn{4}{c}{blocked events $b'$}\\
\cmidrule(lr){3-7}\cmidrule(lr){8-11}
$b$ & rule & 5 & 9 & 17 & 33 & 65 & 2 & 4 & 8 & 16\\
\midrule
\multirow{7}{*}{2} & BM25 & 0.50 & 0.50 & 0.50 & 0.50 & 0.50 & 0.36 & 0.36 & 0.43 & 0.47\\
 & key mention & 0.38 & 0.44 & 0.48 & 0.49 & 0.48 & 0.17 & 0.35 & 0.36 & 0.43\\
 & salience & 0.50 & 0.50 & 0.50 & 0.50 & 0.50 & 0.50 & 0.50 & 0.50 & 0.50\\
 & gradient & 0.49 & 0.47 & 0.51 & 0.51 & 0.47 & 0.53 & 0.48 & 0.49 & 0.49\\
 & recency & 0.48 & 0.43 & 0.43 & 0.33 & 0.14 & 0.50 & 0.50 & 0.50 & 0.50\\
 & key recency & 0.00 & 0.00 & 0.00 & 0.00 & 0.00 & 0.50 & 0.50 & 0.50 & 0.50\\
 & last applied & 0.00 & 0.00 & 0.00 & 0.00 & 0.00 & 0.00 & 0.00 & 0.00 & 0.00\\
 & bound & 0.00 & 0.25 & 0.38 & 0.44 & 0.47 & -- & -- & -- & --\\
\midrule
\multirow{7}{*}{4} & BM25 & 0.49 & 0.50 & 0.50 & 0.50 & 0.50 & 0.27 & 0.26 & 0.28 & 0.39\\
 & key mention & 0.27 & 0.33 & 0.46 & 0.44 & 0.48 & 0.00 & 0.11 & 0.28 & 0.31\\
 & salience & 0.50 & 0.50 & 0.50 & 0.50 & 0.50 & 0.50 & 0.50 & 0.50 & 0.50\\
 & gradient & 0.52 & 0.48 & 0.49 & 0.47 & 0.48 & 0.48 & 0.47 & 0.50 & 0.44\\
 & recency & 0.43 & 0.41 & 0.32 & 0.17 & 0.02 & 0.50 & 0.50 & 0.50 & 0.50\\
 & key recency & 0.00 & 0.00 & 0.00 & 0.00 & 0.00 & 0.00 & 0.50 & 0.50 & 0.50\\
 & last applied & 0.00 & 0.00 & 0.00 & 0.00 & 0.00 & 0.00 & 0.00 & 0.00 & 0.00\\
 & bound & 0.00 & 0.00 & 0.25 & 0.38 & 0.44 & -- & -- & -- & --\\
\bottomrule
\end{tabular}

\caption{Error of the exact reader on retained views. Chain columns follow \Cref{thm:separation}, blocked columns follow \Cref{prop:recency}.}
\label{tab:separation}
\end{table}

\subsection{Recursive memory under a token budget}
\label{sec:exp-recursive}

A recursive memory sees its previous content and the next 16 events, never the query, and must fit a token budget $B$ measured with the reader's tokenizer. Streams have 256 events with the full background of the benchmark, so the replayed state at the end holds about 30 lines or 500 tokens. Each of 40 pairs (80 streams) is read at $t\in\{32,64,128,256\}$ by Qwen2.5-14B. The LLM methods use the same model as writer. The ledger prompt asks for one line per state component with the current decision, its source and any expiry, and tells the writer to drop superseded facts. \Cref{tab:recursive} lists balanced accuracy.

\begin{table}[t]
\centering
\scriptsize
\setlength{\tabcolsep}{3pt}
\begin{tabular}{lcccccccccccc}
\toprule
 & \multicolumn{4}{c}{$B=128$} & \multicolumn{4}{c}{$B=256$} & \multicolumn{4}{c}{$B=512$}\\
\cmidrule(lr){2-5}\cmidrule(lr){6-9}\cmidrule(lr){10-13}
memory & 32 & 64 & 128 & 256 & 32 & 64 & 128 & 256 & 32 & 64 & 128 & 256\\
\midrule
replayed state (oracle) & 0.94 & 0.64 & 0.55 & 0.53 & 0.94 & 0.85 & 0.71 & 0.60 & 0.94 & 0.90 & 0.90 & 0.79\\
last applied event per component & 0.67 & 0.60 & 0.55 & 0.56 & 0.81 & 0.60 & 0.57 & 0.60 & 0.87 & 0.80 & 0.64 & 0.63\\
last event per key & 0.60 & 0.66 & 0.57 & 0.55 & 0.79 & 0.62 & 0.49 & 0.60 & 0.85 & 0.74 & 0.58 & 0.60\\
LLM state ledger & 0.68 & 0.55 & 0.62 & 0.51 & 0.66 & 0.67 & 0.58 & 0.50 & -- & -- & -- & --\\
LLM summary & 0.58 & 0.56 & 0.52 & 0.54 & 0.70 & 0.62 & 0.58 & 0.53 & -- & -- & -- & --\\
gradient eviction & 0.63 & 0.56 & 0.47 & 0.50 & 0.70 & 0.59 & 0.51 & 0.53 & -- & -- & -- & --\\
BM25 retrieval (full archive) & 0.81 & 0.78 & 0.72 & 0.51 & 0.81 & 0.79 & 0.68 & 0.55 & 0.85 & 0.74 & 0.69 & 0.59\\
recency & 0.54 & 0.56 & 0.56 & 0.51 & 0.61 & 0.58 & 0.55 & 0.54 & 0.75 & 0.61 & 0.54 & 0.56\\
\midrule
full log, no budget & \multicolumn{12}{l}{$t=32$: 0.83\quad $t=64$: 0.65\quad $t=128$: 0.58\quad $t=256$: 0.49}\\
\bottomrule
\end{tabular}

\caption{Recursive memories over 256 events, balanced accuracy at four checkpoints and three budgets. The state of the full benchmark needs about 500 tokens at the last checkpoint. With 40 pairs per cell the 95 percent bootstrap interval of each entry is about $\pm0.1$. BM25 retrieval keeps the full archive and sees the query at read time, while every other memory is written without the query.}
\label{tab:recursive}
\end{table}

Three facts stand out. Below the size of the state every memory, the replayed state included, is at 0.60 or lower at 256 events, since a truncated state is missing components. The differences among these memories are within the per-entry intervals of about $\pm0.1$. With 512 tokens the replayed state holds 0.90 at 128 events and 0.79 at 256, while every other memory evaluated at that budget stays at or below 0.69 and 0.63. The full log without any budget is at 0.49.

The LLM ledger does not realize the advantage of state. It is at 0.66--0.68 after two updates and at chance after sixteen, at both budgets, and it is no better than BM25 retrieval over the full archive (0.50 against 0.55 at 256 events with $B=256$, with a 95 percent bootstrap interval of $[-0.15,+0.01]$ for the difference to the best non-filter memory re-selected in each resample). The failure is erasure, not crowding. With $B=256$ the final ledgers use 171 tokens on average and mention the queried key in only 8 of 80 streams. The writer drops components it judges inactive, and a dropped component never returns. The filter structure is what helps, and an instruction-tuned model asked to maintain it does not maintain it.

Keeping raw evidence with the right supersession rule helps but does not close the gap. Keeping the last applied event of each component reaches 0.87 at 32 events and 0.63 at 256 with $B=512$, below the replayed state at the same budget. The reader then has to apply the policy to raw lines, which is the step it performs poorly in \Cref{sec:exp-decay}. The pre-registered comparison between gradient eviction and per-key recency over horizons (\Cref{thm:separation} on uncontrolled streams) is inconclusive at this sample size, with a difference of $+0.08$ and interval $[-0.11,+0.27]$. The controlled test of \Cref{sec:exp-separation} is the decisive one.

\subsection{A filter that can be built}
\label{sec:exp-buildable}

The replayed state assumes perfect parsing. A buildable version replaces it with a model. Each event line is shown on its own to Qwen2.5-14B, without the query and without any other event, together with a fixed instruction and three generic examples, and the model returns a typed update in JSON: event type, source, service, resource, group and expiry. The transition rule of the benchmark applies the updates in time order, and the reader sees the rendered state exactly as in the oracle condition. Nothing about the decision is delegated to the parser. It only has to read one line.

It reads lines almost perfectly. Of 35{,}744 distinct event lines, 35{,}743 are parsed with every field correct. The one error puts a resource name in the service field of a window requirement.

\begin{table}[t]
\centering
\small
\begin{tabular}{llcccc}
\toprule
reader & memory & $t=32$ & 64 & 128 & 256\\
\midrule
Qwen2.5-14B & parsed events + transition rule & 0.94 & 0.90 & 0.88 & 0.80\\
 & replayed state (oracle) & 0.94 & 0.90 & 0.90 & 0.83\\
 & LLM ledger, $B=256$ & 0.66 & 0.67 & 0.58 & 0.50\\
 & LLM applies the parsed updates & -- & 0.72 & -- & 0.59\\
 & LLM fact store (add, update, delete), $B=512$ & -- & 0.83 & -- & 0.59\\
 & BM25 retrieval, $B=512$ & 0.83 & 0.72 & 0.69 & 0.59\\
 & recency $+$ relevance, $B=512$ & 0.83 & 0.80 & 0.66 & 0.55\\
 & full log with reasoning trace, 768 tokens$^\ast$ & -- & 0.78 & -- & 0.54\\
 & full log with reasoning trace, 4096 tokens & -- & -- & -- & 0.64\\
\midrule
phi-4 & parsed events + transition rule & 0.89 & 0.93 & 0.89 & 0.84\\
 & replayed state (oracle) & 0.89 & 0.93 & 0.90 & 0.86\\
 & LLM ledger, $B=256$ & 0.75 & 0.67 & 0.70 & 0.60\\
 & LLM applies the parsed updates & -- & 0.78 & -- & 0.58\\
 & LLM fact store (add, update, delete), $B=512$ & -- & 0.70 & -- & 0.54\\
 & BM25 retrieval, $B=512$ & 0.85 & 0.81 & 0.79 & 0.66\\
 & recency $+$ relevance, $B=512$ & 0.85 & 0.79 & 0.75 & 0.58\\
 & full log with reasoning trace, 768 tokens$^\ast$ & -- & 0.88 & -- & 0.68\\
 & full log with reasoning trace, 4096 tokens & -- & -- & -- & 0.73\\
\midrule
Qwen2.5-32B & parsed events + transition rule & 0.98 & 0.95 & 0.90 & 0.89\\
 & replayed state (oracle) & 0.98 & 0.95 & 0.91 & 0.91\\
 & BM25 retrieval, $B=512$ & 0.83 & 0.77 & 0.71 & 0.68\\
 & recency $+$ relevance, $B=512$ & 0.81 & 0.82 & 0.74 & 0.60\\
 & full log with reasoning trace, 4096 tokens & -- & -- & -- & 0.68\\
\bottomrule
\end{tabular}

\caption{Balanced accuracy over 256 events with the full background (80 streams, 40 pairs), read by three readers. The parsed filter uses Qwen2.5-14B as parser for both readers. $^\ast$The 768-token reasoning-trace reader reads the full log on the separate test streams of \Cref{sec:exp-decay}, and gives no decision within the cap in 30 (Qwen) and 15 (phi-4) of 120 cases at 256 events, which count as errors. The 4096-token reader reads the same streams as the other rows, and reaches the cap without a decision in 5 (Qwen 14B), 0 (phi-4) and 9 (Qwen 32B) of 80 cases. The two rows in which the reader model maintains the state from the same parsed updates, or edits a fact store, were run at 64 and 256 events only. Qwen2.5-32B reads with the Qwen2.5-14B parses.}
\label{tab:buildable}
\end{table}

\Cref{tab:buildable} compares the buildable filter with the non-filter memories of earlier rounds. The ledger-protocol ablation is reported below. At 256 events it reaches 0.80 with the Qwen reader and 0.84 with phi-4, within 0.03 of the oracle state. The strongest alternative in the table for both readers is the full log read with a reasoning trace of up to 4096 tokens on the same streams, at 0.64 and 0.73. The filter leads it by $+0.16$ with paired interval $[0.04,0.28]$ for Qwen and by $+0.11$ with $[0.03,0.21]$ for phi-4, which meets the pre-registered margin of 0.10 with a lower limit above zero for both readers. After a Holm adjustment over the eight registered superiority comparisons of the second and third rounds, the phi-4 lead is no longer significant ($p=0.017$ against a threshold of 0.0167, \Cref{app:prereg}). Every memory in the table other than the reasoning traces stays at or below 0.66 for these two readers (0.68 for Qwen2.5-32B), and the code-kept tables of the ablation below reach 0.70. A shorter cap of 768 tokens on separate streams gives 0.54 and 0.68, mostly because traces stop before a decision. A reasoning trace closes much of the gap at 64 events, from 0.59 to 0.88 for phi-4, and less of it at 256 events. On the controlled backgrounds of \Cref{sec:exp-rate} the parsed filter stays within 0.013 of the oracle state in all four cells run ($K\in\{3,15\}$, $T\in\{64,256\}$) with both readers, while retrieval stays between 0.55 and 0.70. The retrieval baselines here are time-aware at reading: BM25 scores the text of each line without its event number, breaks ties in favour of later lines, and shows the kept lines in time order with their event numbers, so their failure is not the order-blind failure of \Cref{thm:separation}.

A larger reader does not change the ranking. With Qwen2.5-32B reading the same streams, the parsed filter answers 0.89 at 256 events, against 0.91 for the oracle state, 0.68 for BM25 retrieval, 0.60 for recency and relevance and 0.68 for the full log read with a reasoning trace of up to 4096 tokens, which ends without a decision in 9 of 80 cases. The filter leads the better retrieval by $+0.21$ ($[0.11,0.30]$) and the trace by $+0.21$ ($[0.11,0.33]$).

Of the 80 episodes at 256 events, 61 are decided by the queried pair's direct line alone and 19 need a group grant or a change window. The filter's advantage lies mostly in the first kind. QwQ-32B (\Cref{sec:exp-reasoning}) reads the log at 0.84 on direct lookups and 0.88 on compositions, and the parsed state at 1.00 and 0.86. On compositions the parsed state trails the true state, 0.73 against 0.88 for the Qwen 14B reader, because the one parsing error falls on a window requirement.

phi-4 as parser gives the same picture. It parses 99.95 percent of the full-background lines correctly, failing only the five group grants, and its parses give 0.83 at 256 events with the phi-4 reader, against 0.84 with the Qwen parses and 0.86 for the oracle.

\paragraph{Who applies the rule?} The ledger and the parsed filter differ in more than one way: the ledger rewrites free text every sixteen events, while the filter receives typed updates and applies them in code. Two intermediate memories separate these. In the first the reader model receives the same typed updates as the filter, sixteen at a time, together with its current state table and the policy, and returns the updated table with one line per fact. Parsing is therefore exactly as good as in the filter, and only the application of the rule is left to the model. In the second, a re-implementation of the add, update and delete design of Mem0 \citep{chhikara2025mem0} with our own prompts, the model reads the raw lines sixteen at a time and edits a store of facts with add, update and delete operations, within 512 tokens. At 256 events the first reaches 0.59 with Qwen and 0.58 with phi-4 and the second 0.59 and 0.54, close to the ledger and 0.21 to 0.30 below the filter (paired intervals with lower limits from 0.09 to 0.18, rows R2-B1 and R2-B3 of \Cref{app:prereg}). Unlike the ledger, the model-applied table still names the queried pair in 51 to 65 percent of streams, so the component is usually kept, yet accuracy stays near the ledger. Structured input, typed updates and a fixed table format do not rescue the model when it rewrites the table. Taking the bookkeeping away helps the Qwen reader. In a further registered ablation the code keeps a table that can only be overwritten, never deleted from, and the model only writes the new value of each entry an update touches, one update or sixteen at a time. With the Qwen reader this reaches 0.70 and 0.69 at 256 events, about two thirds of the way from the ledger (0.50) to the filter and half of the way from the model-applied table (0.59), and the filter's remaining lead of $+0.10$ ($[-0.03,0.19]$) is no longer significant. With phi-4 the same tables reach 0.53 and 0.61 and the filter leads by $+0.20$ ($[0.09,0.26]$). Rewriting the whole table one update at a time gives 0.64 and 0.63, and the raw log sorted by key 0.61 and 0.64. Part of the ledger's failure is erasure, which a persistent table removes, and part is in the entries the model writes, which an explicit rule removes.

\begin{table}[t]
\centering
\small
\begin{tabular}{p{8.6cm}cc}
\toprule
what is exact, what a model does & Qwen2.5-14B & phi-4 \\
\midrule
exact rule on the true events, no reader & 1.00 & 1.00 \\
exact rule on the parsed events, no reader (parsing) & 0.98 & 0.98 \\
reader on the true state (decoding) & 0.83 & 0.86 \\
reader on the state from parsed events (parsing and decoding) & 0.80 & 0.84 \\
reader on a table the model updates from the parsed events (transition) & 0.59 & 0.58 \\
reader on the full log with a 4{,}096-token trace (everything) & 0.64 & 0.73 \\
exact decision of the rule a 32B code model wrote, after clarification (R4-4b) & \multicolumn{2}{c}{0.93} \\
\bottomrule
\end{tabular}
\caption{Where the errors arise, first domain, 80 full-background streams at 256 events, balanced accuracy. The parser is Qwen2.5-14B in every row that parses. Each row makes one more stage a model's job.}
\label{tab:stages}
\end{table}

\Cref{tab:stages} puts these stages side by side. Parsing costs about 0.03, decoding a correct state costs the reader 0.14 to 0.17, and letting the model run the transition costs another 0.21 to 0.26. The code-writing arms of the next paragraph and of R2-B2 are the computation-matched control: they give a model the same policy text and ask it to supply the transition itself. Two failures should be kept apart. A memory that drops a component, like the ledger that mentions the queried key in 8 of 80 streams, fails to maintain the state. A reader that misreads a correct state, like Qwen on the larger state of the second domain (\Cref{sec:exp-domain2}), fails to decode it.

\paragraph{Can the model write the rule?} The filter above uses the benchmark's transition rule. To see whether that rule could come from the model instead, each model wrote Python code for the state update, a renderer and a decision function from the policy text and the update schema alone, three samples at temperature 0.7, executed on a disposable machine. On this domain no sample is usable. Of six samples, five fail on most streams (two crash in their own example code before any stream runs, one fails on 66 of 80 streams and two on all 80, with missing keys or wrong container types), and the one that runs reaches a balanced accuracy of 0.55 and 0.56 at 64 and 256 events while answering almost always deny. The policy combines ranks, leases, groups and windows, and no sample answers it correctly. On the simpler second domain of \Cref{sec:exp-domain2} the best sample of each model reaches 0.91 and 0.86, but each model also produces a sample that fails. We then gave the task to a stronger code model with supervision. Qwen2.5-Coder-32B wrote 16 programs per domain, the program that answered best on a set of labelled histories of 64 events was selected, and it was shown up to three failing histories and asked once to repair itself. On the second domain 10 of the 16 programs are exactly right, and ten labelled histories suffice to select one that matches the author's rule on every test stream. On the first domain the best program is exact at 64 events and falls to 0.80 at 256, against 0.98 for the human-written rule on the same parsed updates, a gap of $-0.18$ ($[-0.25,-0.10]$). It fails on the priority and restore families, which need rank precedence and reinstatement, and with 50 or 200 histories selection picks the same program and repair does not improve it. An open 32B code model can supply the rule when the policy is simple. When the rule involves rank precedence, it has to come from a stronger writer or from people, and be checked against labelled histories, or be selected from programs among which at least one is right. The reinstatement failures trace to our specification (next paragraph).

\paragraph{A rule written by someone else.} The filter above runs the rule of the benchmark's own replay. To test whether that rule is recoverable from the policy alone, a separate coding agent (Claude Opus 5.5, a proprietary frontier model) with no access to our code, data or paper wrote the update and decision functions from the policy text of \Cref{app:policy} and the update format, and listed 19 points where the text left a choice. On 2{,}000 fresh histories its answers differ from replay in 257 cases, all in the restore family and all from one of those points. The update format names a reconfirmation but does not say what it does to a withdrawn grant, and the agent read it as reaffirming the standing decision, while replay treats it as a grant from its source. The log lines themselves are not ambiguous (``access\_admin confirmed again that svc\_kestrel may write to bucket\_audit''), so the gap concerns the format given to code writers, not the input of any reader. As written, the independent rule reaches 1.00 and 0.85 on the parsed test streams at 64 and 256 events, against 1.00 and 0.98 for ours, a gap of $-0.13$ ($[-0.20,-0.06]$) at 256. It agrees with replay on all 248 histories of the priority family, so rank precedence was recovered from the text. With that single choice changed after seeing this result, read as the log lines state it, and the other 18 choices left as written, it agrees with replay on all 2{,}000 histories and matches our rule on the parsed test streams (1.00 and 0.98). The code models of the previous paragraph were given the same format, so their failures on the restore family are at least partly a failure of our specification. Rerun with one added sentence saying that a reconfirmation acts as a grant from its source, the code model's selected program reaches 0.93 at 256 events against 0.98 for our rule, a gap of $-0.05$ ($[-0.10,-0.01]$) that sits exactly on the registered tolerance. Restore episodes are now all correct. Rank precedence is still wrong in the selected program, whose answers on the priority family are at 0.50.

\paragraph{Paraphrased logs.} The benchmark lines come from a small set of templates. To test whether the result depends on them, phi-4 rewrote every distinct line of the full-background streams in different words and sentence structure, at temperature 0.8, with instructions to keep every fact. Ground truth still comes from the original events. The Qwen parser reads 97.9 percent of the rewritten lines correctly, and most of its errors trace to the rewriter changing identifiers it was told to keep, such as turning \texttt{svc\_heron} into Heron. On the rewritten logs the parsed filter answers 0.85 and 0.83 at 64 and 256 events with the Qwen reader and 0.83 and 0.80 with phi-4, while reading the rewritten log itself gives 0.68 and 0.53, and 0.72 and 0.50. For the Qwen reader the gap between filtering and long-context reading is 0.30 at 256 events on the rewritten logs and 0.31 on the templates, and it shrinks from 0.26 to 0.16 at 64 events. A raw-log baseline for phi-4 on the template streams was not run. Because ground truth stays with the original events, a rewrite that drops an identifier corrupts the input rather than rewording it. In 12 of the 80 streams at 256 events, and 6 at 64, a line that decides the query lost or changed a name, office or number. Removing those streams changes no balanced accuracy above by more than 0.02.

The comparison with the ledger is the point of the experiment. The same model that cannot keep a ledger can parse every line, and an explicit rule does the rest. With phi-4 as both writer and reader the ledger ends at 0.60.

\paragraph{Published memory designs and a query-time retriever.} We registered a comparison with three further memories on the 80 streams of \Cref{tab:buildable}, all written without the query, against the parsed filter cut to the same 512 tokens (row R5-3, Holm adjustment over the family). StateMem \citep{fan2026can} is the closest published design: one model call per event turns the line into typed state units in a persistent store, the model flags which units a new line supersedes, and a deterministic pass re-checks units that depend on a superseded one. No code is released, so we reimplemented it from its description with our own prompt wording and Qwen2.5-14B as the encoder, in a native form and with our policy as task context. The native encoder keeps 209 units on average after 256 events, most of them filler lines, about 9{,}300 reader tokens for Qwen2.5-14B and 7{,}900 for phi-4. After one retry, 63 of the 321 encoder outputs per stream fail to parse in the native form and 7 of 257 with the policy. Read whole it scores 0.55 with Qwen2.5-14B and 0.54 with phi-4, cut to 512 tokens 0.51 and 0.49. With the policy as context it keeps 65 units and scores 0.63 and 0.64. Both forms stay at or near chance on the priority family (0.50 to 0.63 read whole, eight streams), as \Cref{prop:recency} predicts for supersession by recency without source ranks, and writing a 256-event log costs them 0.46 to 1.4 million tokens. The capped filter beats every StateMem variant by $+0.25$ to $+0.33$, significant after the adjustment. Native Mem0 \citep{chhikara2025mem0} ran as the open-source library (mem0ai 2.2.1 with its default prompts and thresholds, Qwen2.5-14B as the encoder, bge-small-en-v1.5 embeddings and a local Qdrant store), receiving each log line as one message. Its update step rarely merges or deletes, and the store grows from 76 memories at 64 events to 386 at 256. At query time its search returns the ten memories most similar to the query, about 300 tokens. These come from the whole log (among the two thirds that keep an event number, mean event index 138 of 256, 30 percent from the last quarter), so similarity finds lines about the queried resource but does not prefer the latest one, which makes the selection close to the order-oblivious rules of \Cref{thm:separation}. It scores 0.46 with Qwen2.5-14B and 0.55 with phi-4, at or below chance in every family except expiry with phi-4, and the capped filter beats it by $+0.33$ and $+0.26$, significant after the adjustment. All six comparisons against StateMem and Mem0 survive the Holm adjustment over the eight registered tests, and the two against the retriever do not. Adding the two second-domain retriever tests of R5-4 to the family, as registered, leaves this unchanged and keeps the phi-4 lead on the second domain significant. Qwen2.5-32B, read outside the registered family, repeats the pattern: the capped filter answers 0.86, StateMem 0.50 and 0.55, Mem0 0.54 and the retriever 0.86, so the filter leads the published designs by $+0.31$ to $+0.36$ and is level with the retriever ($+0.00$, $[-0.11,0.11]$), which this reader too reads poorly at 64 events (0.57).

The third memory keeps the whole archive and selects at query time: every line that names the queried pair, the principal's memberships, group grants on the resource and the resource's window, in time order with event numbers. On these streams replaying only those lines gives the exact answer in all 160 cases, with about three lines per query. Read by Qwen2.5-14B and phi-4 it scores 0.78 and 0.89 at 256 events, level with or above the filter ($+0.01$, $[-0.11,0.13]$, and $-0.08$, $[-0.18,0.03]$), so the registered margin is not met. At 64 events phi-4 reads it at only 0.61, $0.31$ ($[0.20,0.43]$) below the filter, while Qwen2.5-14B reads it at 0.92. On the second domain it is level with the filter for Qwen and at 1{,}024 events for both readers (\Cref{sec:exp-domain2}). By the reading we fixed before the run, what decides is not how much of the log a memory compresses but whether it selects by the state's dependencies. The filter does so at write time, without the query and with storage bounded by the state. The retriever does so at read time, with the query, the whole archive and the dependency structure of the policy written into its selection rule. Where the query is known at read time and the archive can be kept, query-time selection by state dependencies is as good a memory as the filter.

\subsection{Readers trained to reason}
\label{sec:exp-reasoning}

The readers above are instruction-tuned models scored mostly by the likelihood of an answer. Models trained to reason before answering might track the state from the log on their own. We ran two of them, QwQ-32B and DeepSeek-R1-Distill-Qwen-32B, with up to 16{,}384 generated tokens and the same policy, on the streams of \Cref{tab:buildable}. QwQ-32B (revision 976055f) sampled at temperature 0.6, top-$p$ 0.95 and top-$k$ 40, R1-Distill-Qwen-32B (revision 711ad2e) at temperature 0.6 and top-$p$ 0.95, both with seed 0, a context of 24{,}576 tokens and vLLM 0.31.0 in bfloat16. Every trace ended with a decision, and none reached the cap.

\begin{table}[t]
\centering
\small
\begin{tabular}{lccccc}
\toprule
 & \multicolumn{2}{c}{full log} & state from parsed events & \multicolumn{2}{c}{tokens per decision, 256 events}\\
reader & 64 & 256 & 256 & full log & state \\
\midrule
QwQ-32B & 0.90 & 0.85 & 0.98 & 7{,}052 + 2{,}513 & 745 + 787 \\
R1-Distill-Qwen-32B & 0.89 & 0.70 & 0.96 & 7{,}047 + 1{,}755 & 740 + 686 \\
\bottomrule
\end{tabular}
\caption{Reasoning-trained readers, balanced accuracy on the 80 full-background streams. Tokens are prompt plus generated, averaged over requests.}
\label{tab:reasoning}
\end{table}

These readers change the size of the effect, not its sign (\Cref{tab:reasoning}). Reading the full log, QwQ-32B answers 0.850 at 256 events, 0.125 above the best instruction-tuned reader on these streams (phi-4 with a 4096-token trace, 0.725), and on the test seeds of \Cref{sec:exp-decay} it goes from 0.92 at 16 and 64 events to 0.87 at 256. R1-Distill falls from 0.89 to 0.70. Given the state built from parsed events instead, they answer 0.98 and 0.96. The registered test required the state to lead the log by at least 0.10 with a lower limit above zero for each reader, and both pass: $+0.13$ ($[0.04,0.21]$) and $+0.26$ ($[0.15,0.38]$). Reading the true state at 256 events QwQ is at 1.00, the first reader we measured at or above the 95 percent level we registered for state reading, here at the longest horizon only (R1-Distill was not run on these seeds). At read time, once the state has been built, it is also about six times cheaper: a decision from the log costs QwQ about 9{,}500 tokens, one from the state about 1{,}500 (\Cref{sec:exp-cost} counts the cost of building it).

\paragraph{Stronger readers and longer logs.} Reasoning narrows the gap, so before running stronger readers we registered how their results would change our claims (row R4-1 of \Cref{app:prereg}). If the state still led the log by at least 0.10 with a lower limit above zero at 1{,}024 events, the claim would stand. If it did so at 1{,}024 but not at 256 events, the claim would hold only beyond a length that depends on the reader. If it did so at neither, the claim would shrink to cost and to small readers. The test uses the best of three readers on 100 new pairs at 256 and 1{,}024 events, with the same Qwen 14B parser building the state (\Cref{tab:strong}).

\begin{table}[t]
\centering
\small
\resizebox{\linewidth}{!}{\begin{tabular}{lrccccrr}
\toprule
 & & full & state from & true & state from parsed & \multicolumn{2}{c}{tokens per decision}\\
reader & events & log & parsed events & state & minus log & full log & parsed state \\
\midrule
gpt-oss-120b & 256 & 0.95 & 0.99 & 0.99 & $+0.04$ $[0.01,0.08]$ & 6.1k + 1.6k & 0.8k + 0.4k \\
gpt-oss-120b & 1{,}024 & 0.87 & 0.99 & 0.98 & $+0.12$ $[0.07,0.17]$ & 23.4k + 4.7k & 0.9k + 0.4k \\
Qwen2.5-72B (4-bit) & 256 & 0.72 & 0.94 & 0.95 & $+0.22$ $[0.16,0.28]$ & 7.1k + 0.2k & 0.8k + 0.1k \\
Qwen2.5-72B (4-bit) & 1{,}024 & 0.55 & 0.92 & 0.92 & $+0.37$ $[0.31,0.43]$ & 27.5k + 0.1k & 0.8k + 0.1k \\
QwQ-32B & 1{,}024 & 0.52 & -- & -- & -- & 27.5k + 10.1k & -- \\
\bottomrule
\end{tabular}
}
\caption{Stronger readers on 100 pairs of full-background streams at 256 and 1{,}024 events, balanced accuracy. The state from parsed events uses the Qwen2.5-14B parser and the exact rule. Differences are paired over pairs with 95\% bootstrap intervals. Tokens are prompt plus generated, averaged over decisions.}
\label{tab:strong}
\end{table}

gpt-oss-120b, a reasoning model with 120 billion parameters, reads the 256-event log at 0.95. The state from parsed events leads it by only $+0.04$ ($[0.01,0.08]$), short of the registered margin: at this length a strong enough reader recovers the state from the log by itself, at about six times the tokens. At 1{,}024 events its reading of the log falls to 0.87 while the state stays at 0.99, a lead of $+0.12$ ($[0.07,0.17]$). Qwen2.5-72B, answering without a trace, falls from 0.72 to 0.55 while the state stays at 0.94 and 0.92. QwQ-32B, which reads the 256-event log at 0.85 (\Cref{tab:reasoning}), answers 0.52 on the 1{,}024-event log, where 27 of 200 traces reach the cap of 16{,}384 generated tokens and 25 end without a decision. QwQ's 256-event value comes from the 80 streams of \Cref{tab:reasoning}, not from the 100 new pairs, and cannot change the choice of the best reader (0.85 against 0.95). The best reader of the log at both lengths is gpt-oss-120b, so by the registered rule the claim narrows to lengths beyond a reader-dependent horizon. The read-time cost of the state does not grow with the log: at 1{,}024 events a decision from the state costs gpt-oss-120b about 1{,}300 tokens, one from the log about 28{,}000.

\paragraph{Operational horizon.} For a reader $r$ under a protocol $P$, meaning the task distribution, the background composition, the prompt, the inference budget and balanced accuracy on paired episodes, the operational horizon $H_\Delta(r;P)$ is the shortest log length at which the state built from parsed events beats reading the log by at least $\Delta$ in point estimate, with the lower limit of the 95 percent pair-bootstrap interval above zero. We use $\Delta=0.10$, the margin registered for this test, and report $H_{0.10}$ as a bracket between the last tested length at which the criterion fails and the first at which it holds. On the first domain $H_{0.10}$ is at or below 256 events for Qwen2.5-72B, QwQ-32B and R1-Distill-Qwen-32B. For gpt-oss-120b a registered follow-up (row R5-5) places it above 512 events, where the lead is $+0.07$ ($[0.03,0.11]$) with the log read at 0.92, and at about 1{,}024. There the registered sampling seed gives $+0.12$ ($[0.07,0.17]$), a second seed reads the log at 0.90 instead of 0.87, a difference that is not significant, and gives $+0.09$ ($[0.04,0.13]$), and the two seeds together give $+0.10$ ($[0.06,0.14]$), computed after seeing the result and reported only. The state leads significantly at 1{,}024 events under both seeds, but by an amount that sits on the registered margin, so the strongest reader's horizon is not yet clearly passed at the longest length we ran. The horizon is a property of a reader and a protocol, not a consequence of the theory. It differs from the effective context length of RULER \citep{hsieh2024ruler}, which thresholds one reader's accuracy, because it thresholds the gap between two memories read by the same reader. It also differs from the binding decay length $\tau$ of the reader decay model in \Cref{sec:exp-decay}, which is a fitted parameter of a description.

\subsection{How parse errors propagate}
\label{sec:exp-parsers}

The filter of \Cref{sec:exp-buildable} uses a parser that is almost never wrong. The noisy-parser analysis of \Cref{sec:theory} predicts what happens when it is. We ran parsers of 0.5, 1.5, 3, 7 and 14 billion parameters from the same family on the full-background streams. They read 27, 76, 93, 100 and 99.99 percent of the test lines correctly. Before reading any test parse, we classified each parser's errors on separate calibration lines (drop, wrong key, flipped decision, wrong rank, wrong lease, other), simulated each test stream 200 times with errors drawn independently at the calibrated rates, and predicted the balanced error of the exact rule on the resulting state.

The predicted errors rank the cells in roughly the right order, correctly at 64 events and not at 256, where the 3B parser does better than predicted. Over the eight registered cells, the Spearman correlation between predicted and observed error is 0.83, above the registered 0.8, while observed errors range from 0 to 0.45. The cell-wise criterion, a gap of at most 0.03 or two standard errors, holds in 7 of 10 cells (5 of the 8 registered cells). After the run we found that our simulation of the ``other'' class copied a whole calibration parse, keys included, where the registered procedure replaces only the field that was wrong. With that corrected, 9 of 10 cells hold (7 of 8). The remaining miss is the 0.5B parser at 256 events, where the prediction of 0.37 overstates the observed 0.24. The independence assumption itself does not hold. Grouping test lines by wording template, the error indicator of the 0.5B, 1.5B and 3B parsers has an intraclass correlation of 0.63 to 1.00 within templates for at least one event type each, as we registered as the test, so these parsers err by template, not by line. Propagated to the reader, the prediction $\tfrac12+(\mathrm{BA}_{\rm state}-\tfrac12)(1-2\,\mathrm{Err})$ is within 0.10 of the measured balanced accuracy in 16 of 20 cells as registered and in 18 of 20 after the correction, and within 0.03 for the 7B and 14B parsers. Where it misses, the reader does better than predicted.

\paragraph{Correcting a weak parser.} Systematic parse errors can be inverted. The plug-in filter applies the exact rule to whatever the parser says. A calibrated Bayes filter instead keeps a distribution over each state component and updates it by the exact rule mixed over the possible true events behind each parsed line, with the confusion probabilities counted on the calibration lines (\Cref{prop:plugin-map} gives when the two decisions coincide). The method and code were frozen before the filter read any test parse. On the test streams it cuts the balanced error of the 0.5B parser from 0.31 to 0.04 at 64 events and from 0.24 to 0.09 at 256, and of the 1.5B parser from 0.45 to 0.04 and from 0.33 to 0.09, all four reductions with 95 percent intervals below zero and within 0.08 of the reductions estimated on the calibration streams alone. For the 3B, 7B and 14B parsers it leaves the balanced error unchanged (3B: $+0.003$ at 64 events). A registered audit then asked where the gain comes from (row R5-6 of \Cref{app:prereg}, diagnostic because the test result was known). A memoryless correction, which replaces each parsed line by its most likely true event under the same calibration and then applies the plug-in rule, reaches exactly the same balanced error in all ten parser and checkpoint cells and under eight reweightings of the families. An exact joint filter over the ten components a query depends on agrees with the product of marginals on every episode, and maintaining all 68 components without knowing the query gives the same decisions at a storage of 8 to 21 kilobits. The gain is therefore calibrated confusion correction, not sequential inference: the 0.5B and 1.5B parsers misread almost every grant line in the same way, and inverting that confusion needs no memory of earlier lines. Calibrating with the test streams' wording templates held out removes the 0.5B gain at 256 events and keeps more than half of it in the other three cells, so part of the correction is specific to templates. The error that remains, 0.04 and 0.09 for the 0.5B, 1.5B and 3B parsers alike, lies entirely in the window family, where the 3B parser reads every opened window as a window requirement and neither correction can tell the two apart.

\subsection{A second, independently specified domain}
\label{sec:exp-domain2}

The benchmark, its rule and its renderer were written by the same authors as the filter. A second domain was therefore specified and implemented separately, by an agent that had no access to this paper or its code, from a brief that fixed only the interface. In this domain an export shipment may leave the dock if it holds a customs clearance that counts, its payment is confirmed and it carries no hold. Clearances are granted, revoked and reinstated, a reinstatement restores the remembered clearance with its original limit, clearances may be valid only through a given event, a compliance lift removes every hold while a floor lift removes only the floor hold, and many filler lines name shipments without changing anything. Episodes come in pairs that differ only in the decisive lines, labels are balanced and reinstatements make up a third of the pairs. The rules are given to the reader as a 246-word specification.

\begin{table}[t]
\centering
\small
\begin{tabular}{lcccccc}
\toprule
 & parsed lines & \multicolumn{2}{c}{full log} & \multicolumn{2}{c}{state / parsed + rule} & reasoning trace \\
reader & correct & 64 & 256 & 64 & 256 & 256 \\
\midrule
Qwen2.5-14B & 99.95\% & 0.54 & 0.51 & 0.76 / 0.76 & 0.68 / 0.68 & 0.58 \\
phi-4 & 97.07\% & 0.59 & 0.50 & 0.84 / 0.85 & 0.76 / 0.76 & 0.68 \\
\bottomrule
\end{tabular}
\caption{The second domain, balanced accuracy over 40 pairs per horizon. ``State'' is the domain's own rendering of the true state, ``parsed + rule'' parses each line with the reader model and applies the domain author's update function. The reasoning trace has up to 4096 tokens.}
\label{tab:domain2}
\end{table}

Two of the three findings carry over (\Cref{tab:domain2}). Reading the full log falls to chance by 256 events for both readers. Parsing each line and applying the rule beats it by $+0.16$ (paired interval $[0.06,0.26]$) and $+0.26$ ($[0.18,0.34]$) at 256 events, and parsing loses nothing against the true state. The third does not carry over in full. Reading the state itself drops by 0.09 and 0.08 from 64 to 256 events, with intervals that include zero, and at 256 events the lead of the filter over a reasoning trace is $+0.10$ and $+0.09$ with intervals that include zero. The reason is visible in the errors. This domain's renderer lists every shipment, about 800 to 1000 tokens at 256 events, and writes a time-limited clearance as ``through Event $N$'' next to the number of the last event, so the reader still has to compare two numbers. On the expiry family the readers answer 0.50 and 0.55, at chance, and on the joint-condition family 0.63. The first domain's renderer resolves expiry before the reader sees it.

We therefore registered a second rendering before running it, identical to the author's except that a time-limited clearance is written as in force through event $N$ or as expired after event $N$, as the first domain does for leases. With phi-4 the second domain then reproduces all three findings. Reading the true state gives 0.93 and 0.90 at 64 and 256 events, parsing and applying the rule gives 0.93 and 0.89, the expiry family rises from 0.55 to 0.86, and the lead over the reasoning trace at 256 events is $+0.21$ ($[0.08,0.35]$), meeting the registered margin. With Qwen2.5-14B it does not. State reading gives 0.80 and 0.64 at 64 and 256 events, a change of $-0.16$ ($[-0.26,-0.06]$), the expiry family stays at 0.50 at 256 events (eleven pairs), the reinstatement family moves from 0.70 to 0.55 (ten pairs), and the lead over the trace is $+0.06$ ($[-0.04,0.18]$). The state of this domain lists every shipment and reaches about 1000 tokens at 256 events, twice the state of the first domain, and the first domain already showed Qwen's state reading falling with the number of state lines (\Cref{sec:exp-state}). How much of the decision the state leaves to the reader, and how much of the state it shows, decide how much a filter helps.

\paragraph{A frozen replication with fresh wording.} Every choice above was made on the streams it was then evaluated on. To test the second domain on data no analysis had touched, we froze the parser prompt, the domain author's update function, the renderer that resolves expiry and the analysis, timestamped the registration externally (row R5-4), and generated new streams with new seeds. Their wording is new as well: Qwen2.5-72B, which neither parses nor reads in this paper, wrote three phrasings for every event type and one for every filler line, each checked to keep its placeholders and to parse back to exactly one event. Reading the parsed state beats reading the log at 256 events by $+0.22$ ($[0.17,0.28]$) for Qwen2.5-14B and $+0.35$ ($[0.30,0.40]$) for phi-4, and at 1{,}024 events by $+0.20$ and $+0.22$. gpt-oss-120b reads the 256-event log at 0.98, level with the state, and the 1{,}024-event log at 0.72, against 1.00 for the state, a lead of $+0.28$ ($[0.21,0.36]$). The second domain therefore shows the same reader-dependent horizon as the first, with a larger gap beyond it. A bounded store holding three quarters of the mean number of shipments follows its prediction in all twelve reader, policy and length cells, with gaps of 0.003 to 0.083. Against a retriever that, at query time, takes every line naming the queried shipment, the state leads by $+0.20$ ($[0.13,0.26]$) for phi-4 and by $+0.06$ ($[0.00,0.12]$, a lower limit of zero, so not significant) for Qwen at 256 events, and the two are level at 1{,}024 events. In this domain every component of the queried shipment lives on the lines that name it, so selecting those lines at query time recovers most of what the filter computes at write time.

\subsection{Rate thresholds over keys and horizons}
\label{sec:exp-rate}

\Cref{thm:achieve,thm:converse} say that the budget a state memory needs is set by the number of keys and not by the horizon. We test this with backgrounds restricted to direct access events on $K\in\{1,3,7,15\}$ keys besides the target, so that the size of the state is controlled by $K$, at $T\in\{64,256\}$ and budgets from 32 to 512 tokens. Each cell holds 40 pairs read by Qwen2.5-14B. We call $B^\star$ the smallest budget on the grid at which balanced accuracy reaches 0.85.

\begin{figure}[t]
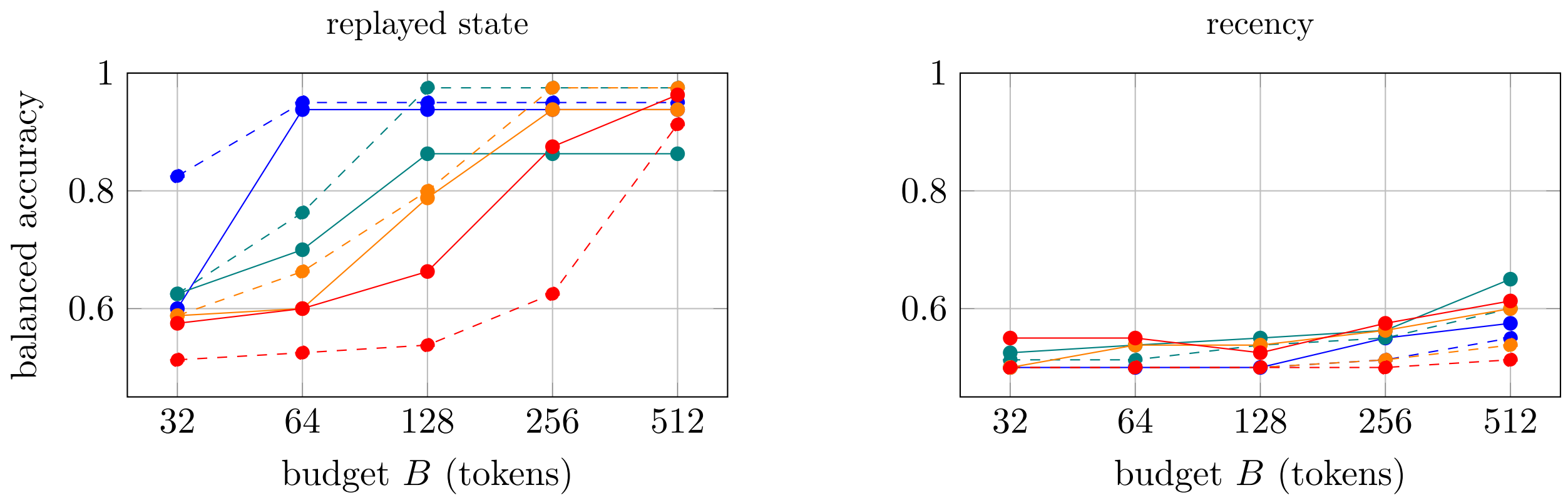

\centering
\begin{tikzpicture}
\begin{axis}[width=0.48\linewidth,height=5cm,xmode=log,log basis x=2,xtick={32,64,128,256,512},xticklabels={32,64,128,256,512},
  xlabel={budget $B$ (tokens)},ylabel={balanced accuracy},ymin=0.45,ymax=1.0,grid=major,title={\small replayed state}]
\input{generated/x4c_plot_state.tex}
\end{axis}
\end{tikzpicture}
\hfill
\begin{tikzpicture}
\begin{axis}[width=0.48\linewidth,height=5cm,xmode=log,log basis x=2,xtick={32,64,128,256,512},xticklabels={32,64,128,256,512},
  xlabel={budget $B$ (tokens)},ymin=0.45,ymax=1.0,grid=major,title={\small recency}]
\input{generated/x4c_plot_recency.tex}
\end{axis}
\end{tikzpicture}
\caption{Balanced accuracy against budget for $K=1,3,7,15$ background keys (blue, teal, orange, red), 64 events (solid) and 256 events (dashed). The threshold of the state memory moves with $K$ and stays put when the horizon quadruples. Recency never reaches 0.85.}
\label{fig:rate}
\end{figure}

For the replayed state, $B^\star$ is 64, 128, 256 and 256 tokens at 64 events and 64, 128, 256 and 512 tokens at 256 events (\Cref{fig:rate}). It grows with $K$ and does not change with a fourfold longer horizon, except for $K=15$, where the state itself grows from 255 to 317 tokens because the shorter background does not touch all fifteen keys. $B^\star$ is the first point of a doubling grid at which a selected accuracy crosses 0.85, so it is a noisy point estimate. In 2{,}000 pair resamples its most frequent value is the reported one in every cell and it otherwise moves by one grid step, except where a resample does not reach 0.85 on the grid at all: 30 percent of resamples at $K=3$ and 64 events, and at most 1.25 percent elsewhere. The equality across horizons is therefore a statement about these point estimates, not a test of equal thresholds. Above its threshold the state memory reads at 0.86 to 0.98. Recency stays between 0.50 and 0.65 at every budget, $K$ and $T$, because the window it keeps is a fixed fraction of a history that is mostly irrelevant. Both outcomes were pre-registered.

Keeping the last applied event of each component reaches the threshold at 128 tokens for $K=1$ at both horizons and for $K=3$ at 256 events only, and for $K\in\{7,15\}$ it stays at 0.74 to 0.83 even at the largest budget. The memory holds the right evidence. The reader has to apply the policy to raw lines, the step it performs poorly in \Cref{sec:exp-decay}. \paragraph{Bits and tokens.} The converse counts bits, about 110 to 281 for this benchmark, while the budgets above count tokens of English. A compact code that writes each component as a short string such as \texttt{kestrel>ledger:G2}, with a fixed legend outside the budget, halves the cost from about 19.5 to about 10.5 tokens per component. Readers exploit it only for the smallest state, $K=1$ at 64 events, where it reaches 0.85 at 32 tokens. Even with the whole compact state in the budget, balanced accuracy stays at 0.71 to 0.85 for Qwen 14B and 0.75 to 0.85 for phi-4, against 0.86 to 0.98 for the natural rendering when it fits, so the pre-registered threshold test fails for the compact code. The rate a memory needs in practice is set by the cheapest encoding the reader decodes reliably, and for these readers that encoding carries roughly twenty tokens per component, far above the few bits the converse requires.

The model-maintained ledger does not follow the state memory down to its threshold. On the same controlled backgrounds it stays at or below 0.65 in all eight cells checked ($K\in\{3,15\}$, $T\in\{64,256\}$, $B\in\{64,256\}$) and drops with the horizon. At $K=3$, $T=256$ and $B=256$ the full state needs about 83 tokens and the ledger writes 80, yet only 11 of 80 final ledgers mention the queried key and balanced accuracy is 0.51, against 0.98 for the replayed state. The failure does not depend on the budget or on the size of the state.

A first sweep with an uncontrolled background, in which memberships and windows kept the state growing with $T$, is reported in \Cref{app:x4a} together with a post-hoc analysis by the fraction of the state that fits the budget.

\subsection{Predicting a budgeted memory}
\label{sec:exp-budget}

The bit thresholds of \Cref{sec:theory} lie far below every token budget a reader can use. In units of decodable components, \Cref{prop:budget} gives a prediction that does bind. A read-time projection, which holds the exact state but shows the reader only a fraction $f$ of its rendered lines, dropped at random or keeping the most recently touched ones, should be read with balanced accuracy $\tfrac12+(\mathrm{BA}(1)-\tfrac12)(1-2\bar P_e(f))$. The factor $1-2\bar P_e(f)$ depends only on the generator and the eviction rule, and we computed it and registered it before running any reader (\Cref{tab:budget}). The only number taken from the run is the reader's accuracy on the full state in each cell.

\begin{figure}[t]
\centering
\begin{tikzpicture}\begin{axis}[width=0.47\linewidth,height=5cm,xlabel={fraction of state lines kept},ylabel={balanced accuracy},ymin=0.45,ymax=1.0,grid=major,title={\small random eviction},legend style={font=\scriptsize,at={(0.5,-0.32)},anchor=north,legend columns=2}]
\addplot[blue,thick,no markers] coordinates {(0,0.5000) (0.125,0.5357) (0.25,0.5756) (0.375,0.6199) (0.5,0.6772) (0.625,0.7329) (0.75,0.7924) (0.875,0.8566) (1.0,0.9250)};
\addlegendentry{Qwen2.5-14B predicted}
\addplot[blue,only marks,mark=*,mark size=1.5pt] coordinates {(0,0.5000) (0.125,0.5625) (0.25,0.5750) (0.375,0.7375) (0.5,0.6750) (0.625,0.7875) (0.75,0.8000) (0.875,0.8375) (1.0,0.9250)};
\addlegendentry{Qwen2.5-14B observed}
\addplot[red,thick,no markers] coordinates {(0,0.5000) (0.125,0.5315) (0.25,0.5667) (0.375,0.6058) (0.5,0.6564) (0.625,0.7055) (0.75,0.7580) (0.875,0.8146) (1.0,0.8750)};
\addlegendentry{phi-4 predicted}
\addplot[red,only marks,mark=*,mark size=1.5pt] coordinates {(0,0.5000) (0.125,0.5750) (0.25,0.6125) (0.375,0.6375) (0.5,0.7125) (0.625,0.8125) (0.75,0.7875) (0.875,0.8375) (1.0,0.8750)};
\addlegendentry{phi-4 observed}
\end{axis}\end{tikzpicture}
\hfill
\begin{tikzpicture}\begin{axis}[width=0.47\linewidth,height=5cm,xlabel={fraction of state lines kept},ylabel={balanced accuracy},ymin=0.45,ymax=1.0,grid=major,title={\small recency eviction},legend style={font=\scriptsize,at={(0.5,-0.32)},anchor=north,legend columns=2}]
\addplot[blue,thick,no markers] coordinates {(0,0.5000) (0.125,0.5106) (0.25,0.5319) (0.375,0.5319) (0.5,0.5425) (0.625,0.5531) (0.75,0.5744) (0.875,0.6594) (1.0,0.9250)};
\addlegendentry{Qwen2.5-14B predicted}
\addplot[blue,only marks,mark=*,mark size=1.5pt] coordinates {(0,0.5000) (0.125,0.5125) (0.25,0.5500) (0.375,0.5500) (0.5,0.5625) (0.625,0.5750) (0.75,0.6125) (0.875,0.6875) (1.0,0.9250)};
\addlegendentry{Qwen2.5-14B observed}
\addplot[red,thick,no markers] coordinates {(0,0.5000) (0.125,0.5094) (0.25,0.5281) (0.375,0.5281) (0.5,0.5375) (0.625,0.5469) (0.75,0.5656) (0.875,0.6406) (1.0,0.8750)};
\addlegendentry{phi-4 predicted}
\addplot[red,only marks,mark=*,mark size=1.5pt] coordinates {(0,0.5000) (0.125,0.5125) (0.25,0.5375) (0.375,0.5375) (0.5,0.5500) (0.625,0.5875) (0.75,0.6125) (0.875,0.7375) (1.0,0.8750)};
\addlegendentry{phi-4 observed}
\end{axis}\end{tikzpicture}
\caption{Budgeted filters on the controlled backgrounds with $K=15$ keys at $T=256$ events, 40 pairs per point. Lines are the predictions of \Cref{prop:budget}, fixed before the run except for the end point at $f=1$. Points are the measured balanced accuracies.}
\label{fig:budget}
\end{figure}
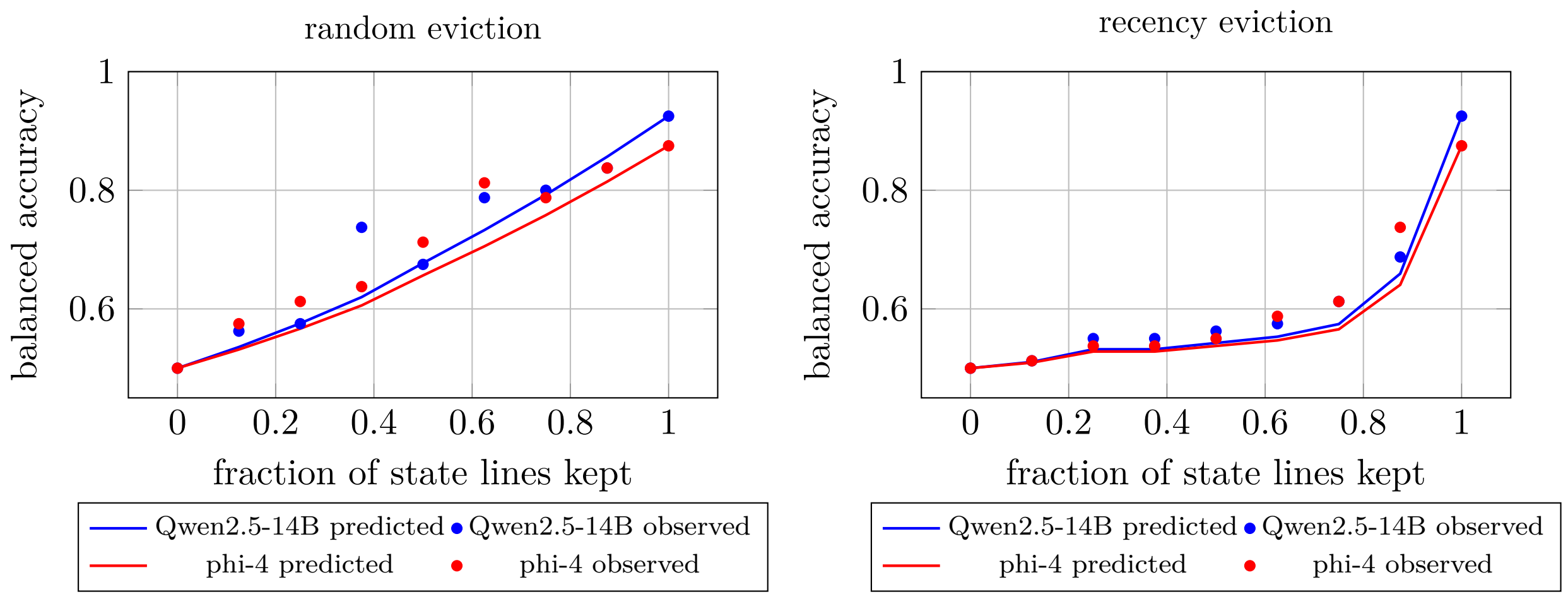

The prediction holds for both readers on all six backgrounds ($K\in\{3,7,15\}$, $T\in\{64,256\}$). Over the 84 interior cells the mean absolute gap between measured and predicted balanced accuracy is 0.026 for Qwen2.5-14B and 0.030 for phi-4, and 99 percent of the cells are within 0.10 (\Cref{fig:budget}). With nothing kept both readers sit at exactly 0.50. Under random eviction balanced accuracy rises on a straight line in $f$ whose slope matches $\mathrm{BA}(1)-\tfrac12$ within 0.08 and whose intercept is within 0.02 of one half, in every cell with $K\in\{7,15\}$. Keeping the most recent lines is worse than keeping random ones, as the theory predicts, because the deciding component of a long history is often an old one: at $K=15$ and $T=256$ the mean advantage of random eviction over $f\in[0.25,0.75]$ is $+0.15$ ($[0.09,0.20]$) and $+0.15$ ($[0.08,0.22]$). All four criteria were registered in advance with their thresholds. For read-time projections, a calibrated component-eviction model of the reader therefore predicts the measured curve from one measured number per reader and background, in units of components the reader can decode.

\paragraph{A bounded recursive store.} A projection keeps the full state hidden and limits only what the reader sees, so it does not test the recursive constraint of \Cref{def:memory}. A store of at most $C$ component records that applies each event by the exact rule to what it holds, and deletes what it evicts, does (\Cref{def:store}). An evicted higher-rank withdrawal can no longer block a later lower-rank grant, and a component written early must survive every later insertion. Before any reader ran we computed and registered its factor $1-2\bar P_e$ on fresh streams (\Cref{tab:store} and row R5-1 of \Cref{app:prereg}). Under random eviction it falls far below the projection at long horizons: at $K=15$ and $f=\tfrac12$ it is 0.36, 0.11 and 0.05 at 64, 256 and 1{,}024 events, against 0.43, 0.39 and 0.45 for the projection with the same capacity. A store below the size of the state therefore has a retention horizon of its own, set by its capacity and the rate of insertions, while a store at or above the state size has none. Least-recently-touched eviction stays within 0.05 of its projection in six of the seven cells and within 0.075 in the seventh ($K=7$, 256 events), so the recency curve above already describes an irreversible store of that kind. Read by Qwen2.5-14B and phi-4 on fresh streams, with $\mathrm{BA}(1)$ measured on separate calibration seeds, the store follows its registered prediction in all five registered tests for both readers. Over the 147 interior cells the mean gap between measured and predicted balanced accuracy is 0.013 and 0.018, and every cell is within 0.10. At $K=15$ and 256 events under random eviction, measured accuracy averaged over $f\in[0.25,0.75]$ is 0.56 and 0.57, below the projection's prediction of 0.67 by $+0.11$ ($[0.08,0.14]$) and $+0.10$ ($[0.05,0.13]$), as the store's own prediction says. At the same capacity fractions, with $C$ a fixed fraction of the mean state size (12.8 components at 64 events and 16.3 at 256), it falls from 0.72 at 64 events to 0.56 and 0.57 at 256, a drop of $+0.16$ ($[0.10,0.21]$) and $+0.15$ ($[0.07,0.21]$), and to 0.51 and 0.54 at 1{,}024. Under least-recently-touched eviction measured accuracy stays within 0.014 of the projection on average over interior capacities at $K=15$ and 256 or 1{,}024 events. On the full-background streams, a bounded store built from Qwen 14B parses reaches 0.80 and 0.84 with 32 records, against 0.80 and 0.84 for the unbounded filter, and falls to 0.55 to 0.61 with 8 or 16. A rate-limited recursive filter therefore behaves as the theory says: exact above the size of the state, and below it forgetting on a time scale set by its capacity.

\subsection{What each memory stores and what it costs}
\label{sec:exp-cost}

Memories differ in what they keep between events, what they keep as an archive, what they spend when an event arrives and what the reader finally sees, and a comparison at equal reader input is not a comparison at equal storage or equal total cost (\Cref{tab:resources}).

\begin{table}[t]
\centering
\small
\begin{tabular}{p{3.6cm}p{2.6cm}p{1.6cm}p{2.0cm}p{2.6cm}}
\toprule
memory & persistent store & raw archive & model calls per event & reader sees \\
\midrule
full log & all lines & yes & 0 & all lines \\
BM25, recency and relevance & all lines & yes & 0 & top lines, 512 tokens \\
dependency retriever (R5-3) & all lines & yes & 0 & lines naming the query's components \\
LLM ledger, summary & text, 256 tokens & no & 1 per 16 & the text \\
model-applied table (R2-B1) & table, at most 512 generated tokens & no & 1 per 16 & the table \\
Mem0-style fact store (R2-B3) & facts, unbounded & no & 1 per 16 & the facts, cut to 512 tokens \\
code-kept tables (R3-2) & table, unbounded & no & 1 per 1 or 16 parsed updates & the table \\
StateMem, Mem0 (R5-3) & units or facts, unbounded & no & 1 to 2 & a state block or the top 10 facts \\
parsed filter & exact state & no & 1 & the state, uncut or cut to 512 tokens \\
read-time projection (\Cref{tab:recursive,tab:budget}; \S\ref{sec:exp-rate}) & exact state & no & 0 or 1 & a cut of the state \\
bounded recursive store (R5-1) & $C$ records & no & 0 or 1 & the $C$ records \\
\bottomrule
\end{tabular}
\caption{What each memory keeps and what it costs. ``0 or 1'' model calls: none when the exact events are used, one parser call per event when they are parsed.}
\label{tab:resources}
\end{table}

\paragraph{Read-time projections.} The state memories under a budget in \Cref{tab:recursive}, the rate sweep of \Cref{sec:exp-rate} and the budget test of \Cref{sec:exp-budget} recompute the exact state from the whole history and cut it only when the reader reads it. In \Cref{tab:recursive} and the rate sweep the cut keeps the first $B$ reader tokens of the rendering, which lists direct records alphabetically and then memberships, group grants and windows, so a cut removes windows and group lines first and may end inside a line. These memories test the reader model of \Cref{prop:budget}. They do not test a store that forgets, which is the bounded recursive store above.

\paragraph{Equal budgets.} The filter state is not always within the 512 tokens given to the retrieval and fact-store memories of \Cref{tab:buildable}. At 256 events 24 of the 80 states exceed 512 reader tokens (mean 501, maximum 569). Cut to 512 tokens, the parsed filter answers 0.79 with Qwen2.5-14B, 0.81 with phi-4 and 0.86 with Qwen2.5-32B, against 0.80, 0.84 and 0.89 uncut, and the same cut explains why the replayed state reads 0.79 at 512 tokens in \Cref{tab:recursive} and 0.83 uncut in \Cref{tab:buildable} on the same streams.

\paragraph{Writing and reading.} The token ratios of \Cref{tab:reasoning,tab:strong} are read-time reductions after the state has been built. Building it costs one parser call per event. With the prompt of \Cref{app:policy} a parser call costs 438 prompt tokens, of which 410 are a prefix shared by every call, and 18 generated tokens on average, so writing a 256-event log costs about 117{,}000 tokens, or about 12{,}000 when the shared prefix is cached. Reading that log once costs the Qwen readers about 7{,}000 tokens and reading the state about 740. For $Q$ queries against the same history the filter costs $C_{\mathrm{write}}(T)+Q\,C_{\mathrm{read,state}}$ and the log $Q\,C_{\mathrm{read,log}}(T)$, so the filter is cheaper in tokens after about 19 queries without prefix caching and about 2 with it. Per token, a call to the 14-billion-parameter parser is also cheaper than a call to a 120-billion-parameter reader. A memory that is read once is cheaper as a log. One that is queried repeatedly, or read by a large model, is cheaper as a state.

\section{Related work}
\label{sec:related}
\IfFileExists{related/related_work.tex}{

\paragraph{Data-rate theorems and real-time coding.}
Our problem belongs to a long line of work on estimation and control through finite-rate links.
The data-rate theorem states that a linear plant can be stabilized or tracked only if the channel rate exceeds the sum of the logarithms of the magnitudes of its unstable eigenvalues~\cite{wong1999systems,nair2004stabilizability,tatikonda2004control}.
Later work extended the threshold to noisy channels, where Shannon capacity is not the right quantity and anytime capacity takes its place~\cite{tatikonda2004noisy,sahai2006necessity}, to nonlinear plants through topological feedback entropy~\cite{nair2004topological}, and to state estimation with a prescribed accuracy~\cite{matveev2007analogue,liberzon2018entropy}.
Surveys and monographs collect these results~\cite{nair2007feedback,matveev2009estimation}, and rate--cost tradeoffs refine them for LQG control~\cite{kostina2019ratecost}.
A second line studies causal coders that update a finite memory one symbol at a time.
Witsenhausen showed that an optimal real-time coder for a Markov source needs only the current symbol and the decoder memory~\cite{witsenhausen1979structure}, and Walrand and Varaiya identified the decoder posterior as a sufficient statistic~\cite{walrand1983optimal}.
These structural results were generalized to sequential decision formulations of real-time communication~\cite{teneketzis2006structure,mahajan2009optimal} and complemented by causal rate--distortion theory~\cite{neuhoff1982causal}.
Hellman and Cover gave an early sharp link between the number of memory states and the achievable error of a hypothesis test~\cite{hellman1970learning}.
The same idea appears in planning as the belief state of a POMDP~\cite{kaelbling1998planning} and its learned approximation, the approximate information state~\cite{subramanian2022approximate}.
We use this body of work as our foundation.
The memory update $M_t=U(M_{t-1},O_t)$ is a causal finite-memory coder, and the quantity to preserve is the decision $f_Q(S_T)$ rather than the source itself.
What changes in the agent setting is that the latent process is a finite-state machine driven by events that supersede one another.
Supersession is what makes the sufficient statistic independent of the history length $T$.
To our knowledge, none of the work on LLM memory or context compression that we surveyed connects to data-rate theorems or real-time coding.

\paragraph{Filtering and estimation from imperfect observations.}
For a hidden Markov model the exact Bayes filter is the forward recursion~\cite{rabiner1989tutorial,ephraim2002hidden}, and Wonham derived the corresponding filter for finite-state Markov chains in continuous time~\cite{wonham1964some}.
Estimation from coarse observations is studied with finite memory~\cite{hellman1970learning}, under the data-rate constraints discussed above, and with quantized measurements, for example Kalman filtering from the signs of innovations~\cite{ribeiro2006soikf}.
Event-triggered estimation sends a measurement only when it is informative enough to justify the communication cost~\cite{astrom2002comparison,wu2013event}.
Our noisy parser is an observation channel in this sense.
Each line is parsed independently and may be dropped, flipped, or affected by a rank-replacement error, and an exact transition rule then updates the state.
The plug-in filter we analyze applies the exact rule to the noisy observations as if they were correct.
A Bayes filter over the supersession automaton, with the parser errors as the emission model, is the HMM forward recursion.
We run it against the plug-in filter in \Cref{sec:exp-parsers}.

\paragraph{Rate--distortion views of memory and context.}
Several recent papers treat the context of a language model as a source to be compressed.
Nagle et al.\ cast prompt compression for black-box models as a rate--distortion problem and computed the optimal distortion--rate curve, separating the query-aware and query-agnostic cases~\cite{nagle2024fundamental}.
DeMem measures memory quality by the loss in decision quality and derives a forgetting boundary, a memory--distortion frontier, and an information lower bound of Fano type~\cite{zou2026remember}.
Its contexts are drawn i.i.d., the memory is a one-shot partition into $K$ states, and the current query is not charged to the budget.
Colaco and Lahjouji unify KV-cache eviction, prompt pruning, bounded recurrent state, and agent memory consolidation as one compaction problem with a Fano-type lower bound~\cite{colaco2026keep}.
They analyze a single compaction event and state the accumulation of error under repeated compaction as a hypothesis.
Kim formulates online KV-cache compression as sequential Wyner--Ziv coding and characterizes suffix-only cache policies, leaving open whether recurrent summaries can do better~\cite{kim2026polynomial}.
Our analysis differs from all three in being recursive and across time.
The budget is a per-step rate, the bounds hold uniformly in $T$, and our separation concerns order-blind retention rules, which ignore the time order of events.
A one-shot encoder can see the order of the events it compresses.
What a one-shot formulation cannot express is the update constraint, that $M_t$ is computed from $M_{t-1}$ and $O_t$ alone.
For finite-state decision queries the question Kim leaves open has a classical answer through the Myhill--Nerode theorem and streaming space bounds~\cite{nerode1958linear,alon1996space}.
The same equivalence of histories underlies predictive state representations~\cite{littman2001psr} and the causal states of computational mechanics~\cite{shalizi2001computational}, and the $\epsilon$-transducer of Barnett and Crutchfield is its general form for input--output processes~\cite{barnett2015computational}.
Our achievability result instantiates this answer for agent memory and does not resolve the general question left open in~\cite{kim2026polynomial}.
Our converse uses the same Fano argument as~\cite{zou2026remember,colaco2026keep} and adds the recursive, uniform-in-$T$ form.
The fooling-set argument behind our space lower bound is the standard technique for space lower bounds of streaming algorithms~\cite{alon1996space}.

\paragraph{Long contexts and state tracking.}
Language models use long inputs unevenly.
Accuracy depends on where the evidence sits~\cite{liu2024lost} and falls as the input grows even when the task is unchanged~\cite{levy2024same} or retrieval is perfect~\cite{du2025context}.
RULER defines the effective context length of a model as the longest input at which its accuracy stays above a fixed threshold~\cite{hsieh2024ruler}.
Our operational horizon $H_\Delta$ is also a threshold, but on the gap between reading the state and reading the log rather than on the accuracy of one reader.
Closest to our observation, PI-LLM streams repeated updates to the same keys and asks for the final values.
Accuracy declines log-linearly toward zero as interference accumulates, and the errors mostly return overwritten values~\cite{wang2025unable}.
Entity tracking degrades toward chance as operations accumulate~\cite{kim2023entity}, and mechanistic analysis shows that models aggregate the relevant facts at the query token rather than updating a state incrementally, with removals implemented by a fragile suppression signal~\cite{tang2026entities}.
Memory benchmarks report the same trend at the scale of conversations~\cite{wu2025longmemeval,tavakoli2026beyond}.
On the theory side, transformers can simulate automata through shortcuts that are brittle out of distribution~\cite{liu2023transformers}, state tracking is limited for both transformers and state-space models~\cite{merrill2024illusion}, and attention can be modeled as a bandwidth-limited channel~\cite{schnabel2025lost}.
Our measurement is consistent with these findings and adds two elements.
First, each test case is one half of a pair that shares background, event positions, and wording and differs only in the order of the target events or in one field, so the correct answers swap.
Second, we separate a shift in the decision threshold from a loss of discrimination using AUC and the best post-hoc threshold.
At $T=256$ the loss of discrimination dominates, which a threshold shift cannot explain.

\paragraph{Benchmarks for evolving state.}
A growing set of benchmarks asks whether memory reflects the current state rather than a superseded one.
StateMemBench renders symbolic programs of typed state operations into dialogue and grades whether an answer reflects the current or the superseded state~\cite{fan2026can}.
The same paper proposes a memory method, StateMem, which we discuss with structured state tracking below.
STALE tests implicit conflicts in which a later observation invalidates an earlier memory without explicit negation~\cite{chao2026stale}.
Memora penalizes answers that rely on invalidated memories~\cite{uddin2026recall}, MemoryAgentBench includes a fact-consolidation task for selective forgetting~\cite{hu2026evaluating}, and MEMAUDIT evaluates memory writing before the query is known, including the preservation of validity state~\cite{bhargava2026memaudit}.
OrgAccess evaluates role-based access-control reasoning but has no temporal sequence of grants and revocations~\cite{sanyal2026orgaccess}.
Our generator complements these benchmarks with paired order counterfactuals, families for restoration, decoys, precedence, expiry, implicit membership changes, and conjunctions, and a reader validity gate that checks the reader on the true current state before any memory is evaluated.
We use the benchmark as an instrument for testing the theory rather than as a contribution in its own right.
None of these benchmarks, ours included, draws its state-bearing events from real logs.
GitHub issue timelines, available at scale through GHTorrent~\cite{gousios2012ghtorrent}, are a candidate source for future work because labels, assignees, and status are set and later superseded in time, although they carry no authority ranks.

\paragraph{Memory systems and compression.}
Agent memory systems manage a hierarchy of context~\cite{packer2023memgpt}, retrieve from a memory stream by recency, importance, and relevance~\cite{park2023generative}, or summarize recursively~\cite{wang2025recursively}.
A-Mem stores each interaction as a note with keywords, tags, and a context written by an LLM, links it to similar notes, and lets the new note rewrite the context of its neighbours~\cite{xu2025amem}.
AriGraph builds a semantic knowledge graph with episodic links from the observations of a text-game agent, and an LLM removes the triplets that a new observation makes outdated~\cite{anokhin2025arigraph}.
Neither ranks sources. A-Mem never marks a note as superseded, and in AriGraph the newer observation always replaces the older triplet.
MemAgent overwrites a fixed-length memory chunk by chunk and trains the update with reinforcement learning~\cite{yu2026memagent}, and MEM1 learns a constant-size internal state~\cite{zhou2026mem1}.
Both are recursive filters in form.
They differ from our setting in that the query is known while the memory is written, and neither paper analyzes the rate required.
Prompt compression removes tokens by perplexity or a learned classifier~\cite{jiang2023llmlingua,jiang2024longllmlingua,pan2024llmlingua2}, and KV-cache methods keep heavy-hitter or recent tokens~\cite{zhang2023h2o,xiao2024efficient} or quantize the cache~\cite{liu2024kivi,hooper2024kvquant}.
These retention rules do not represent which events supersede which.
Those whose scores ignore the time order of events are order-blind in our sense, and our separation result applies to them.
A separate line keeps procedural lessons rather than world state. Self-Direction distils each corrected mistake into a low-rank adapter and later loads the adapter whose source task is most similar to the current one~\cite{kansal2026selfdirection}. Such lessons accumulate rather than supersede one another, which is why selection by similarity suffices there. A lesson made obsolete by a later change would still be selected, since nothing in that retrieval records which lesson replaced which.

\paragraph{Structured state tracking.}
The filter we build, in which an LLM parses each line into a typed update and an explicit transition rule applies it, follows a standard design.
Task-oriented dialogue state tracking maintains slot values that are updated turn by turn~\cite{budzianowski2018multiwoz,wu2019transferable}, event-sourced systems rebuild state by replaying a log of updates, and knowledge representation formalizes how events start and end facts in the event calculus~\cite{kowalski1986logic} and how roles are enabled and disabled over time in temporal role-based access control~\cite{bertino2001trbac}.
Letting the LLM write the transition rule as code follows program-aided reasoning~\cite{gao2023pal}.
Structured agent memories apply explicit add, update, and delete operations~\cite{chhikara2025mem0} or invalidate the edges of a temporal knowledge graph when new facts contradict them~\cite{rasmussen2025zep}.
We have not run Graphiti, so the following is an argument and not a measurement.
Because this invalidation orders facts by time, it fails on the blocked chains of our recency proposition, where a later event from a lower-authority source must not override an earlier higher-authority one, unless authority is modelled explicitly.
The closest published method is StateMem~\cite{fan2026can}.
One LLM call per turn parses the turn into typed state units kept in a persistent store, and the same call flags existing units that the turn supersedes.
A deterministic re-check then marks every unit that depends on a changed unit.
Only a later turn can supersede an earlier unit, so conflicts are resolved by recency, and the store records the turn and speaker of each unit without ranking sources.
StateMem differs from our parsed filter in one place: the LLM, not a rule, decides supersession.
We compare the two in \Cref{sec:exp-buildable}.
Table~\ref{tab:methods} summarizes these designs.
Our contribution is therefore not the architecture.
It is the measurements, the taxonomy of retention rules into order-blind, position-based, and last-applied rules together with its bound, and the evidence that the model fails specifically at the transition step.

\paragraph{Evidence against bounded memory.}
Patel reports that replacing the full context with a bounded, self-maintained memory lowers accuracy on the knowledge-update subset of LongMemEval, that accuracy falls further as the conversation grows, and that a proportionally larger memory does not recover it~\cite{patel2026supersede}.
This finding cuts against our claim that a recursive memory can make accuracy independent of $T$.
The memory in that study is free-form text with a fixed character budget and no structure by key.
Our theory predicts that such a memory fails once the number of live keys exceeds what the budget can hold, and that a ledger keyed by entity with explicit supersession keeps its accuracy as long as the budget covers the per-key cost.
Our experiments test this prediction directly.
}{\todo{related work from related session}}
\IfFileExists{related/method_table.tex}{
\begin{table}[t]
  \centering
  \small
  \setlength{\tabcolsep}{4pt}
  \caption{Design of memory methods for evolving state. Calls are LLM calls made while writing the memory, per event (turn, message, observation, or chunk).}
  \label{tab:methods}
  \begin{tabular}{@{}>{\raggedright\arraybackslash}p{2.6cm}
                  >{\centering\arraybackslash}p{1.3cm}
                  >{\raggedright\arraybackslash}p{2.7cm}
                  >{\centering\arraybackslash}p{1.6cm}
                  >{\raggedright\arraybackslash}p{3.3cm}
                  >{\raggedright\arraybackslash}p{2.9cm}@{}}
    \toprule
    Method & Query-blind writing & Who decides supersession & Source\newline authority & Persistent store bound & LLM calls per event \\
    \midrule
    StateMem~\cite{fan2026can} & \checkmark & LLM flags, newer turn wins & $\times$ & none, superseded units kept & 1 \\
    Mem0~\cite{chhikara2025mem0} & \checkmark & LLM (delete or update) & $\times$ & none & $1+{}$one per extracted fact \\
    Zep/Graphiti~\cite{rasmussen2025zep} & \checkmark & LLM detects, time resolves & $\times$ & none, invalidated edges kept & several (extraction, resolution, temporal) \\
    A-Mem~\cite{xu2025amem} & \checkmark & none, notes are never superseded & $\times$ & none, one note per event & $2+{}$one per linked neighbour \\
    AriGraph~\cite{anokhin2025arigraph} & \checkmark & LLM selects outdated triplets & $\times$ & none & 2 \\
    MemAgent~\cite{yu2026memagent} & $\times$ & LLM rewrites memory & $\times$ & 1024 tokens & 1 \\
    MEM1~\cite{zhou2026mem1} & $\times$ & LLM rewrites state & $\times$ & one internal state, size learned & 1, shared with acting \\
    \midrule
    Parsed filter (ours) & \checkmark & rule & \checkmark & one entry per live component & 1 \\
    \bottomrule
  \end{tabular}

  \vspace{2pt}
  {\footnotesize\raggedright StateMem records the turn and speaker of each unit but does not rank sources. Zep invalidates an edge when an LLM finds a contradiction, and the newer fact always wins. MemAgent and MEM1 write their memory with the question or task in the prompt.\par}
\end{table}
}{}

\section{Discussion}
\label{sec:discussion}

\paragraph{What to build.} The results point to a division of labour, and \Cref{sec:exp-buildable} measures it. A language model is good at reading one event and saying what it changes. It is poor at holding the consequences of hundreds of such changes in a long context, and poor at maintaining a compact record of them when asked to. The replayed state and the parsed filter succeed because the structure is enforced by the memory, not requested from the model. The last-applied rule supplies the right evidence but leaves the reader to apply the policy to raw lines, and it falls short of the state on long streams. A practical memory therefore parses each incoming event into a typed update, applies it with an explicit transition rule that knows supersession, rank and expiry, and shows the reader the resulting state. The budget such a memory needs grows with the number of tracked keys, which \Cref{sec:exp-rate} measures directly. Retention that ignores time order cannot tell which retained line is still in force (\Cref{thm:separation}). Retention that uses time but not authority fails behind blocked events (\Cref{prop:recency}). Retention that executes the rule is exact with a budget set by the number of components (\Cref{prop:aware}), and the two weaker classes are not.

\paragraph{Long histories.} The logs are long in events, not in tokens. A log of 256 events is about 7{,}000 tokens, well inside the native window of every reader (16{,}384 tokens for phi-4 and 32{,}768 for the Qwen models), and the reader decay model of \Cref{sec:theory} is stated per line. At 1{,}024 events the log is about 23{,}000 to 27{,}500 tokens, still inside the window of every reader of \Cref{tab:strong}, so their decline there is not truncation. In this benchmark longer logs hurt mainly through the amount of unrelated material, and the number of updates to the queried key has no detectable effect within the tested range. This differs from the interference account of repeated key overwrites \citep{wang2025unable}, and the difference may come from the task. Overwrite benchmarks ask for a value, while ours asks for a decision under a policy that combines several components. The reinstatement and decoy failures suggest a second mechanism on top of dilution: prohibitions and alarm words act as strong cues that a later reinstatement does not cancel.

\paragraph{Domain of validity.} Both domains are synthetic and their rules are stated in the prompt. Events come from templates, and one paraphrased variant written by a model, not by people, is the only test of other wording. No line refers to another line, so the context-free parsing that the filter relies on is never tested against cross-line references, and every event fits the update schema. An event outside the schema would be dropped by the parser, which is the drop error of the noisy-parser analysis, not a safe default. Readers are open models between 7 and 120 billion parameters, scored mainly by the likelihood of a fixed answer format, without tool use. The readers above 32 billion parameters are measured only in the test of \Cref{tab:strong}. Two reasoning-trained 32B models are measured on the 80 filter streams, and QwQ-32B also on the decay seeds. Proprietary frontier models as readers, trained recursive memories such as MemAgent, KV-cache eviction methods and real operational logs are not measured. The transition rule of the filter is written by people. A frontier coding agent wrote one that matches it after one clarification of the update format. An open 32B code model wrote a rule within the registered tolerance on the first domain only after that clarification, and that rule still fails rank precedence. On the simpler second domain it did so after selection among samples. The theory is exact for the access process defined in \Cref{sec:theory} and its rate results assume bounded leases. The reader decay model is a fitted description with stated assumptions, not a theorem. phi-4 both paraphrased the logs and read them. Real operational logs are not tested, and issue-tracker timelines, which supersede by time but have no authority ranks, are a natural first source. A state with many keys can be rendered as the query cone, the queried pair, its groups and its window, a handful of lines, if the query is known when the memory is read. Persistent rank errors behind blocked events could be corrected by re-parsing the lines of a key when a query depends on it.

\bibliographystyle{plainnat}
\bibliography{refs}

\appendix
\section{Policy text and prompt}
\label{app:policy}
\begin{quote}\small\ttfamily
\textbf{Access policy.}

1. Sources are ranked: security\_office above access\_admin above ops\_console. An access event (grant, withdrawal or reconfirmation) changes the current decision only if its source ranks at least as high as the source of the decision it would replace.

2. A grant stated as valid through Event N no longer applies after Event N.

3. A service may write to a resource if it holds a current grant, or if it belongs to a group that holds a grant on that resource.

4. If a resource accepts writes only while its change window is open, a write needs both permission and an open window. Windows start closed.

5. Without an applicable grant the answer is DENY.
\end{quote}
\noindent The system prompt is the persona sentence followed by this policy. The persona is ``You are an automated decision and safety governance agent.'' unless stated otherwise. The user turn is ``Based strictly on the memory below, evaluate the incoming request.'' followed by a \texttt{[Memory]} block and a \texttt{[Request]} block, and the assistant turn is prefilled with \texttt{DECISION:}.

\section{Separation experiment with the LLM reader}
\label{app:x5}

\begin{table}[h]
\centering
\small
\begin{tabular}{llccccccccc}
\toprule
 & & \multicolumn{5}{c}{chain length $m$} & \multicolumn{4}{c}{blocked events $b'$}\\
\cmidrule(lr){3-7}\cmidrule(lr){8-11}
$b$ & rule & 5 & 9 & 17 & 33 & 65 & 2 & 4 & 8 & 16\\
\midrule
\multirow{7}{*}{2} & BM25 & 0.49 & 0.50 & 0.50 & 0.50 & 0.50 & 0.34 & 0.38 & 0.43 & 0.47\\
 & key mention & 0.40 & 0.44 & 0.48 & 0.49 & 0.48 & 0.30 & 0.39 & 0.43 & 0.46\\
 & salience & 0.49 & 0.50 & 0.51 & 0.50 & 0.50 & 0.44 & 0.48 & 0.49 & 0.49\\
 & gradient & 0.47 & 0.45 & 0.51 & 0.53 & 0.48 & 0.51 & 0.48 & 0.50 & 0.52\\
 & recency & 0.48 & 0.43 & 0.43 & 0.33 & 0.14 & 0.50 & 0.51 & 0.50 & 0.50\\
 & key recency & 0.06 & 0.02 & 0.06 & 0.02 & 0.03 & 0.50 & 0.50 & 0.50 & 0.50\\
 & last applied & 0.01 & 0.00 & 0.00 & 0.00 & 0.00 & 0.00 & 0.01 & 0.00 & 0.00\\
 & bound & 0.00 & 0.25 & 0.38 & 0.44 & 0.47 & -- & -- & -- & --\\
\midrule
\multirow{7}{*}{4} & BM25 & 0.49 & 0.50 & 0.50 & 0.50 & 0.50 & 0.39 & 0.50 & 0.46 & 0.48\\
 & key mention & 0.41 & 0.35 & 0.43 & 0.44 & 0.48 & 0.31 & 0.45 & 0.47 & 0.47\\
 & salience & 0.52 & 0.49 & 0.51 & 0.50 & 0.50 & 0.48 & 0.49 & 0.48 & 0.49\\
 & gradient & 0.49 & 0.46 & 0.48 & 0.45 & 0.53 & 0.48 & 0.47 & 0.50 & 0.48\\
 & recency & 0.43 & 0.41 & 0.33 & 0.21 & 0.08 & 0.50 & 0.50 & 0.50 & 0.50\\
 & key recency & 0.31 & 0.30 & 0.25 & 0.28 & 0.18 & 0.31 & 0.50 & 0.50 & 0.50\\
 & last applied & 0.01 & 0.01 & 0.00 & 0.00 & 0.00 & 0.01 & 0.01 & 0.00 & 0.00\\
 & bound & 0.00 & 0.00 & 0.25 & 0.38 & 0.44 & -- & -- & -- & --\\
\bottomrule
\end{tabular}

\caption{Error of Qwen2.5-14B reading the retained views of \Cref{sec:exp-separation}. The bound rows refer to every reader of an order-oblivious memory.}
\label{tab:separation-llm}
\end{table}

\section{Second-order retention criteria}
\label{app:curvature}
\label{sec:exp-curvature}

This appendix reports a negative result that the rest of the paper does not depend on: a second-order criterion for retaining KV-cache entries, tested against first-order scoring.

For a binary decision with margin $d(u)=z_{\rm DENY}-z_{\rm APPROVE}$ and $p=\sigma(d)$, the Fisher information of the decision distribution with respect to a memory representation $u$ is $F=p(1-p)\,gg^\top$ with $g=\nabla_u d$. The local KL divergence caused by a residual $e$ is
\begin{equation}
\tfrac12\, e^\top F e=\tfrac12\, p(1-p)\,(g^\top e)^2 ,
\end{equation}
the square of the first-order change of the margin. The trace of the diagonal block of token $i$ is $p(1-p)\|g_i\|^2$, and $p(1-p)$ is shared by all blocks of one query. Ranking memory blocks by Fisher trace is therefore ranking by gradient norm. Information beyond gradient magnitude can only come from the direction of the residual, from cancellation across blocks and from weighting across queries.

We tested these three sources on real KV-cache compression. The memory block of a prompt is prefilled in full precision, its keys and values in all 28 layers of Qwen2.5-7B are quantized (keys per channel over groups of 32 tokens, values per token, 2 to 4 bits, stochastic rounding), and only the request and label tokens attend to the quantized cache. On 200 episodes with 4 queries and 12 perturbations each, we predicted the true KL divergence on held-out episodes with seven predictors. The full rank-one quadratic is worse than the best first-order predictor by 6.1 percent (95 percent interval $[-9.3,-3.6]$). Averaging over queries gains 1.8 percent ($[-3.9,+6.6]$). Gradient magnitude times residual magnitude, which ignores direction, is within one percent of the per-block directional predictor. The linear prediction of the margin change has correlation 0.89 with the truth at 4 bits and 0.23 at 2 bits, where 23.5 percent of decisions flip.

\paragraph{Protocol.} Episodes come from a simpler generator with grants, withdrawals and reinstatements over 24 events and four queries per episode. The memory block is prefilled once in full precision. Its post-RoPE keys and values in all 28 layers of Qwen2.5-7B-Instruct are quantized in the KIVI layout (keys per channel over groups of 32 tokens, values per token) with 2, 3 or 4 bits and stochastic rounding. Seed 0 quantizes every memory line and seeds 1 to 3 quantize a random half of the lines. Only the request and the label tokens attend to the quantized cache, so the residual $e=\hat u-u$ is exact. Label logits are computed in single precision. The KL floor is $10^{-4}$, set from the margin noise between cached and one-shot forwards (up to 0.28). There are 200 episodes, 4 queries and 12 perturbations, 9600 rows in all. Even episode indices train and odd indices test.

\paragraph{Predictors.} With $c=p(1-p)/2$ and blocks $b$ indexed by token, layer and key or value:
$P_1=\|e\|^2$,
$P_2=c\,\|g\|^2\|e\|^2/D$,
$P_3=c\sum_b\|g_b\|^2\|e_b\|^2/D_b$,
$P_{3b}=c\,(\sum_b\|g_b\|\|e_b\|)^2$,
$P_4=c\sum_b(g_b^\top e_b)^2$,
$P_5=c\,(\sum_b g_b^\top e_b)^2$.
$P_6$ averages $P_5$ over the four queries and $P_7$ averages $(g^\top e)^2$ without $p(1-p)$. Each predictor enters an ordinary least squares fit of $\log$ KL on $\log P$ on the training episodes.

\begin{center}
\small
\begin{tabular}{lcccccccc}
\toprule
 & $P_1$ & $P_2$ & $P_3$ & $P_{3b}$ & $P_4$ & $P_5$ & $P_6$ & $P_7$\\
\midrule
per query, held-out MSE & 11.67 & 5.98 & 5.92 & 5.92 & 5.92 & 6.28 & & \\
per query, Spearman & 0.33 & 0.87 & 0.88 & 0.87 & 0.88 & 0.86 & & \\
query average, held-out MSE & 7.12 & 5.40 & 5.34 & 5.31 & 5.33 & & 5.21 & 6.43\\
\bottomrule
\end{tabular}
\end{center}

The pre-registered gate required $P_5$ or $P_6$ to reduce the held-out error of the best of $P_1$ to $P_4$ by at least 10 percent. The per-query reduction is $-6.1$ percent (95 percent bootstrap interval over episodes $[-9.3,-3.6]$) and the query-average reduction is $1.8$ percent ($[-3.9,6.6]$). Both upper limits are below 10 percent, so the mechanism claim was terminated as registered. $P_6$ beats $P_7$ by 18.8 percent ($[6.2,29.9]$), which only confirms that $\tfrac12p(1-p)$ is the right scale of the local KL divergence.

By bit width, the median KL is $1.2\times10^{-5}$, $8.7\times10^{-5}$ and $2.3\times10^{-3}$ at 4, 3 and 2 bits, the correlation between the true margin change and $g^\top e$ is 0.89, 0.74 and 0.23, and decisions flip in 1.8, 5.4 and 23.5 percent of rows. Restricted after the fact to 4 bits, $P_5$ improves on $P_4$ by 8.3 percent with a bootstrap interval of $[5.3,11.2]$. This slice was chosen after seeing the data and is not a registered test.

A first quantizer that grouped keys per token produced a relative key residual of 25 percent at 4 bits, because Qwen2.5 keys carry outlier channels with magnitude up to 420 against about 3 for typical channels, measured on the smoke-test episodes. The linearity check failed and the quantizer was replaced before the formal run.

\section{First rate sweep}
\label{app:x4a}

The first sweep restricted only the direct access keys of the background. Membership and window events were still drawn, so the state grew with $T$ at fixed $K$ (for $K=15$ from 223 to 370 tokens), and the extractive rule kept the newest event per key without regard to rank. The pre-registered horizon-independence test failed for the state memory in the $K=15$ cell only ($B^\star=256$ at 64 events against more than 512 at 256 events) and failed for the extractive rule throughout. Recency never reached 0.85. After the fact, grouping all cells by the fraction of the state that fits the budget gives balanced accuracies of 0.58, 0.66, 0.73, 0.86 and 0.94 for fractions below one quarter, one half, three quarters, below one, and the full state. The controlled sweep of \Cref{sec:exp-rate} was designed and registered after this result and before running.

\IfFileExists{theory/theory_appendix.tex}{
\section{Proofs for the cost of state}\label{app:theory-proofs}

\begin{proof}[Proof of Theorem~\ref{thm:achieve}]
If $h\equiv h'$ then $ho\equiv h'o$ for every $o\in\cO$, since every continuation of $ho$ is a continuation of $h$ with $o$ in front.
So $\Phi([h],o):=[ho]$ is well defined.
Taking the empty continuation in Definition~\ref{def:filter} shows that $f_q(S(h))$ depends on $h$ only through $[h]$, so $g(Z,q)$ is well defined and exact at every time.
\end{proof}

\begin{proof}[Proof of Theorem~\ref{thm:converse}]
(a) A memory cell $m$ that holds $c$ of the $h_i$ produces one fixed answer vector on $\Gamma$, because the decision after test $(w,q)$ is $g(U_w(m),q)$ with $U_w$ the composition of the updates along $w$.
The answer vectors of distinct $h_i$ differ, so at most one history in the cell is answered correctly on all of $\Gamma$.
Summing over at most $2^R$ cells gives at least $N-2^R\ge1$ histories with a wrong decision, made with certainty.

(b) Write the randomized memory as deterministic maps $U_t(m,o,\xi_t)$ and $g_t(m,q,\zeta_{t,q})$ driven by a sequence of random seeds $\omega=(\xi_t,\zeta_{t,q})_{t\le T,q\in\cQ}$, one realization per run, independent of the history.
Suppose $D_{\mathrm{wc}}(T)=0$. For each of the finitely many triples (history of length at most $T$, time, query), the set of seeds on which that decision is wrong has probability zero.
The union of these sets also has probability zero, so some realization $\omega^\ast$ answers every triple correctly.
Fixing $\omega^\ast$ gives a deterministic recursive memory with the same alphabet and zero error, which contradicts (a) when $2^R<N$.
The same argument works for countably many triples, so it extends to the anytime criterion over all horizons.
\end{proof}

\begin{proof}[Proof of Proposition~\ref{prop:count}]
\emph{Upper bounds.} Map each record to $\cD$ by $(0,1)\mapsto\bot$, records with $u<t$ to $\bot$, records with $u=t$ to $(1,\ast,0)$, and otherwise keep $(a,\ell)$ and replace $u$ by $\rho=u-t$.
Merged records behave identically on every continuation.
A rank-one withdrawal and $\bot$ are both replaced by every direct event on that key, are preserved by all other events, and both give direct-access indicator $0$ (the first disjunct of~\eqref{eq:answer}).
An expired record has rank $0$ and direct-access indicator $0$, exactly like $\bot$.
A record with $u=t$ has direct-access indicator $1$ now and rank $0$ from time $t+1$ on, whatever its $\ell$. The merge into $(1,\ast,0)$ is made only after the event at time $t$ has been processed, since during that step the record still holds rank $\ell$.
On the reduced states the update is well defined: age every residual by one (a residual that would become negative gives $\bot$), apply the event with the rule of Definition~\ref{def:process}, and merge again.
Hence the product of $\cD^{\cK}$ with the $C_{\mathrm{aux}}$ bits $(\mu,\gamma,\nu,\omega)$ is a recursive memory that answers \eqref{eq:answer} exactly, and $\NQ\le|\cD|^K2^{C_{\mathrm{aux}}}$.

\emph{Separating tests without leases.} We show that two product states that differ in one component are separated by a continuation of length at most $|\mathcal{G}|+2$. Let $L$ also denote the top rank.
For a key $\kappa=(p,r)$, first \emph{unmask} it: remove $p$ from every group and open $r$ (at most $|\mathcal{G}|+1$ events, which do not touch $d(\kappa)$).
Afterwards the answer to $q=\kappa$ equals the direct decision.
Querying now reads $a$.
If $a=1$ with rank $\ell$, a withdrawal of rank $\ell'$ is applied if and only if $\ell'\ge\ell$, so the tests ``withdraw at rank $\ell'$, then query'' for $\ell'=1,\dots,L-1$ read off $\ell$.
If $a=0$, grants at ranks $\ell'=1,\dots,L-1$ read off the rank of the withdrawal, with $\bot$ behaving as rank one.
For a membership bit $(p,g)$: add $\mathrm{gg}(g,r)$, withdraw $\kappa$ at rank $L$, open $r$ and remove $p$ from every other group, then query $\kappa$. The answer is the bit.
For a group grant $(g,r)$: withdraw $\kappa$ at rank $L$, remove $p$ from every other group, add $p$ to $g$, open $r$ and query. The answer is the bit.
For $\nu(r)$: grant $\kappa$ at rank $L$, close $r$ and query. The answer is $1-\nu(r)$.
For $\omega(r)$: issue $\mathrm{req}(r)$ unless already required in both states, grant $\kappa$ at rank $L$ and query. The answer is $\omega(r)$.
In each test the final answer depends only on the tested component, so two states that differ anywhere are separated.

\emph{Reachability without leases.} Any product state is produced by at most one event per component in any order, followed or preceded by $\varnothing$. This proves (a).

\emph{Lower bound with leases.} Put $s:=K_{\mathrm{tot}}$ and $k_u:=|\mathcal{G}|+1$ (the unmasking length).
Restrict every key to residuals $\rho\in\{k_u+1,\dots,\Lambda-s\}$ besides the $2L$ lease-free states, which gives $L(\Lambda-s-k_u+2)$ states per key.
They are reachable at time $s$: a lease issued at step $s'\le s$ with $\lambda=\rho+s-s'\le\Lambda$ has residual $\rho$ at time $s$, and each key is set once. For $t>s$ the null padding goes in front, so that residuals are not aged.
After unmasking, residuals have dropped by at most $k_u$ and are still at least one, so ranks are still visible to the withdrawal tests one step later.
Two different residuals $\rho<\rho'$ are separated by querying at time $t+\max(k_u,\rho+1)\le t+\rho'$, and finite residuals are separated from $\infty$ in the same way.
The tests of the auxiliary components reset $d(\kappa)$ by a rank-$L$ event without lease ($\lambda=\infty$) and are unaffected by leases.
\emph{Single key.} In the direct-only submodel there are no masks, and the four test types above separate all of $\cD$ (ranks of a live grant need $\rho\ge1$, which is why $(1,\ast,0)$ is one state).
\end{proof}

\begin{proof}[Proof of Theorem~\ref{thm:rd}]
Conditionally on $M_t$, run every test of $\Gamma_\kappa$ on an independent copy of the memory and let $\hat Z_\kappa$ be the state whose answer vector matches the outputs (arbitrary if none does). Then $Z\to M_t\to\hat Z$ is a Markov chain and $\hat Z_\kappa=Z_\kappa$ outside $E_\kappa$. Let $\delta_\kappa=\Prob(\hat Z_\kappa\ne Z_\kappa)\le\Prob(E_\kappa)$.
Since the $Z_\kappa$ are independent, $R\ge I(Z;M_t)\ge I(Z;\hat Z)\ge\sum_\kappa\bigl(H(Z_\kappa)-H(Z_\kappa\mid\hat Z_\kappa)\bigr)$.
Fano gives $H(Z_\kappa\mid\hat Z_\kappa)\le\varphi(\delta_\kappa)$ with $\varphi(\delta)=h(\delta)+\delta\log_2(n-1)$.
If $\bar\delta>1-1/n$ there is nothing to prove. Otherwise, $\varphi$ is concave on $[0,1]$ and increasing on $[0,1-1/n]$, and $\varphi(\delta)\le\log_2 n=\varphi(1-1/n)$ for all $\delta$, so $\sum_\kappa\varphi(\delta_\kappa)\le K\varphi(\bar\delta')$ with $\bar\delta'=K^{-1}\sum_\kappa\delta_\kappa\le\bar\delta$. Hence $\log_2 n-R/K\le\varphi(\bar\delta)$, which is~\eqref{eq:rd}.
\end{proof}

\begin{proof}[Proof of Proposition~\ref{prop:timeshare}]
The joint reduced state of $K'$ keys takes $n^{K'}\le2^R$ values and is updated exactly by the rule of Proposition~\ref{prop:count}, so tracked keys have no error. An untracked key has all of its tests answered correctly exactly when $Z_\kappa=z_0$, which has probability $1/n$ under the uniform law, and otherwise some test separates $Z_\kappa$ from $z_0$.
\end{proof}

\begin{proof}[Proof of Corollary~\ref{cor:fano}]
$H(A)-R\le H(A\mid\hat A)\le\sum_q H(A_q\mid\hat A_q)\le\sum_q h(P_{e,q})\le|\cQ|\,h(\bar P_e)$, using Fano for binary variables and concavity of $h$.
\end{proof}
}{}

\section{Evidence for the benchmark comparison}
\label{app:benchmark-evidence}
\IfFileExists{related/benchmark_evidence.tex}{
\begin{table}[ht]
  \centering
  \footnotesize
  \caption{Evidence for each cell of Table~\ref{tab:benchmarks}, taken from each paper's description of its own design.}
  \label{tab:benchmark-evidence}
  \begin{tabular}{@{}>{\raggedright\arraybackslash}p{3.2cm}>{\raggedright\arraybackslash}p{12.8cm}@{}}
    \toprule
    Benchmark & Evidence \\
    \midrule
    StateMemBench~\cite{fan2026can} &
      Answers come from replaying each symbolic program through a deterministic evaluator. Scenarios contain value updates, scoped exceptions, commitments, and retractions. Restoration of a retracted item, source authority, order-only minimal pairs, and an answerer given the correct state are not described. A scoped override that may lapse leaves expiry ambiguous. Memory stores are built before the probe. Set~A and Set~B share the probe structure and differ in length. Programs are rendered into dialogue by an LLM, and public data supplies vocabulary only. \\
    \addlinespace
    STALE~\cite{chao2026stale} &
      Scenarios are generated by an LLM, edited by experts, and graded by an LLM judge. Each instance contains one implicit conflict that invalidates an earlier belief, and the paper limits itself to one-shot transitions, so no state is restored. Source authority, expiry, order pairs, and a correct-state control are not described. Frameworks ingest the dialogue before probing. Contexts reach 150K tokens, but length is not an experimental factor. \\
    \addlinespace
    Memora~\cite{uddin2026recall} &
      Questions derive from simulated memory traces and are graded by LLM judges, so programmatic ground truth is ambiguous. The simulator introduces, updates, and invalidates entities. Users may reverse earlier statements, which leaves reinstatement ambiguous, and temporal constraints leave expiry ambiguous. Source authority, order pairs, and an oracle-memory condition are not described, and whether memory is written before the query is not stated. Weekly, monthly, and quarterly horizons are compared. Conversations are simulated. \\
    \addlinespace
    MemoryAgentBench~\cite{hu2026evaluating} &
      The selective-forgetting data reuse counterfactual edit pairs from MQuAKE, with the new fact placed after the original. The only conflict rule is to prefer later information. Reinstatement, expiry, order-swapped pairs, and a correct-state control are not described. Chunks are ingested incrementally before questions. Contexts of 6K, 32K, 64K, and 262K tokens are used. Inputs are reconstructed from existing long-context datasets to simulate multi-turn interaction. \\
    \addlinespace
    MEMAUDIT~\cite{bhargava2026memaudit} &
      Package optima are computed exactly by branch-and-bound with MILP certification. Tombstones mark superseded or deleted facts. Reinstatement, source priority, and order pairs are not described. A temporal-interval distribution leaves expiry ambiguous, and a reader audit in the appendix is not a correct-state control. Memory is written before future queries are known. Sweeps vary the budget, not the stream length. The natural package is a LongMemEval-style slice built with an LLM. \\
    \addlinespace
    OrgAccess~\cite{sanyal2026orgaccess} &
      Expected responses are largely LLM-generated and human-reviewed. There are no grant-then-revoke sequences or reinstatement. Competing directives are mentioned without a defined priority order. Expiry dates and temporal access limits appear in examples. Order pairs, a correct-state control, and memory writing are not described. The controlled variable is the number of concurrent permissions (1, 3, or 5). \\
    \addlinespace
    LongMemEval~\cite{wu2025longmemeval} &
      Questions are filtered and rewritten by human experts. The knowledge-update type tests changed user information. Reinstatement, source priority, expiry, and order-only pairs are not described. The oracle setting gives evidence sessions only, which is not the same as the correct current state. Sessions are processed online before the question. The S and M settings differ in history length. Evidence sessions are simulated by LLM self-chat, while a quarter of the filler sessions come from ShareGPT, so the real-log cell is ambiguous. \\
    \addlinespace
    PI-LLM~\cite{wang2025unable} &
      The target is defined by rule as the value at the last occurrence of each key. Keys are overwritten repeatedly, and the target is always the last update, so no earlier value is restored. Source priority and expiry are not described. Random and sequential update orders are compared, but not as minimal pairs. No condition shows only the final values. The keys to track are given before the stream. The number of updates varies from 3 to 400, and one experiment holds input length fixed. The key--value streams are synthetic. \\
    \bottomrule
  \end{tabular}
\end{table}
}{}

\section{Pre-registration log}
\label{app:prereg}

Every test below was written with its threshold and seed range before the corresponding run. Except for R5-3 and R5-4d, whose registrations include a joint Holm adjustment, each registered test is interpreted on its own, without correction for multiplicity. Across the eight registered second- and third-round superiority comparisons (filter or state against an alternative, R2-A, R2-C2b, R3-1 and R3-2, two readers each), a Holm adjustment of one-sided bootstrap p-values for a positive difference keeps five significant. The lead of the filter over the 4096-token trace for phi-4 ($p=0.017$ against a Holm threshold of 0.0167) does not survive, and the two comparisons that failed as registered remain failed. Rows marked $^\dagger$ were designed after seeing the result of an earlier test, which the row names, and before their own run. Rows have one of three statuses. Confirmatory rows carry a threshold registered before the quantity they test had been computed on any of the streams they use: E0, E1, X1a, X1b, X2, X3a, X3b, X4a, X4 recency, X4b, X5a--c, X6a, X6b, X8a, X9, R3-1, R3-2, R3-3, R3-4, R3-6, R4-1, R4-3, and the round-five rows R5-1, R5-3, R5-4 and R5-5. R5-6 is diagnostic: it was registered before its quantities were computed, after the result of R4-3 was known. Several of them reuse streams that earlier rows had read with other memories (the X6 streams from R2 to R4, the X4c streams in R3-3), which is weaker evidence than fresh streams. R5-1, R5-4 and the 512-event test of R5-5 use fresh seeds. R5-3 reuses the X6 streams, the second seed of R5-5 the R4-1 pairs, and R5-6 the round-three parses. Adaptive rows were designed after an earlier result: X4c$^\dagger$, written after X4a failed because its uncontrolled background let the state grow with $T$; R2-A$^\dagger$, written after the 768-token traces of X7 stopped before a decision; R2-C2a$^\dagger$ and R2-C2b$^\dagger$, written after R2-C showed that the second domain's renderer left expiry to the reader; E2-0b, whose stopping consequence was changed after E2-0 failed; and R4-4b, written after R4-2 had exposed a gap in the update format. Report-only rows have no threshold: X1c, X7, X10, R2-B1, R2-B2, R2-B3, R2-C, R2-D and R4-2. Within R3-6 and R4-2 the corrected or changed variants were found after the run and are labelled as such. The log is a dated text file kept with the code. Entries before the fifth round have no external timestamp, so their dates rest on our record. Before any fifth-round run, the log including the fifth-round registrations and the hashes of their code and frozen predictions was hashed, and the hash was timestamped through OpenTimestamps on 28 October 2026; the proof is released with the log. Revisions made after smoke tests and before formal runs are part of the log and are listed where they matter. The full text, including reasons for each revision and the earlier validation-range decisions, is released with the code as \texttt{PREREG.md}.

{\small
\begin{longtable}{p{2.2cm}p{6.8cm}p{4.6cm}}
\toprule
Test & Registered criterion & Outcome \\
\midrule
\endhead
E0 & Gradient retention beats longest-line and BM25 retention on five hand-written scenarios & Tied with longest line, demoted \\
E1 (gate 4) & Rank-one or query-averaged Fisher quadratic predicts KL at least 10\% better than the best first-order predictor & Failed, mechanism claim terminated \\
E2-0, E2-0b & Full-context reader recalls both classes and reinstatements at 0.90 at 64 events & Failed for Qwen 7B and 14B \\
X1a & Reader AUC decreases with $T$ & Holds for all three readers \\
X1b & Neutral persona changes balanced accuracy at 64 events by 0.10 or more & No ($+0.008$) \\
X2 & State reading at 0.95 or above at every $T$ and flat in $T$ & Fails: below 0.95 for all readers; for phi-4 no detectable dependence on $T$ (no equivalence test) \\
X1c & Report which of same-key events, other-key events and length dominates & Length, same-key effect not detectable \\
X5a & Order-oblivious rules err at least 0.40 at $m=65$, $b=2$ & Holds with both readers \\
X5b & Key recency errs $\tfrac12\pm0.08$ on blocked chains with $b'\ge b$ & Holds (exactly 0.50, as the proposition implies for the exact reader, so this checks the implementation) \\
X5c & Last applied event is exact with the exact reader, at most 0.10 error with the LLM reader & Holds (0.00, an implementation check, and 0.01) \\
X3a & LLM ledger beats the best non-filter memory by 0.10 at 256 events, $B=256$ & Failed ($-0.05$) \\
X3b & Gradient eviction degrades more over the horizon than per-key recency & Inconclusive \\
X4a & State memory threshold flat in $T$ and monotone in $K$ (uncontrolled background) & Failed for the state memory at $K=15$ and for newest-event-per-key retention throughout \\
X4 recency & Recency threshold grows with $T$ or exceeds the grid at 256 events & Holds \\
X4c$^\dagger$ & Same as X4a with a controlled background, plus last-applied retention & Holds for the state memory. Holds only formally for last-applied retention, which reaches 0.85 only for $K=1$, and for $K=3$ at 256 events \\
X4b & Report the LLM ledger on the X4c grid (revised to match X4c before running) & At most 0.65 in every cell \\
X6a & Parser exact on at least 95\% of event lines & Holds (35{,}743 of 35{,}744) \\
X6b & Parsed filter beats the best non-filter memory by at least 0.10 (point estimate) with a 95\% lower limit above 0 at 256 events & Holds for both readers (lower limits 0.10 and 0.05) \\
X7 & Report a reasoning-trace reader on full logs & Reported \\
X8a & Compact code threshold flat in $T$ and monotone in $K$ & Failed (readers do not decode it reliably) \\
X9 & Attaining rule within 0.05 of the bound & Holds (at most 0.029) \\
X10 & Report parsing and reading on paraphrased logs & Reported \\
R2-A$^\dagger$ & Parsed filter beats a reasoning trace of up to 4096 tokens over the full log by at least 0.10 with a 95\% lower limit above 0 at 256 events, each reader & Holds ($+0.16$ $[0.04,0.28]$ and $+0.11$ $[0.03,0.21]$) \\
R2-B2 & Report transition code written by the reader models from the policy text, three samples each, run on a disposable machine & Reported. First domain: no sample's own decision above 0.56 (reading a code-rendered state reaches 0.84 for one sample at 64 events and 0.50 at 256). Second domain: best samples 0.86 to 0.94, every model has a failing sample \\
R2-C & Report whether decay of log reading, flat state reading and the lead of the filter over long context recur on an independently specified domain & Decay and the lead over the raw log recur. State reading drops by 0.08 to 0.09 with $T$ (intervals include zero), and the lead over a reasoning trace is not significant \\
R2-B1, R2-B3 & Report the reader model applying the parsed updates to a state table, and a Mem0-style fact store, against the filter and the ledger & Reported. 0.58 to 0.59 and 0.54 to 0.59 at 256 events, 0.21 to 0.30 below the filter (Qwen $+0.21$ $[0.09,0.34]$ and $+0.21$ $[0.10,0.33]$, phi-4 $+0.26$ $[0.15,0.38]$ and $+0.30$ $[0.18,0.43]$) \\
R2-C2a$^\dagger$ & With expiry resolved in the second domain's rendering, state reading on the expiry family at least 0.80 at 256 events, each reader & Holds for phi-4 (0.86), fails for Qwen 14B (0.50) \\
R2-C2b$^\dagger$ & Same rendering, parsed filter beats a reasoning trace of up to 4096 tokens by at least 0.10 with a 95\% lower limit above 0 at 256 events, each reader & Holds for phi-4 ($+0.21$ $[0.08,0.35]$), fails for Qwen 14B ($+0.06$ $[-0.04,0.18]$) \\
R2-D & Report whether Qwen2.5-32B changes the decay of log reading and the lead of the filter over long context & Decay recurs (AUC 0.96 to 0.65), state AUC at least 0.99, filter leads retrieval by $+0.21$. With a reasoning trace of up to 4096 tokens the 32B reader reaches 0.68, and the filter leads it by $+0.21$ $[0.11,0.33]$ \\
R3-1 & Reasoning-trained 32B readers: state from parsed events beats the full log by at least 0.10 with a 95\% lower limit above 0 at 256 events, each reader & Holds ($+0.13$ $[0.04,0.21]$ QwQ-32B, $+0.26$ $[0.15,0.38]$ R1-Distill-Qwen-32B) \\
R3-2 & Filter beats the best of four ledger protocols (code-kept table with 16 or 1 updates per call, whole-table rewrite with 1 update, log sorted by key) by at least 0.10 with lower limit above 0 at 256 events, each reader & Holds for phi-4 ($+0.20$ $[0.09,0.26]$), fails for Qwen 14B ($+0.10$ $[-0.03,0.19]$) \\
R3-3 & Budgeted filters follow the registered prediction: mean gap at most 0.05 with 90\% of cells within 0.10, chance with nothing kept, a straight line under random eviction, random eviction better than recency & All four hold for both readers (mean gap 0.026 and 0.030) \\
R3-4 & A code model's rule, selected on 50 labelled histories and repaired once, within 0.05 of the human-written rule at 256 events, each domain & Holds on the second domain (1.00), fails on the first ($-0.18$ $[-0.25,-0.10]$) \\
R3-6 & Parse errors predicted from calibration rates: per-cell error within $\max(0.03,2\,\mathrm{SE})$, ordering with Spearman at least 0.8, reader accuracy within 0.10 & Ordering holds (0.83). Cells hold in 7 of 10 (5 of the 8 registered) and reader accuracy in 16 of 20 as registered, 9 of 10 (7 of 8) and 18 of 20 after correcting a substitution error found after the run \\
R4-1 & Stronger readers (gpt-oss-120b, Qwen2.5-72B, QwQ-32B on the log only) at 256 and 1{,}024 events. Rule fixed in advance: state from parsed events leads the best log reader by at least 0.10 with lower limit above 0 at 1{,}024 events keeps the claim; at 1{,}024 but not at 256 narrows it to beyond the reader's horizon; at neither shrinks it to cost and small readers & Best log reader gpt-oss-120b at both lengths: $+0.04$ $[0.01,0.08]$ at 256 (not met), $+0.12$ $[0.07,0.17]$ at 1{,}024 (met). Claim narrowed as registered, title changed accordingly \\
R4-2 & A rule written by an agent that sees only the policy text: disagreement with the exact replay, and accuracy on Qwen 14B parses (report only) & Written by Claude Opus 5.5. 257 of 2000 histories disagree, all in the restore family, from one gap in the update format, and 0 of the 248 priority histories. On Qwen 14B parses 1.00 and 0.85 against 1.00 and 0.98 ($-0.13$ $[-0.20,-0.06]$ at 256). After the result, with that gap closed: 0 of 2000 and 1.00, 0.98 \\
R4-3 (H-BF1--3) & Calibrated Bayes filter beats the plug-in filter for the 0.5B and 1.5B parsers (upper limit below 0 in all 4 cells), within 0.15 of cross-validation, at most $+0.05$ worse for 3B, 7B and 14B & All hold (0.5B: $-0.27$ and $-0.15$; 1.5B: $-0.41$ and $-0.24$ in balanced error) \\
R4-3 (H-IND) & Template-level intraclass correlation of parse errors, against independence across lines & Independence rejected for 0.5B, 1.5B and 3B: errors follow the wording template \\
R4-4b & As R3-4 on the first domain, with one sentence added to the update format saying what a reconfirmation does & Holds at the boundary ($-0.05$ $[-0.10,-0.01]$); restore family 0.50 to 1.00, priority family still 0.50 \\
R5-1 & Bounded recursive store (at most $C$ records, evicted records deleted), fresh seeds, prediction frozen before reading: mean gap at most 0.05 with 90\% of cells within 0.10; chance with an empty store; below the projection at $K=15$, $T=256$ by at least half the predicted gap; accuracy falls from 64 to 256 events at a fixed capacity fraction $f$; least-recently-touched eviction within 0.05 of its projection & All five hold for both readers (mean gap 0.013 and 0.018, 147 of 147 cells within 0.10; $+0.11$ and $+0.10$ below the projection's prediction; $+0.16$ and $+0.15$ from 64 to 256 events) \\
R5-3 & Published memory designs and a query-time retriever on the 80 streams of \Cref{tab:buildable} at 256 events: the parsed filter cut to 512 tokens beats each comparator cut to 512 tokens (StateMem native, StateMem with the policy, native Mem0, the dependency retriever) by at least 0.10 with lower limit above 0, each reader, Holm over the eight tests; if the retriever is not beaten, the claim is restated as selection by dependencies & Met against StateMem and Mem0 for both readers ($+0.25$ to $+0.33$, all six significant after the adjustment). Not met against the retriever ($+0.01$ $[-0.11,0.13]$, $-0.08$ $[-0.18,0.03]$), and the claim is restated \\
R5-5 & Operational horizon of gpt-oss-120b: state from parsed events leads the log by at least 0.10 with lower limit above 0 at 512 events; a second sampling seed at 1{,}024 events reads the log within sampling error of the first & $+0.07$ $[0.03,0.11]$ at 512 (not met, horizon above 512). Seeds indistinguishable ($+0.03$ $[-0.03,0.08]$), but with the second seed the lead at 1{,}024 is $+0.09$ $[0.04,0.13]$, below the margin; both seeds together $+0.10$ $[0.06,0.14]$ \\
R5-4 & Second domain, frozen, fresh seeds and fresh wording by a model that neither parses nor reads: parsed state beats the log by at least 0.10 with lower limit above 0 at 256 events (each reader) and at 1{,}024 (gpt-oss-120b); bounded store within 0.10 of its prediction; parsed state beats a query-time retriever by at least 0.10 with lower limit above 0 at 256 events, Holm jointly with R5-3 & Holds for Qwen2.5-14B ($+0.22$) and phi-4 ($+0.35$), not for gpt-oss-120b at 256 ($+0.02$); holds for gpt-oss-120b at 1{,}024 ($+0.28$ $[0.21,0.36]$); bounded store 12 of 12 cells; retriever beaten for phi-4 ($+0.20$, significant in the joint Holm family of ten), not for Qwen ($+0.06$ $[0.00,0.12]$, $p=0.036$ against a threshold of 0.0167) \\
R5-6 & Diagnostic audit of the calibrated filter, registered before computing: query-blind version, exact joint filter, a memoryless confusion correction, held-out templates, shifted family priors; if sequential inference never beats the memoryless correction, the gain is called confusion correction & The memoryless correction ties the filter in all 10 cells and 8 weightings; joint equals product; query-blind version identical; held-out templates remove the 0.5B gain at 256 events \\
\bottomrule
\end{longtable}
}

\end{document}